*To be submitted*

# Building heterogeneous multistep workflows in automated microscopy: reward design and knowledge transfer for autonomous structure–property discovery

Sergei V. Kalinin[1,2]

[1] Department of Materials Science and Engineering, University of Tennessee, Knoxville, TN 37996, USA
[2] Department of Electrical Engineering and Computer Science, University of Tennessee, Knoxville, TN 37996, USA

## Abstract

Autonomous microscopy can turn fast acquisition into scientific progress when it decides not only where to measure, but also what information to acquire, at what scan rate, and when to recalibrate the instrument. We develop a budget-constrained framework for such multistep structure–property workflows. It combines explicit scientific rewards, action costs, instrument observation models, transfer of pre-experiment knowledge into the online campaign, and evaluation kept independent of the controller. The transferred knowledge includes predictor accuracy and computational cost, empirical learner-maturity functions that weight model-derived rewards, and drift characteristics established before controller training. Using a real electron energy-loss spectroscopy dataset, we run two complementary campaigns. In the first, the controller chooses between imaging and spectroscopy, image size, and scan speed. In the second, the instrument drifts and the controller also decides when to recalibrate. The campaigns show that response discovery and structure–property learning favor different surrogate models and acquisition strategies. They also show that simple acquisition rules form strong, practical baselines. Independent error analysis reveal where model uncertainty is and is not a reliable learning signal. In the drift campaign, explicit registration greatly improves positional accuracy, and the framework measures what this costs in spectroscopy throughput. Together, these results provide a reproducible way to design, compare, and audit autonomous experiments, and they define what microscopes must provide for deployment: independently executable actions, characterized scan and detector responses, timestamped registration telemetry, and complete action records.

## 1. Introduction

Automation in microscopy offers more than the replacement of repetitive manual operations. When instrument electronics and detectors can acquire data faster than an operator can inspect them and choose the next action, manual decision-making leaves part of the instrumental capability unused. Connecting acquisition, analysis, and control can increase the amount of scientifically useful work completed during a session. Autonomous probe conditioning, extended scanning-probe operation, and data-driven electron-microscope alignment illustrate different ways of reducing this bottleneck [1–5]. Self-driving laboratories have demonstrated the same benefit in chemistry and materials science, where closed-loop selection of experiments accelerates synthesis and discovery [6–11]. However, the objective of any experiment is not simply to collect more data, but to increase the rate at which measurements answer a scientific question while retaining the context needed to interpret and reuse them.

An effective automated workflow must combine a meaningful scientific objective with practical timing constraints and a record of its decisions. The record should identify what the system knew, which action it selected, the actual execution settings, and the observation that followed. The predictors themselves do not need to be simple or analytically interpretable, but the experiment as a whole must be possible to inspect, diagnose, and reproduce. Human-in-the-loop methods allow the scientist to change the target without issuing every instrument command, while linked instrument, sample, and decision histories support subsequent analysis and integration [2,12]. Such linked records complement the FAIR data principles and executable acquisition frameworks [13,14].

Important components of this vision have been demonstrated by several groups. Rashidi and Wolkow automated STM probe assessment and conditioning [3], and Krull and colleagues combined image classification, region selection, and probe maintenance in autonomous scanning-probe operation [4]. DeepFocus addresses focus and astigmatism correction in SEM [5], while BEACON and emittance-based Bayesian optimization address aberration correction in STEM [15,16]. Reward-driven tuning has also been demonstrated for tapping-mode scanning probe microscopy [17]. Moving from instrument adjustment to scientific sampling, autonomous piezoresponse spectroscopy has identified regions of strong electromechanical response [18], and image-conditioned DKL has guided discovery of ferroelectric structure–property relationships [19]. Roccapriore et al. applied this principle to EELS in nanoplasmonic systems [20], and related work extended scalarized objectives to four-dimensional STEM [21]. Similar closed-loop strategies are used in sparse-data-guided STEM [22], AI-driven scanning X-ray microscopy [23], Gaussian-process-driven acquisition at synchrotron and neutron facilities [24], and Bayesian tuning of accelerator-based light sources [25], and deep learning is now used throughout electron microscopy [26]. These studies largely separate the problem of maintaining a useful instrument state from that of deciding which scientific measurement to acquire next.

The available scientific objectives and the role of the operator have also broadened. Explainability and intervention strategies have been developed for autonomous scanning probe microscopy [27]. Active recommender systems allow a scientist to shape a spectral target during acquisition [28], and Scientific Exploration with Expert Knowledge (SEEK) incorporates task-specific human guidance [29]. Multi-objective optimization makes competing rewards explicit [30], while adaptive channel selection asks which structural measurement is most informative [31]. Curiosity-driven approaches use an auxiliary model of prediction error to seek unexplored structure–property relationships [32]. Hypothesis learning goes a step further and uses the experiment itself to discriminate between competing physical models. Ziatdinov et al. used it to select measurements on a combinatorial Sm-doped $BiFeO_3$ library, ranking candidate models of the composition-driven phase transition as data accumulated [33]. Liu et al. implemented hypothesis learning in autonomous piezoresponse force microscopy and showed that domain growth in a $BaTiO_3$ film is controlled by the kinetics of domain-wall motion rather than by thermodynamics or screening [34]. Slautin et al. removed the need for a predefined set of models: candidate equations are generated by symbolic regression from sparse measurements and ranked for physical plausibility by a large language model, which led to a creep-like domain-growth law in a PZT film [35]. The relevant question is therefore not only whether an algorithm can optimize a chosen statistic, but whether that statistic represents the scientific progress sought by the experiment.

Until now, these workflows are universally myopic. At each step they select the measurement with the highest immediate acquisition score, while the type, size, and fidelity of the measurement are fixed in advance. This is particularly true for structure–property discovery: image-conditioned spectroscopy workflows assume that the structural image and its representation (i.e. state variables) are already available, and the only decision is which spectrum to acquire next [19–21]. A real experiment can also change the image window, the scan speed and hence the image fidelity, and the division of time between imaging and spectroscopy, and it must maintain the instrument, for example by correcting drift. These decisions have delayed consequences: in the present supervised workflow, an image earns no immediate reward but can improve the selection and interpretation of a later spectrum, and time spent on recalibration is time not spent on measurement. The surrogate model adds a further time constraint, since a model that performs well on a complete dataset may be too slow to update, or unreliable, early in acquisition. Making such decisions requires a description of how each observation is produced, what it costs, and how it changes the usefulness of later measurements, that is, a non-myopic formulation. Cost-aware and multi-fidelity Bayesian optimization formalize related choices about which observation to acquire and at what cost [36–40], and reinforcement learning provides a general framework for sequential decisions with delayed rewards [41].

This larger decision space makes prior knowledge valuable. Online DKL learns a structure–property relationship from the spectra acquired during the experiment, rather than relying on a predictor previously fitted to the same specimen [19,20,42]. It should not also have to discover every useful descriptor, training schedule, and instrument-response model by trial and error on the microscope. Existing paired datasets can establish likely sample requirements and computational trade-offs before acquisition begins. Instrument knowledge describes how an action produces an observation; scientific knowledge describes the structure–property relationship inferred from that observation. Such information can initialize later workflows, but it must be kept separate from the unmeasured spectral responses and full-data references used to evaluate them. Similarly, problem-specific segmentation or feature extraction can make the surrogate model more data-efficient, but it requires prior knowledge in the form of pretrained models. These two kinds of prior knowledge should be validated separately when the specimen or instrument changes. Learning-curve extrapolation offers a complementary way to decide how much model-development effort is worthwhile before committing further computation [43].

Here we develop and evaluate a common workflow framework through two separate campaigns built on the same real image–EELS dataset. In the first, the experimental degrees of freedom are the measurement location, the choice between imaging and spectroscopy, the image-window size, and the scan duration, and image and spectrum positions are assumed to be perfectly registered. We examine how the spatial support and numerical representation of the structural descriptor, the accuracy and uncertainty of the regressor, and the available acquisition time jointly determine the outcome. In the second campaign, the window size is fixed, but the instrument drifts, and the controller must decide when to image, when to recalibrate its position by image registration, and when to acquire a spectrum. In both campaigns, we separate finding a strong local response from learning the structure–property relationship across the specimen, and independently assess model prediction, reward alignment, and completed-campaign performance. Finally, we identify which information can be established from pre-acquired data, which quantities can serve only as evaluation references, and what the microscope and its manufacturer must provide before such workflows can be deployed.

## 2. Experimental data and a configurable measurement workflow

Both campaigns use the same pre-acquired plasmonic image–EELS dataset [20]. Electron energy-loss spectroscopy in the scanning transmission electron microscope maps localized surface plasmons with nanometer resolution [44,45], so the spectral response depends strongly on the local particle arrangement. We use these data to ask how an experiment should divide its time between observing structure and measuring a spectral response. The structural image and the spatial distribution of the selected band area are shown in Figure 1a,b. Example spectra in Figure 1c show the target interval, and Figure 1d shows a second response map integrated over 0.8–1.0 eV. The task is to predict the band response from the local structure while deciding which images and spectra are worth acquiring. Sections 2–5 describe the first campaign, in which image and spectrum positions are perfectly registered; Section 6 describes the second campaign, which adds instrument drift. The dataset, image reconstruction, and data partition are described in Supplementary Section S1; no new microscope data were acquired for this work.

The scalar response at candidate location $i$ is the integrated spectral intensity over a fixed interval:

$$y_i = \int_{E_1}^{E_2} S\,(E, \mathbf{r}_i)\, dE. \qquad (1)$$

The integration interval (shaded in Figure 1c) is the same for all predictors and campaigns. It defines an operational target and does not imply assignment to a particular plasmon mode; relating an integrated band to a specific excitation requires separate energy calibration and analysis [46]. The 0.8–1.0 eV map in Figure 1d is shown only for comparison, to illustrate that a different choice of band gives a different response landscape and hence a different scientific objective; it is not used for training. The spectra in Figure 1c are the stored, downsampled measurements, whereas the training targets are the band integrals stored with the original dataset. The complete response map is available to the evaluator, but the online learner sees a spectrum only after it has been acquired.

The workflow treats acquiring structural information and acquiring a label as separate actions (Figure 1e). An acquisition rule selects a candidate location based on the information currently available. A controller then chooses the operation and, for imaging, the window side $L$ and the relative scan time $\tau$:

$$a_t \in \{\mathrm{IMAGE}(i, L, \tau),\ \mathrm{SPECTRUM}(i)\}. \qquad (2)$$

IMAGE updates the structural descriptor at the selected center but does not reveal a spectrum. After an image, the controller may image another candidate or acquire a spectrum at any previously imaged center. SPECTRUM adds one measured image–response pair and updates the predictor. A coarse preview of the whole field, acquired at the start, allows candidates to be ranked before they are imaged in detail. The learner only ever sees the observations it has actually paid for; neither clean hidden images nor unmeasured responses are used in their place.

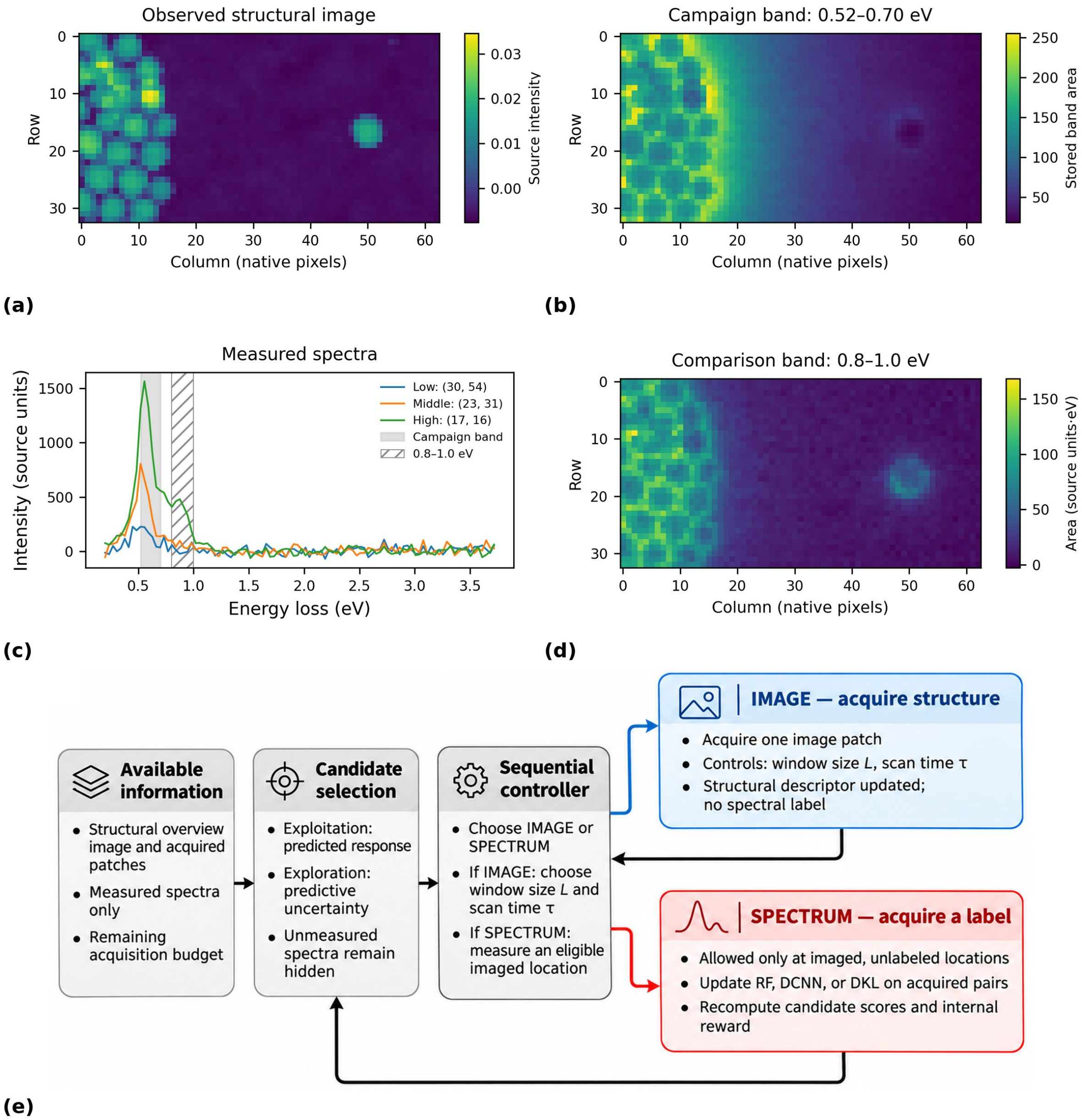

**Figure 1.** Experimental data and configurable acquisition. (a) Observed structural image. (b) Stored campaign target over 0.52–0.70 eV. (c) Three measured 100-channel spectra from the selected low-, middle-, and high-area locations, with both intervals indicated. (d) Comparison map for 0.8–1.0 eV, using linear interpolation at the band endpoints and trapezoidal integration of the cached spectra; missing display pixels in (b,d) are filled by the mean of available neighboring map values. The comparison map and gap filling are for display only: no missing spectrum is added to training or evaluation, and no original target, predictor, or policy is changed. (e) Independent IMAGE/SPECTRUM operations, candidate selection, tunable image-window size and scan time, and measurement feedback.

Because the data were acquired in advance, changes in imaging conditions are represented by an observation model rather than by new live measurements. A window is extracted from the observed image and blurred along the fast-scan direction, with larger blur assigned to faster scanning:

$$\tilde{I}_{i,L,\tau} = \mathcal{B}_{\sigma(\tau)}\left[\mathcal{C}_{i,L}(I)\right]. \tag{3}$$

Here, $\mathcal{C}_{i,L}$ extracts the window of side $L$ centered at candidate $i$, and $\mathcal{B}_{\sigma(\tau)}$ applies a Gaussian blur of width $\sigma(\tau)$ along the fast-scan direction before the numerical descriptor is formed. In the principal scenario, imaging time grows with window side and selected scan duration, whereas a spectrum has a separate fixed cost. These assumptions convert the static data into a controllable acquisition environment; they are not measured characteristics of the microscope. In a live workflow, the observation operator and costs should be obtained from the measured response of the instrument, including the dependence on scan direction, dwell time, detector bandwidth, and actual execution timing. The exact laws and numerical constants are given in Supplementary Section S1.

In this first campaign, the simulator retains exact image–spectrum registration and a fixed specimen. It does not include drift (introduced in the second campaign, Section 6), sample evolution, dose damage, repeated measurements, or additional spectral noise. Only labelled pairs update the scientific model; images without spectra are not used to train a structural representation. Offline calibration and online measurements are kept separate, and the finite acquisition budget includes the declared initialization. These restrictions isolate the choice of information, spatial context, and fidelity; they are not intended as a complete dynamical model of the microscope.

## 3. Pre-experiment calibration, empirical maturity, and reference performance

Before sequential acquisition can be evaluated, the structural descriptor and scientific predictor must be chosen. We use the pre-acquired dataset to compare their accuracy, computational cost, and sample requirements, and to establish reference performance. This stage provides practical settings for an online experiment while distinguishing them from full-data quantities that are available only for retrospective evaluation.

### 3.1 Structural descriptors and uncertainty-aware regressors

The first task is to choose how much structure the descriptor contains and how that structure is represented numerically. A larger physical window supplies more context, whereas a finer numerical grid supplies more features; these are not the same choice. Previous image-conditioned autonomous experiments used DKL to map image patches to a spectral response [19–21,42], but DKL is not the only suitable regressor. For uncertainty-guided active learning [47–49], a model must provide both a predicted response and an uncertainty estimate: the prediction identifies promising responses, and the uncertainty indicates where additional data may be informative. Prediction-variance reduction is a classical active-learning criterion whose interpretation depends on the statistical model and evaluation domain [50].

We compare three different ways of meeting these requirements. RF is relatively inexpensive and uses variation among bootstrap-trained trees as an uncertainty proxy [51]. DCNN learns a convolutional representation and supplies a direct response prediction; it has no intrinsic posterior variance, so we estimate uncertainty by repeated dropout evaluations of its regression head [52,53]. DKL learns a representation jointly with a Gaussian process and provides a conditional posterior variance in that representation [42]. This comparison tests a lightweight ensemble, a flexible deterministic predictor with an added uncertainty approximation, and a probabilistic kernel model. Deep ensembles are another option for estimating uncertainty but are not used here [54].

The DCNN uncertainty estimate may be less dependable than its mean prediction because head dropout explores only part of the fitted model. DKL provides a more explicit probabilistic construction, but its variance is also conditional on a learned representation and fitted parameters. Neither construction guarantees calibration under changing inputs or adaptive sampling [55]. Prediction accuracy and uncertainty quality must therefore be assessed separately. Numerical profiles, optimization settings, and the calibration-based rule for selecting a practical online model are given in Supplementary Sections S2–S3. Learned-kernel uncertainty must also distinguish conditional GP variance from uncertainty in the optimized representation [56,57].

### 3.2 Calibration, computational trade-offs, and reference performance

Pre-experiment calibration answers practical questions: which representation predicts the response, which model can be updated fast enough, and how much data are needed before the prediction becomes useful. The accuracy–training-time Pareto frontier identifies settings that cannot be made both faster and more accurate within the tested set. These settings, learning curves, and observation/cost models can initialize an experiment. They must not be confused with the complete response map or full-data reference errors, which are available only to the evaluator (Table 1). Related cost-aware model-selection and learning-curve approaches motivate measuring these trade-offs explicitly [37,43].

**Table 1. Separation of pre-experiment knowledge, online information, and evaluator-only references.**

| Information | Role in the workflow | Restriction |
|---|---|---|
| Calibration-based profile and maturity parameter | Fix a practical predictor and learning-stage weight | Not evidence of transfer to a new specimen |
| Observation and cost laws | Define available actions and their consequences | Assumed here; require instrument calibration |
| Acquired images and measured spectra | Fit the online learner and rank candidates | Revealed only through legal actions |
| Full-data predictor reference | Measure family-specific prediction gap | Never an online target or reward |
| Maximum accessible measured response | Measure discovery regret and success | Hidden until post-campaign evaluation |

We retain one empirical full-data prediction reference for each regressor family. Its configuration is selected using calibration data, and its error is then assessed on held-out responses. The difference between this reference and an online model shows how far a campaign remains from a well-informed predictor in that family. It is not a guaranteed lower bound, and it can include differences in both training-set size and model configuration. The discovery reference is different: it is the largest response present in the accessible dataset, independent of the regressor. These two references answer different scientific questions.

Figure 2a shows that RF provides economical operating points, while DKL extends the tested accuracy–computation frontier toward lower error. The tested DCNN settings do not lie on this particular Pareto frontier. Increasing network size alone is therefore not a useful selection rule: a more costly network can fail to improve prediction. This conclusion applies to the screened representations and training schedules, not to every possible implementation of the three model families.

DKL gives the strongest selected full-data reference in this dataset, but that does not determine the best model for a short acquisition campaign. Matched-condition evaluation asks what can be predicted when training and evaluation use the same image quality; common-input evaluation compares the models on an unchanged structural representation. Both are needed because a live campaign accumulates images of different sizes and fidelities. Complete profiles, errors, timings, and descriptor landscapes are reported in Supplementary Section S3 and Figures S2–S3.

### 3.3 Learner maturity

Early in an experiment, the regressor is fitted to few measured spectra, so its learning signal may change substantially after each new observation. We introduce maturity to describe this stage of learning and to reduce the influence of model-derived rewards before sufficient labelled data have accumulated. Maturity is calibrated from the sample-count dependence of prediction error, rather than assigned from the nominal model size. For regressor $m$, the empirical learning curve and its associated weight are

$$E_m(N) = E_{\infty,m} + \left(E_{0,m} - E_{\infty,m}\right)\frac{N_{0,m}}{N + N_{0,m}}, \tag{4a}$$

$$\omega_m(N) = \frac{N}{N + N_{0,m}}. \tag{4b}$$

Here, $N$ is the number of measured spectra, $E_{0,m}$ is the baseline error, $E_{\infty,m}$ is the fitted asymptote, and $N_{0,m}$ is the sample count at which half the fitted recoverable improvement has been obtained. Figure 2b shows that RF approaches a useful error level relatively quickly, while DKL requires more data to develop its stronger full-data predictor. The compact DCNN saturates comparatively early but at a less favorable error level. Early saturation is therefore not synonymous with a better regressor. Fitted parameters and their variability are given in Supplementary Table S3.

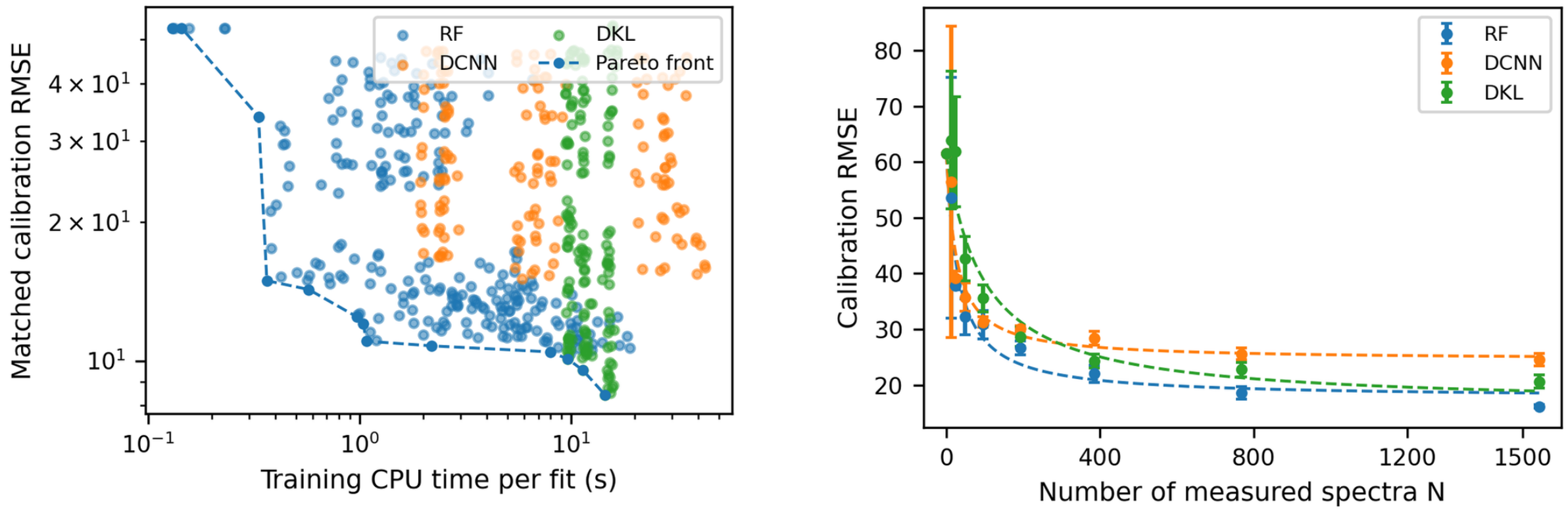

**(a)** **(b)**

**Figure 2.** Pre-experiment calibration. (a) Matched-condition accuracy versus measured training CPU time for the screened RF, DCNN, and DKL configurations, with the non-dominated frontier. (b) Learning curves for the primary online models and their fitted saturation functions. The maturity parameter describes learning stage; the fitted asymptote is distinct from the separately selected full-data reference. Detailed settings and numerical results are in Supplementary Section S3.

Maturity does not measure the probability that a particular prediction is correct. A model can mature toward an unfavorable error level, and a mature model can still be overconfident. The weight increases only when a spectrum is acquired, not after an image-only action. It is held fixed as a function of sample count throughout the campaign, rather than refitted after each change in window size or sampling pattern. Independent prediction-error measurements are therefore still needed to determine whether a maturity-weighted reward reflects useful learning. The fitted curve describes learning stage [43], not the probability that the acquisition ranking is correct.

## 4. Sequential objectives, acquisition functions, and reward design

The sequential problem combines two decisions: which candidate is scientifically interesting and which instrument action is worth taking next. We first define the learning or discovery objective, then distinguish the acquisition score used before a measurement from the reward obtained afterwards. Finally, we describe how the controller uses those rewards and the remaining budget to allocate imaging and spectroscopy, while full-data references remain outside its decision process.

### 4.1 Scientific objective and candidate ranking

We consider two limiting objectives separately. Exploitation seeks a high response in the selected spectral interval; pure exploration seeks to reduce predictive uncertainty. The predictor returns the mean $\mu_t(i)$ and a variance-like quantity $v_t(i)$ for each currently available candidate descriptor. Candidate scores are

$$A_t^{\text{exploit}}(i) = \frac{\mu_t(i) - \bar{y}_0}{s_0}, \tag{5a}$$

$$A_t^{\text{explore}}(i) = \frac{\sqrt{v_t(i)}}{s_0}, \tag{5b}$$

$$s_0 = \max(\text{std}(y_{\text{seed}}), 1). \tag{5c}$$

The seed mean and scale are fixed after initialization. For an image action, the highest-scoring unseen candidate is chosen from its coarse preview. For a spectrum, the choice is restricted to already imaged, unlabelled candidates and uses their acquired descriptors. A high predicted mean favors a strong response; a high uncertainty favors a measurement about which the model is less certain. The score does not anticipate what each possible scan setting would reveal, which would require a separate model of unacquired images. The response-driven score is greedy mean selection, not expected improvement.

This construction keeps acquisition functions and rewards separate. The acquisition function ranks prospective candidates before a measurement. The reward evaluates the realized change after an action. The controller, rather than the acquisition score, selects IMAGE versus SPECTRUM and the imaging settings. The implementation is thus a composition of model-based candidate selection and learned operation allocation, allowing the value of each component to be examined against simpler baselines.

### 4.2 Predictive uncertainty as a learning reward

The natural learning objective is to reduce the prediction error over the whole specimen. In a live experiment, however, this error cannot be computed, because the responses at unmeasured locations are unknown. Using the complete EELS response map to compute the reward would turn the task into a different problem, in which the controller has access to an oracle. Holding back a set of measured spectra for validation is possible, but it costs spectroscopy time, and an adaptive policy can overfit to a small validation set that it uses repeatedly. We therefore keep the true test error hidden from the controller and use it only for evaluation. As the observable exploration signal, we use the predictive uncertainty on a fixed panel of images whose acquisition is charged to the budget.

For a fixed panel $Q$ of 32 images, we define the normalized uncertainty

$$U_t = \frac{1}{|Q|\, s_0^2} \sum_{i \in Q} v_t\,(i). \tag{6}$$

The panel membership and descriptor quality do not change during a campaign. This avoids artificial uncertainty reduction caused merely by deleting difficult candidates or evaluating the model on an easier image configuration. Reference spectra are not revealed online. RF tree variance, DCNN head-dropout variance, and DKL latent variance nevertheless retain different meanings and scales. The fixed panel makes their temporal changes reproducible, not automatically calibrated or numerically comparable across families. Integrated variance reduction has a classical experimental-design interpretation [50], but the present estimators remain subject to misspecification and changing inputs [54–58].

The exploration reward is

$$r_t^{\text{explore}} = \omega_m(N_t)\,[U_t - U_{t+1}], \tag{7}$$

where $N_t$ is the number of labels before the action. Negative rewards are retained. A newly added spectrum can increase estimated uncertainty after a from-scratch refit; suppressing such increases would change the implemented objective. Since IMAGE neither refits the predictor nor changes the reference panel, its immediate exploration reward is zero. Its potential benefit is delayed: the acquired descriptor can make a subsequent spectrum more useful. The choice of modality is therefore a genuine sequential decision, rather than an image and a spectrum bundled into a single action.

### 4.3 A directly observed reward for response discovery

Let $b_t$ be the best response among spectra already measured in the campaign. The exploitation reward is

$$r_t^{\text{exploit}} = (b_{t+1} - b_t)/s_0, \tag{8a}$$

$$b_t = \max\{y_i\colon\, i \in \mathcal{M}_t\}. \tag{8b}$$

A new record gives a positive reward; a lower response gives zero, as does an image. No maturity weight is applied, because an improved record is a measured fact rather than a statement about model confidence. The environment uses the stored response as the measurement and adds no spectral noise, so noise-induced bias in the record and repeated measurements are outside the present model.

Before any controller-specific scaling, the cumulative discovery reward therefore depends only on the initial and final records:

$$\sum_t r_t^{\text{exploit}} = \frac{b_T - b_0}{s_0}. \tag{9}$$

The exploration reward behaves differently. Because each uncertainty decrease is multiplied by a weight that grows with the number of labels, the rewards do not add up to the total change in uncertainty between the start and the end of the campaign. The cumulative exploration reward, the final unweighted uncertainty, and the final prediction error are therefore three different quantities. Maturity weighting changes how credit is distributed over time, but it cannot turn an uncertainty signal that is wrongly ordered with respect to scientific improvement into a correct one. For this reason, we report all three quantities in the results rather than treating a rising reward curve as evidence of a successful experiment. Potential-based reward shaping preserves the optimal policy under specific conditions [59]; no such guarantee holds for the time-varying multiplicative maturity weight used here.

### 4.4 Budget, controller, and independent performance measures

Time enters as a hard resource constraint,

$$\sum_t c\,(a_t) \le B, \tag{10}$$

where the sum includes initialization. No additional cost penalty is applied, and rewards are not divided by action duration. Instead, the cost determines which actions remain possible and how many further decisions can be made. An image is allowed only when an unseen candidate and space in the pending-image queue are available, and when enough time remains to image and then measure a spectrum. A spectrum requires an imaged, unlabelled candidate and sufficient remaining time. Actions of variable duration naturally lead to semi-Markov formulations [60]; here we retain the simpler finite-budget, undiscounted update.

The controller is an on-policy TD(0) actor–critic [61,62]. Its actor assigns probabilities to the feasible operation/settings choices; a separate acquisition score chooses the candidate center. Its critic estimates the value of the current experimental state: the cumulative reward expected from future actions before the budget is exhausted. This value is not the predicted spectral area, predictive variance, or hidden test error. In the implementation, the critic operates in a fixed rescaling of the scientific reward, denoted $\tilde{r}_t$.

The observable state has twenty components: a constant; remaining and elapsed budget fractions; the labelled-pair count; new image and spectrum counts; the pending-image count; reference-panel uncertainty; maturity; the best measured response; the largest acquisition scores among pending and unseen candidates; the unseen fraction; an objective indicator; the previous reward; the mean acquired window size and scan-time setting of labelled pairs; indicators for available pending and unseen candidates; and a logarithmic label count. Counts and scores are normalized before use. This state summarizes acquired information and workflow history; it contains no unmeasured responses or oracle errors. Exact normalizations are listed in Supplementary Table S13.

If $T$ denotes termination at the budget limit, the policy value and the one-step temporal-difference error are

$$V^{\pi}(s_t) = \mathbb{E}_{\pi}[\textstyle\sum_{k=t}^{T-1} \tilde{r}_k \mid s_t], \tag{11a}$$

$$\delta_t = \tilde{r}_t + V_w(s_{t+1}) - V_w(s_t), \qquad V_w(s_T) = 0. \tag{11b}$$

The terminal state has zero remaining value. After each real action, the sum of its reward and the critic's estimate for the next state is compared with the critic's previous estimate. The signed discrepancy updates the critic and the probability of the chosen action. An image can consequently have positive long-term value even though its immediate reward is zero, because it makes a useful later spectrum possible. Figure 3 shows the information flow. The deployed policy is frozen during evaluation; training uses one update per real operation, without replay. Numerical learning rates, reward scaling, and clipping are specified in Supplementary Section S4.

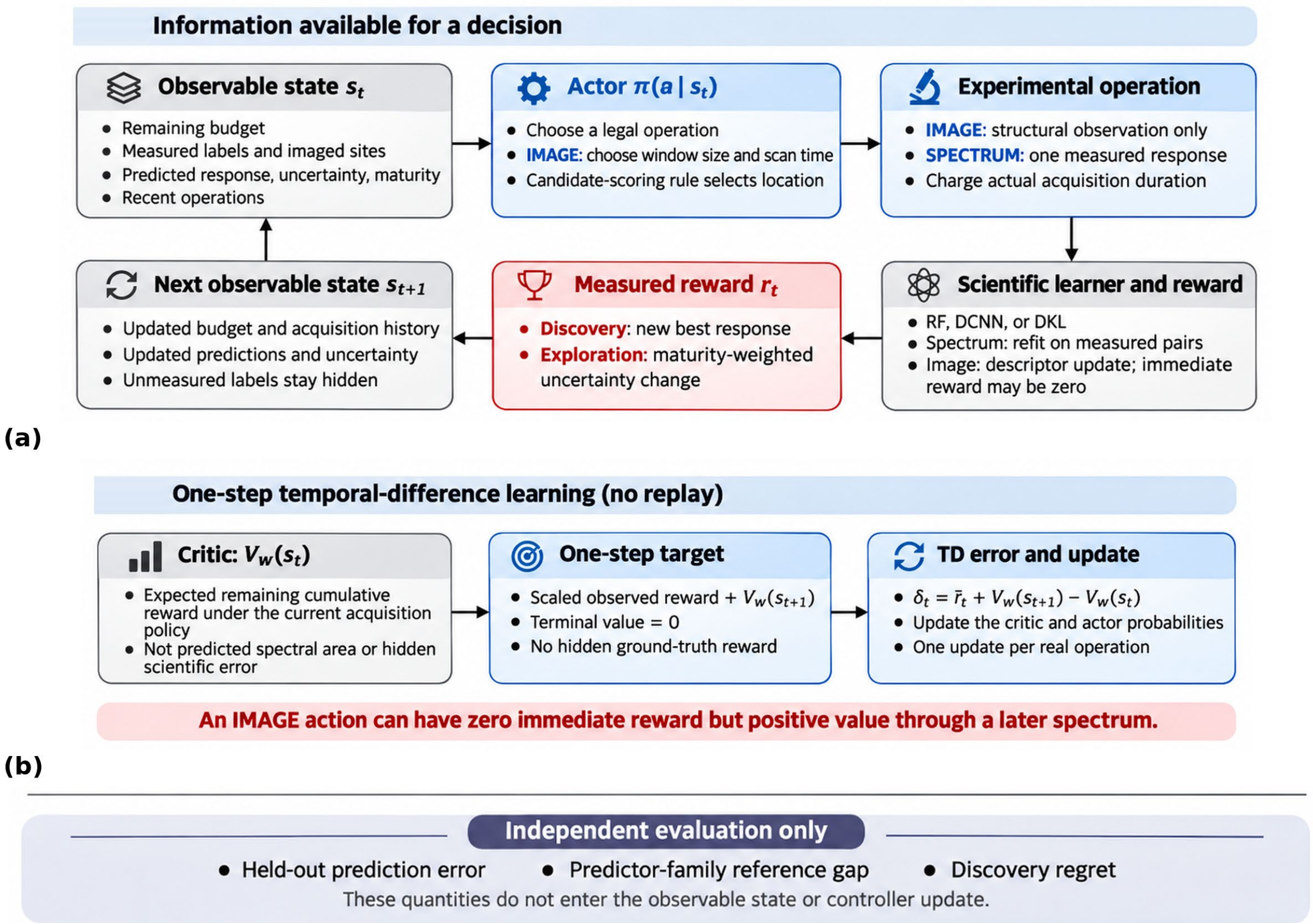

**Figure 3.** Observable state and TD(0) control. (a) Acquired images, measured spectra, model summaries, and the remaining budget define the state. The actor selects a legal operation and imaging settings, while the acquisition rule chooses the candidate location. Only spectroscopy adds a label and refits the scientific regressor. (b) The critic predicts the remaining cumulative reward in controller-scaled units. The difference between the observed one-step target and the current value updates actor and critic during training. IMAGE can receive delayed credit through a subsequent spectrum. Hidden prediction error and oracle gaps are evaluated separately and are not inputs to the controller.

For evaluation, held-out RMSE on common clean 17-pixel inputs supports comparison across model families. A second RMSE uses each family's fixed oracle descriptor and defines its reference gap, $G_{m,t} = E_{m,t} - E_{m,\text{ref}}$. Discovery is evaluated through $y_{\max} - b_t$, time to a strong response, and exact-maximum hits. The oracle descriptor can differ substantially from the mixed-quality training descriptors encountered online, so its gap combines sample limitations and descriptor/profile mismatch. Both evaluation panels are reported, regardless of which model each favors.

## 5. Learning, discovery, and the allocation of experimental effort

We now compare the completed campaigns by what they discover, what they learn across the specimen, and how they allocate the acquisition budget. All three regressors are evaluated under the same information restrictions. The comparison includes both learned operation policies and simpler acquisition-based controls, so improvements in candidate selection are not automatically attributed to reinforcement learning.

### 5.1 Campaign design and distinct scientific outcomes

For each predictor, separate policies were trained for response discovery and uncertainty-driven learning, then evaluated with frozen parameters on paired experimental initializations. The comparisons include acquisition-driven, random-candidate, and fixed-alternation controls under the same budget and access rules. These are repeated campaigns on one specimen, not independent specimens. Replication details are provided in Supplementary Section S4.

Figure 4 compares the strongest response actually measured with prediction error over the complete evaluation panel. Response-driven acquisition concentrates on promising candidates, whereas uncertainty-driven sampling can improve coverage of the structure–property relationship. Their success must be judged by both outcomes: discovering a high response does not establish a globally accurate predictor, and a lower average error does not guarantee discovery of the strongest response.

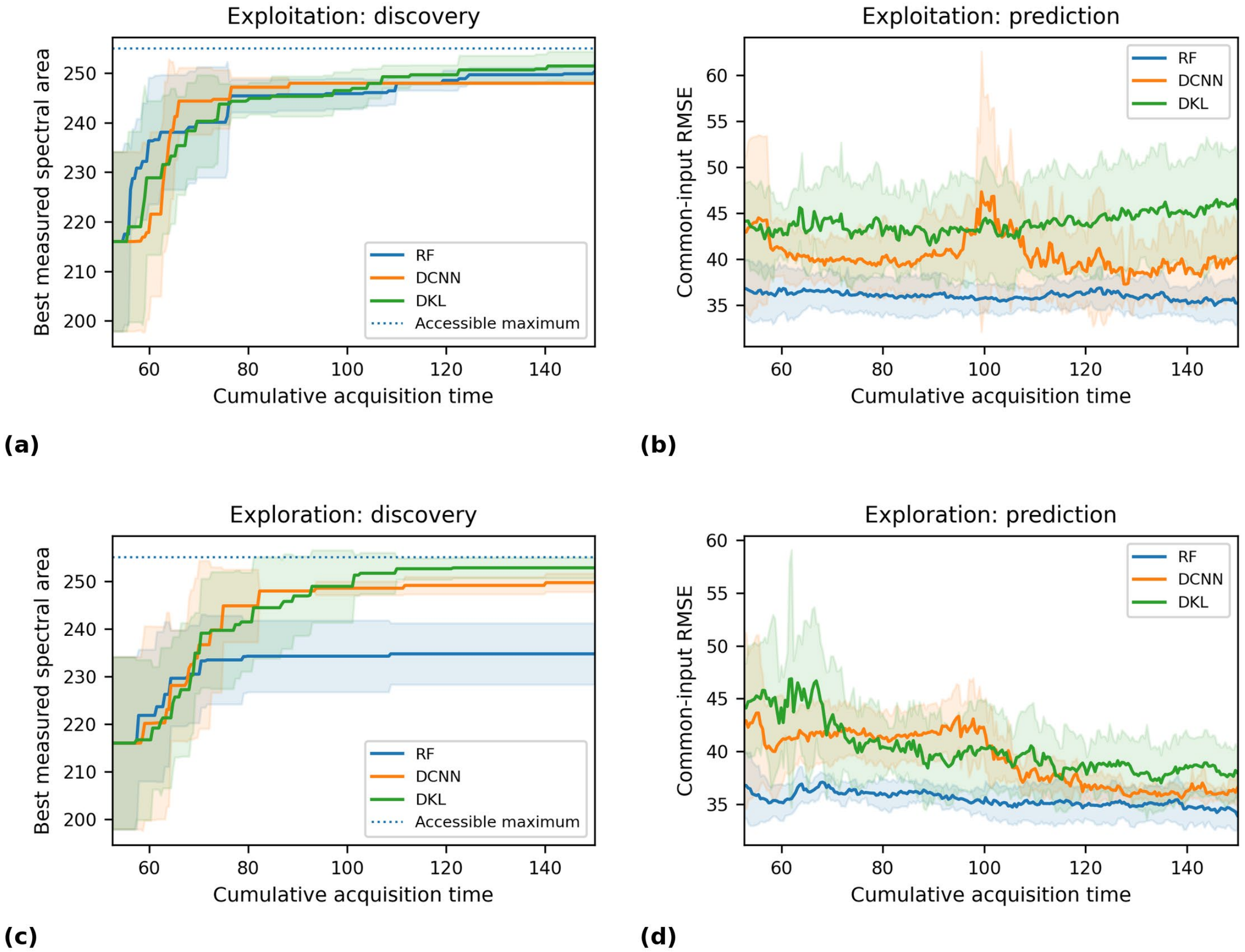

**Figure 4.** Discovery and global prediction under distinct scientific objectives. (a,b) Exploitation and (c,d) pure exploration, showing best actually measured spectral area (left) and RMSE on the common clean-L17 test panel (right). The dotted response reference is the accessible maximum, 255.0. Lines summarize twelve frozen-policy deployments per model/objective: two policy seeds are averaged within each of six experimental initializations, after which the mean and ± one standard deviation are plotted. The same 300 hidden test responses and initial acquisition cost are used. Curves start at the end of initialization, $t = 52.6$, and end at budget 150. Successful response discovery need not coincide with low error over the complete specimen.

RF exhibits the clearest objective trade-off: response-driven sampling finds stronger responses, whereas uncertainty-driven sampling gives lower global prediction error. In the tested

DCNN configuration, exploration improves both outcomes relative to greedy response sampling. DKL exploration is especially effective at discovering the strongest response, but RF remains the better finite-budget global predictor on the common input panel. Thus the best full-data model, the most successful discovery strategy, and the most accurate online predictor need not be the same choice. The numerical endpoints and their variability are given in Supplementary Table S4.

### 5.2 Value of learning the operation policy

The comparison isolates the additional value of learning the operation policy after candidate ranking has already used the regressor's predictions. Some baseline runs discover a stronger response even when the learned policy gives lower global prediction error. This means that an improvement in one measure cannot establish an overall policy advantage. The relevant question is whether the controller allocates imaging and spectroscopy in a way that serves the declared objective, compared with a simpler rule using the same acquired information. The settings chosen by the learned controller also remain close to the average of the available settings: the mean selected window side (about 16 pixels) and relative scan time (about 0.55) nearly equal the means over the action set (16.3 and 0.56; Supplementary Table S11). These averages show no strong shift in the mean settings, but do not by themselves establish a uniform selection distribution. The comparable outcomes identify the acquisition-rule baseline as a substantive operating point, rather than a weak control. Full baseline and alternative-profile comparisons are given in Supplementary Figures S5–S6 and Tables S5–S6.

Random candidate sampling provides broad coverage but can spend measurements far from a rare high-response region. Greedy sampling can find that region efficiently while leaving other structural environments poorly represented. Learning the operation policy adds a separate choice, when and how to image, but it does not remove this scientific trade-off. The present results show how those contributions can be evaluated separately; they do not establish convergence to a universal acquisition strategy.

### 5.3 Explicit imaging and spectroscopy histories

Figure 5 merges IMAGE and SPECTRUM into a single chronological lane for each predictor. Blue and red encode operation identity; the thin yellow–orange–red strip above each lane encodes the duration of its IMAGE events. Bar length represents actual acquisition time. This presentation distinguishes choosing a modality from choosing how long to spend on it, without using the same color scale for both meanings.

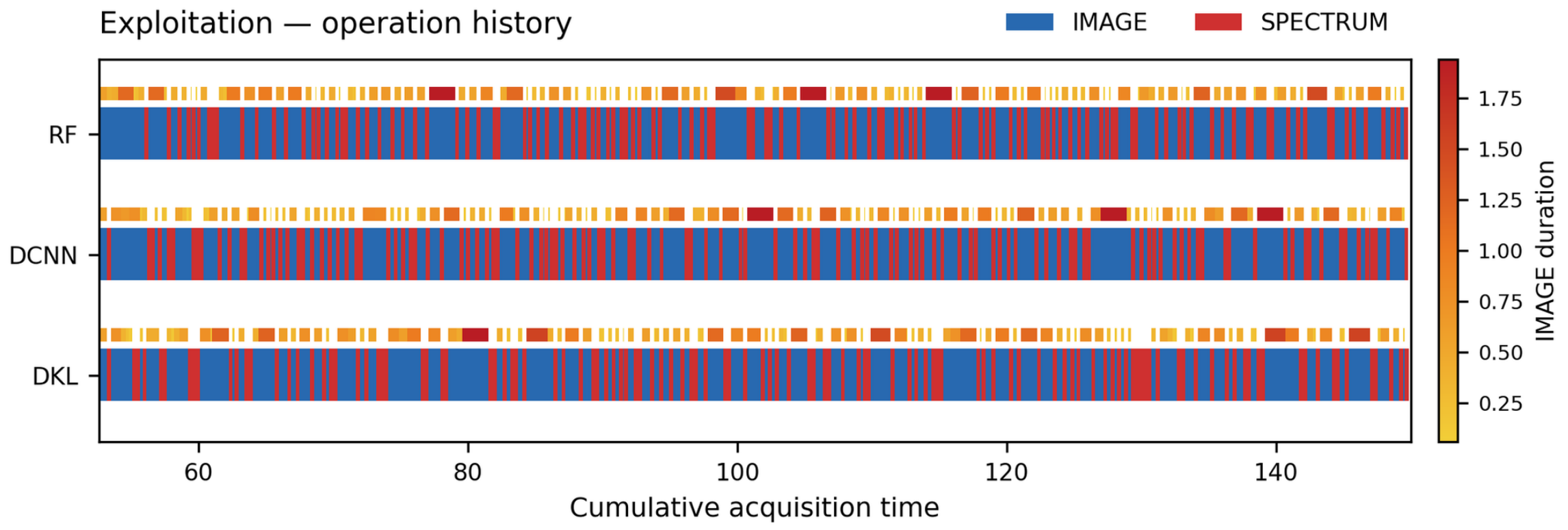

**(a)**

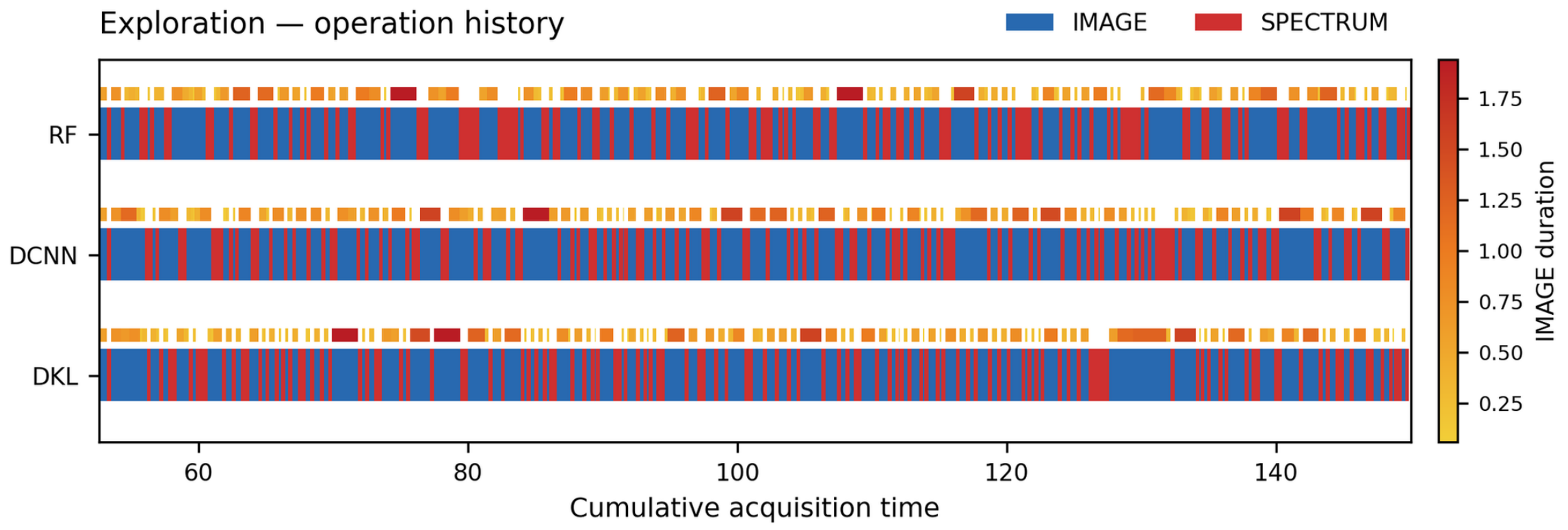

**(b)**

**Figure 5.** Merged acquisition histories for RF, DCNN, and DKL under (a) response discovery and (b) uncertainty-driven learning, using experimental initialization 1001 and policy seed 11. Each model has a single operation lane: IMAGE is blue and SPECTRUM is red, with bar width equal to the actual normalized duration. The narrow strip above each lane uses a yellow–orange–red scale for IMAGE duration, not modality or chronological order. The controlled interval begins after the common initialization. Successive images or spectra can occur independently; spectra are legal only at previously imaged, unlabelled centers. These are predeclared representative deployments, not trajectories chosen for their final outcomes.

The histories show that images and spectra are interleaved flexibly rather than acquired in fixed image–spectrum pairs. Images remaining without spectra at termination show the opportunity cost of acquiring structural information that is not subsequently labelled. Similar image and spectrum counts can still produce different outcomes because the selected locations and observation qualities differ. Acquisition counts and timing summaries are retained in the Supplementary Information.

### 5.4 Trajectories in control space and specimen space

Figure 6 follows one predeclared DKL deployment under each objective. The left panels show the chosen window and relative scan time as acquisition proceeds; the right panels place these decisions in the discrete control plane. Grid boundaries separate the available settings. Repeated selections are displaced within their own cell for visibility, without changing the underlying action. The bottom panels show measured spectroscopy locations on the original structural image. Jet colors encode cumulative measurement time throughout.

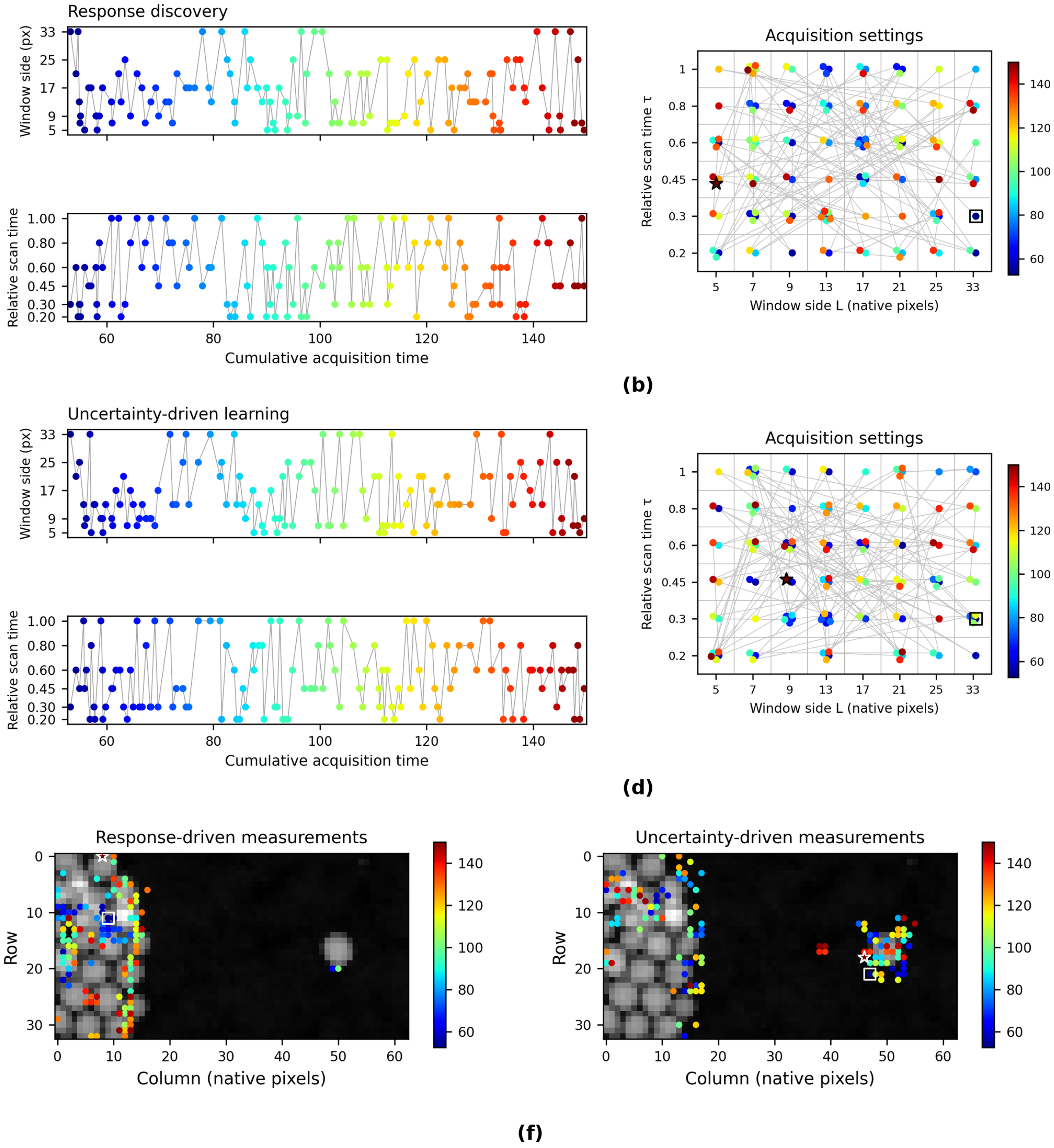

**Figure 6.** Imaging settings and measured locations for the predeclared DKL campaigns. (a,c) Window side and relative scan time versus cumulative acquisition time. (b,d) The same IMAGE actions in the window–scan-time plane; equal-sized categorical cells identify the discrete settings, and small deterministic offsets separate repeat visits within each cell. These offsets are display-only and do not introduce additional control values. (e,f) SPECTRUM locations on the structural image, shown as points without connecting paths. Jet colors represent cumulative acquisition or measurement time, not response. Open square and star symbols mark the first and last displayed actions. The compact-RBF profile, experimental initialization 1001, and policy seed 11 are unchanged.

In the displayed realization, response-driven sampling is concentrated in the high-response structural region, while uncertainty-driven sampling also visits the isolated feature on the other side of the image. These are examples, not universal spatial patterns. Their purpose is to distinguish the context and fidelity selected in control space from the locations labelled in specimen space.

The specimen panels show discrete measurements only; no beam path, intervening exposure, or stage-motion cost is implied.

The trajectories show continued changes in selected window size and scan time during the campaign. This is consistent with the mixed-quality, limited-data setting with the bounded policy-training schedule, and with the modest changes in mean settings noted in Section 5.2, but it does not identify a unique optimum. The complete three-model window, scan-time, implied blur, and duration histories are retained in Supplementary Figures S7–S10. The main-paper trajectory should therefore be read as an auditable example of the decision process, while the repeated-campaign results support the quantitative comparisons.

### 5.5 Internal reward is not global scientific performance

Figure 7 places cumulative exploitation and exploration rewards side by side on separate vertical scales. Under exploitation, the reward is an observed increase in the best spectral response, and its cumulative value is fixed by the normalized difference between final and initial records. Under exploration, the reward accumulates maturity-weighted changes in a model-specific uncertainty estimate. These curves should be interpreted together with Figure 4, not substituted for its discovery and prediction outcomes. A larger RF uncertainty reward than a DKL uncertainty reward does not imply a larger absolute scientific gain because the underlying variance constructions differ.

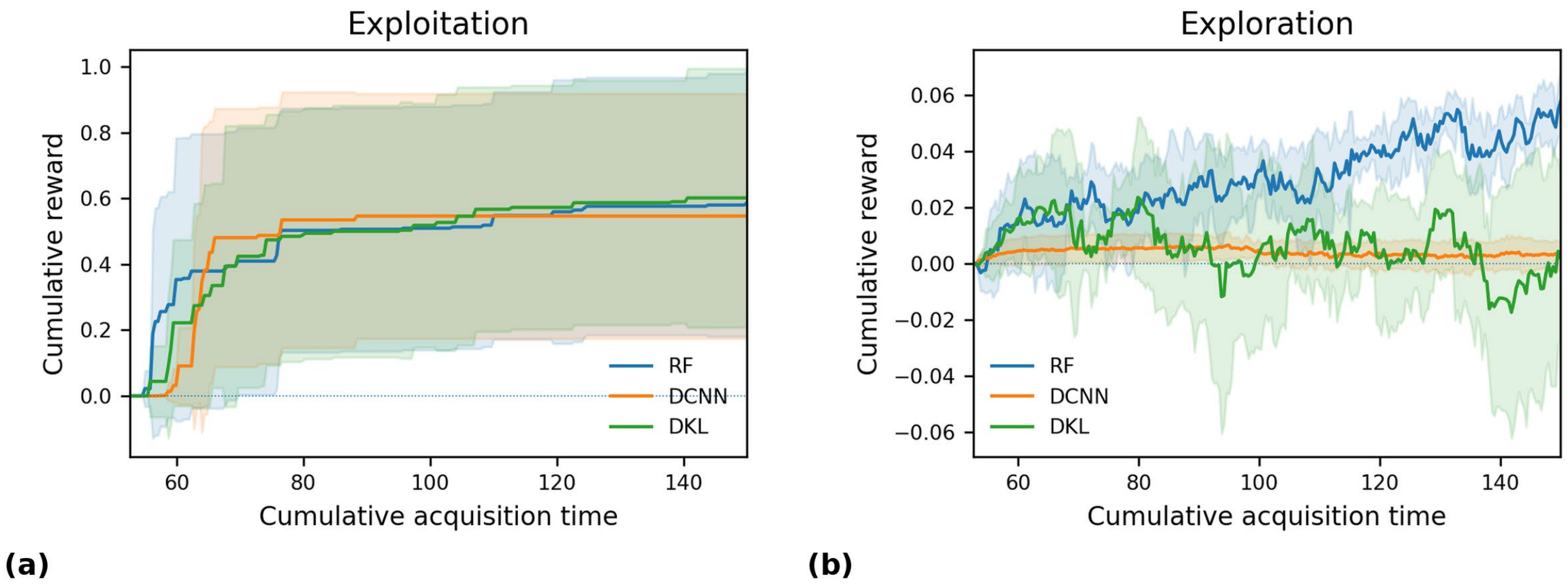

**Figure 7.** Accumulated scientific rewards, separated by objective. (a) Exploitation return: the increase in the best measured response divided by the fixed seed-response scale. (b) Exploration return: accumulated maturity-weighted decrease in normalized uncertainty on the fixed 32-image reference panel. IMAGE has zero immediate reward in both cases; negative exploration gains are retained. Lines and ribbons use the same paired six-initialization aggregation as Figure 4. The side-by-side panels have deliberately independent vertical scales. Uncertainty estimators have different numerical meanings, so larger exploration return across model families is not automatically greater scientific value. Independent discovery and prediction performance must be read from Figure 4 and the error audit.

The independent error analysis identifies where uncertainty rewards require further calibration. For RF and DKL, a decrease in internal uncertainty has little one-step association with improvement in held-out prediction error. For this DCNN implementation, the association is negative: confidence can increase while prediction becomes worse. This diagnostic separates a useful sampling signal from direct evidence of improved prediction, without ruling out longer-horizon benefits of uncertainty-guided selection. The correlations, failure fractions, instantaneous rewards, and both evaluation-panel diagnostics are given in Supplementary Section S7 and Figures S11–S12.

The distinction is especially important for DKL. Its GP variance is conditional on optimized neural weights and covariance parameters; it does not integrate uncertainty over those quantities. Relearning the representation after each spectrum changes the feature space in which the conditional variance is calculated. RF tree disagreement and DCNN head dropout have their own limitations. The present analysis does not prove that these uncertainty methods are generally ineffective, nor that the observed mismatch has a single cause. It demonstrates that uncertainty construction and reward weighting are separate design problems. Maturity can attenuate early model-derived rewards, but cannot repair incorrect action rankings or guarantee positive correlation with error reduction. These qualifications are consistent with the literature on learned kernels and uncertainty estimation [55–57].

### 5.6 Sensitivity to the instrument response model

The principal scenario places a direct time cost on a larger image window. A complementary scenario instead keeps the total image cost independent of window size and increases blur with the speed required to cover a larger raster in that time. All other primary settings are held fixed, and the policies are retrained under the changed observation law. This is a controlled test of the assumed instrument model, not a second experimental calibration.

The alternative analysis gives mixed changes in discovery and prediction rather than a general improvement. It reinforces the need to measure the actual relation between scan settings, image quality, and elapsed time before selecting a policy for a microscope. The full formulation, numerical results, and paired comparisons are in Supplementary Section S8 and Figures S13–S23; the common reference-window setting provides an exact implementation check.

## 6. Drift-induced structural misregistration and closed-loop acquisition control

The first campaign treated image and spectrum positions as perfectly registered, isolating the choice of structural context and scan fidelity. In the second campaign, we ask how the same scientific objectives change when the instrument drifts during and between operations. This campaign is separate from the first: the structural image and measured responses are again real, whereas the spatial drift, blur, cross-correlation observations, action costs, and control decisions are simulated. The window is fixed at 17 × 17 pixels so that changing context does not obscure the influence of scan duration, registration, and spectroscopy throughput. The two objectives remain discovery of the largest measured band response and learning the image–spectrum relationship. Studies of scan noise, drift distortion, and image registration provide the relevant physical context [63–67].

### 6.1 Drift as motion through structural and predictive representations

A measurement intended at position $\mathbf{r}$ is associated with the patch centered at $\mathbf{r}$ only while that position remains registered. A displacement $\mathbf{d}$ replaces its structural input with a patch centered at $\mathbf{r} + \mathbf{d}$, while a later spectrum may be acquired at yet another position. When a spectrum keeps the nominal target, this produces feature–label misregistration rather than ordinary image noise. Because the field is fully acquired, the effect can be visualized without any controller (Figure 8a–c): drifted patches from the experimental image are passed through a frozen two-dimensional autoencoder and compared with their undrifted representations. Each reference point keeps its original measured EELS-band response as its color (Figure 8b,c). The resulting latent trajectories spread anisotropically and increasingly overlap structural environments associated with different responses (Figure 8). The overall spreading grows substantially with displacement, whereas neighborhood-based color mixing is less regular, because the nearest-neighbor graph changes as points move. These observations identify spatially varying sensitivity to drift. The autoencoder provides a convenient nonlinear low-dimensional representation [68], but distances in the two-dimensional embedding are not a calibrated measure of information loss.

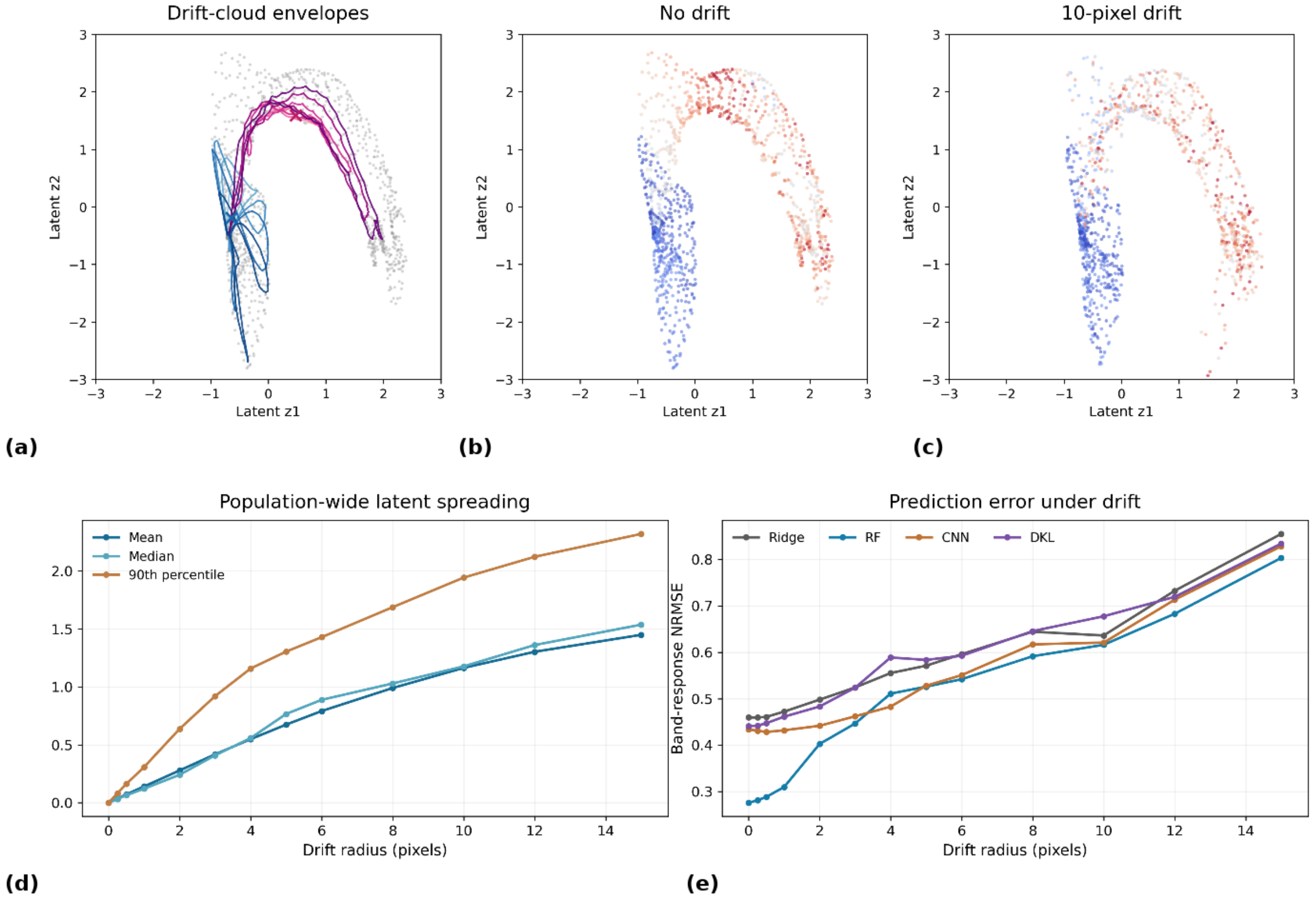

**Figure 8.** Static displacement of structural image patches and consequences for EELS learning. (a–c) Square panels with common latent-coordinate limits of −3 to 3 on both axes: two-site drift-cloud envelopes on the clean background (a), and response-coded distributions without drift (b) and after 10-pixel displacement (c). In (b,c), blue-to-red colors encode increasing measured band intensity at the nominal center and remain fixed under drift; their intermixing shows feature–label misregistration in the frozen latent space. (d,e) Population latent spreading and band-response prediction NRMSE for Ridge, RF, CNN, and DKL. Labels remain assigned to nominal centers. Additional diagnostics are in Figures S24–S26.

Because the clean-trained regressor receives a different structural patch after drift, its prediction can deteriorate even if the EELS spectrum itself is unchanged. Figure 8d shows increasing latent spreading, while Figure 8e and Supplementary Figure S26 show degradation of band-area and downsampled full-spectrum prediction for Ridge, RF, CNN, and DKL. The details depend on the model: RF is the most accurate clean-data regressor in the tested implementation, while the gap to CNN narrows at large misregistration. Stronger drift therefore modifies not only the apparent location of structural states but also the reliability of the structure–property mapping; the full model-by-model curves and population-spreading statistics are in Supplementary Section S10.1.

## 6.2 Imaging speed, blur, and the value of another measurement

We couple drift to the same directional scan-blur law used in the first campaign. Fast images preserve acquisition time but have a wider effective blur kernel; slow images preserve higher spatial-frequency information but allow additional instrument motion before the next spectrum. The first comparison therefore retains a simple IMAGE–SPECTRUM protocol: the controller chooses the scan duration, acquires one image, and then takes one EELS spectrum at the nominated target. Reimaging heuristics add the option to take a second image before the spectrum. These are controlled baselines; no cross-correlation correction is required for this scan-speed-only comparison. Across the tested drift speeds, rapid imaging typically improves the rate of discovering strong responses because it permits more spectra, whereas an intermediate duration can favor prediction quality by reducing blur. The balance is objective-dependent, and a fixed speed is a scientifically meaningful benchmark rather than an inadequate control.

### 6.3 From imaging to explicit calibration and sequential decisions

We then separate the three operations IMAGE, CALIBRATE, and SPECTRUM. IMAGE has six allowable durations under the same blur law; CALIBRATE registers the latest acquired image against a structural reference by normalized cross-correlation and applies an estimated command correction; SPECTRUM acquires one measured EELS response and completes the current target. The maximum image duration is three normalized time units, while calibration and spectroscopy each cost one unit. A position correction invalidates the previous image–spectrum association until a fresh image is acquired. In this experiment, cross-correlation is performed on the actual simulated blurred observation, so a faster scan can make calibration less reliable. An unregistered image is not assumed to provide a correct drift estimate. Subpixel and non-rigid registration address different transformation models [65,69]; the present experiment retains its stated cross-correlation procedure.

Three decision architectures are compared under the same physical action set. Model-free Q-learning selects operations without an explicit drift-dynamics filter. A generic Kalman filter (K1) propagates a stochastic constant-velocity state; a structure-aware Kalman model (K2) additionally represents candidate periodic components. Both filters use the same type of cross-correlation observation, while their decision rules and matched periodic schedules are analyzed separately. Fixed-fast imaging and periodic calibration are retained as strong reference protocols. The hidden oracle records true drift, actual image–spectrum mismatch, measured response, and held-out scientific error, but the controller is allowed to use only acquired images, spectra, registration signals, model uncertainty, and its remaining budget. This separation assesses three contributions independently: instrument-state prediction, alignment of the online reward with scientific progress, and the value of the resulting decisions. Q-learning [70] selects actions, whereas Kalman filtering [71] estimates dynamical state from observations; the workflow specifies how each model's information is used to choose or execute an action.

The two operational objectives preserve the reward logic of Section 4. Response discovery rewards an increase in the best actually measured spectral-band response, with no immediate reward for imaging or calibration. Structure–property learning rewards maturity-weighted reduction in RF inter-tree uncertainty on a fixed reference bank. Image and calibration actions acquire their value through later spectra. The oracle independently monitors actual prediction error, the fraction of accessible response improvement realized, residual registration error, and experimental throughput. The full 300-unit campaigns are summarized in Figure 9; reward equations, controller-observable state, and complete comparisons are given in Supplementary Sections S10.2–S10.5.

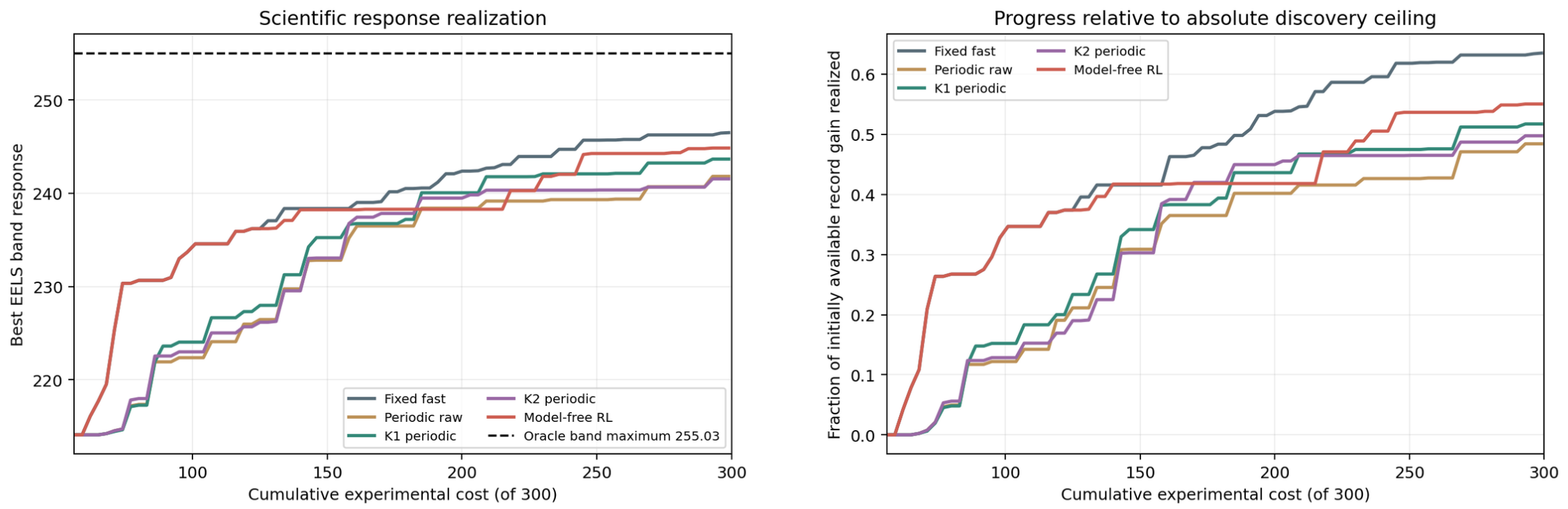

**(a)** **(b)**

**Figure 9.** Discovery performance referenced to the complete EELS response field. (a) the best band response measured as the campaign budget is consumed by each controller. (b) the fraction of improvement beyond the initial measured record that has been realized relative to the highest accessible experimental response. Curves are reconstructed from the completed 300-unit campaign traces. This accessible ceiling is an evaluator-only reference, not an online reward or a guaranteed attainable budget-limited optimum.

### 6.4 Realized control sequences and physical-state accuracy

The main operational difference between controllers is the allocation of time among imaging, calibration, and spectroscopy. Fixed-fast imaging progresses through many targets with no calibration overhead. Periodic and Kalman-based protocols spend time checking and correcting position, typically requiring a subsequent image before spectroscopy. Model-free RL can learn a mixed schedule; its chosen sequence depends on whether the objective is discovery or uncertainty reduction. In the recorded campaigns the model-free RL controller does use CALIBRATE: its calibration actions occur intermittently, rather than on the regular schedule of K1 and K2. They can be difficult to discern when hundreds of operations are compressed into one timeline. There is no separately trained RL–Kalman hybrid in this comparison. Figure 10 displays representative, predeclared action histories and shows why a more accurate physical controller can complete fewer spectroscopy measurements. These trajectories illustrate specific held-out drift histories rather than an optimized best-case realization. The zoomed lower panel of Figure 10 makes the model-free RL calibration operations visible: they are actual selected actions, rather than inferred corrections. The lower panel uses the same exploitation policy, drift regime and realization as the upper timeline.

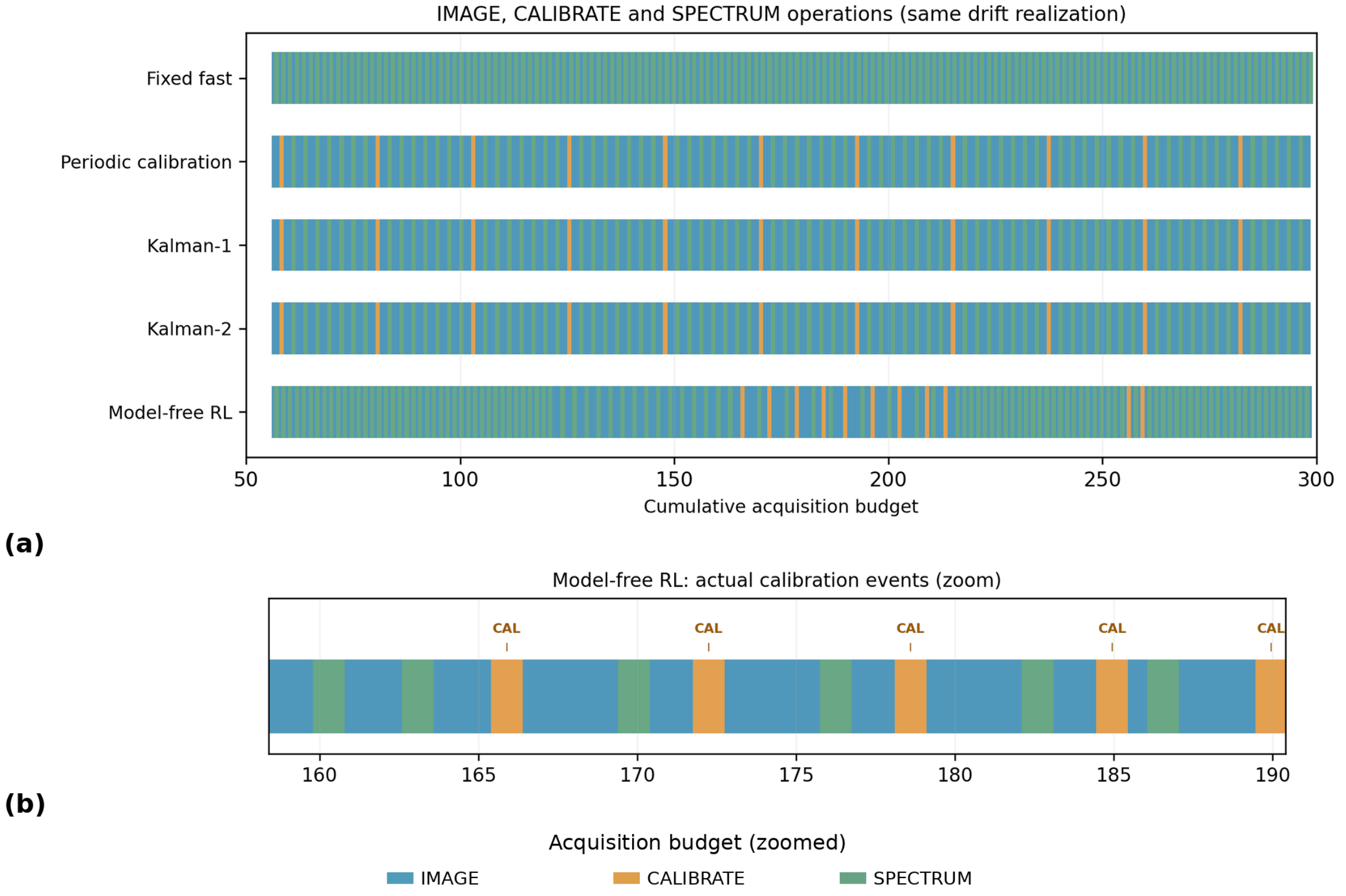

Figure 10. Recorded imaging, calibration, and spectroscopy actions under a common periodic-drift realization (exploitation objective, seed 500). (a) complete chronological sequences for fixed-fast, periodic calibration, Kalman-1, Kalman-2, and model-free RL; bar width represents action duration. (b) an expanded segment of the same model-free RL trace centered on its first CALIBRATE event, with each calibration explicitly annotated. Blue denotes IMAGE, orange CALIBRATE, and green SPECTRUM. The RL controller performs calibration, but only intermittently rather than at a fixed interval; the zoom avoids hiding these short events on the full-campaign time axis.

Position accuracy improves substantially with periodic compensation, and K2 can reduce the residual error relative to K1 under a matched calibration schedule. The framework quantifies the scientific cost–benefit of that improvement: time used for registration reduces the number of EELS measurements, and uncertainty reduction is an imperfect surrogate for true prediction-error reduction. The representative positional-error and relative-mismatch trajectories in Figure 11 demonstrate these distinct metrics. In the long campaigns, model-free RL approaches the scientific performance of periodic calibration, while the paired controls establish a practical reference for judging the added value of learned scheduling.

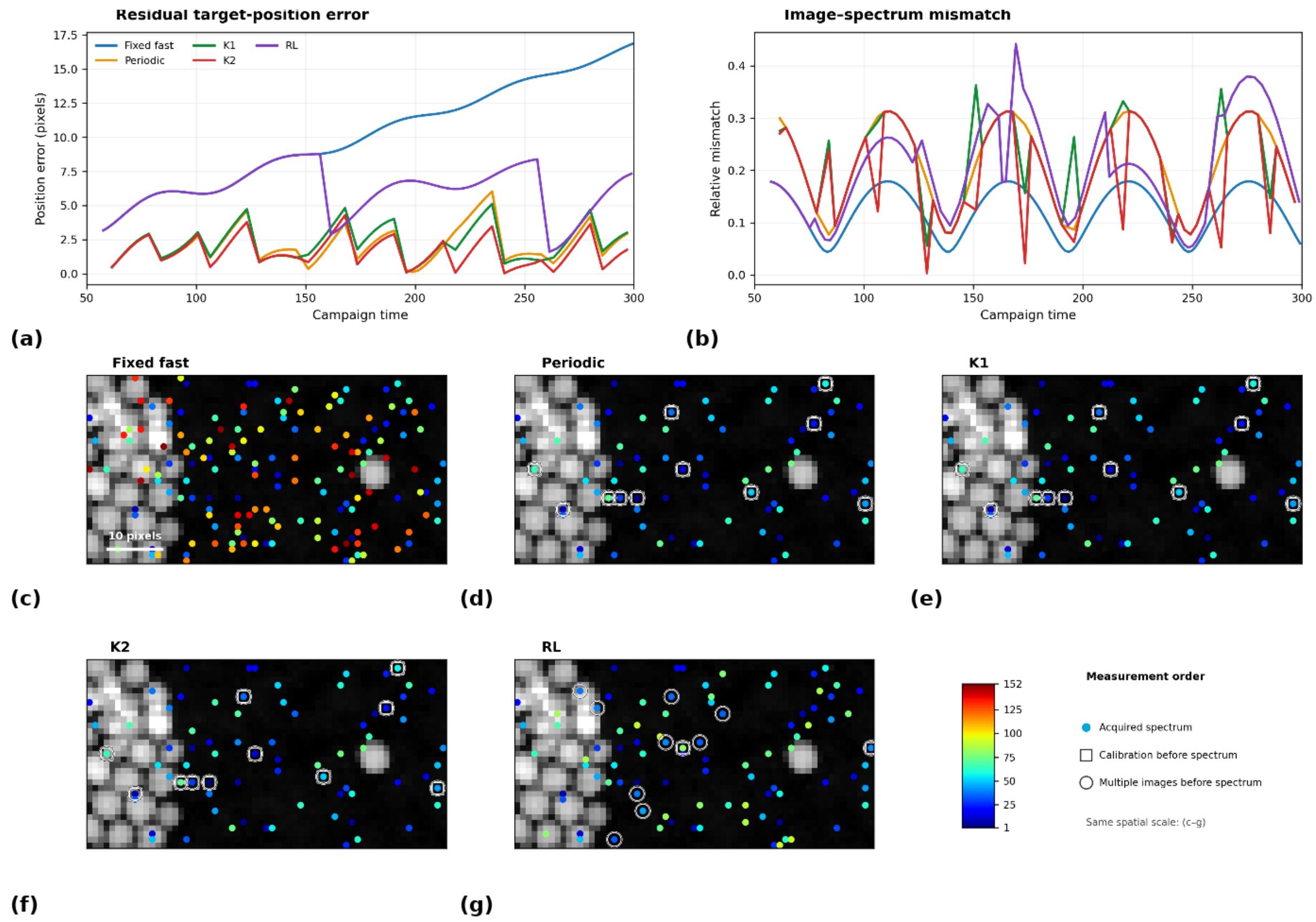

**Figure 11.** Physical fidelity and measured locations for the 300-unit exploration campaign under the same periodic-drift realization. (a,b) Equal-sized evaluator-only residual target-position error and relative image–spectrum mismatch. (c–e, second row) Fixed fast, Periodic, and K1. (f,g, third row) K2 and RL, with shared keys in the remaining cell. Spatial panels have identical coordinates; the 10-native-pixel scale bar in c applies to c–g. Jet color denotes measurement order. Squares indicate calibration before spectroscopy; circles indicate more than one image.

Together, the results of the second campaign show that drift can be characterized in feature and latent space before controller training, that scan speed controls a measurable fidelity–throughput compromise, and that explicit registration reduces physical error at a real acquisition cost. They also distinguish a full-field discovery ceiling from the more limited attainable outcome of a finite-budget controller. The comparison establishes a reproducible way to select and assess calibration strategies by both physical fidelity and scientific outcome, with strong baselines defining the benefit required of additional model complexity.

## 7. What non-myopic workflows need from the microscope

Building the non-myopic workflows required more than defining a state, a reward, a predictor, and an acquisition function. The controller had to know which actions it could take, what each action cost, how a chosen scan setting turns the specimen into an image, whether an image and a spectrum refer to the same location, how the spectral reward is calculated, how long each step actually takes, and what has already happened in the session. In our benchmark, all of this information was supplied by the simulator: legal actions and coordinates, an assumed observation law, assumed costs, exact or simulated registration, a fixed spectral target, and a complete record of every action. On a real microscope, each item must be provided by the instrument or measured by the facility. Below we list these items in the order in which the workflow uses them, and for each we indicate what the manufacturer is best placed to provide.

First, the controller needs an action interface. Imaging and spectroscopy must be callable independently, with explicit coordinates, field of view, pixel count, pixel size, scan direction, dwell time, and detector settings. The instrument should report the settings that were actually executed, not only those requested, and should signal whether an acquisition was completed, cancelled, timed out, or only partly acquired. Limits on window size, scan speed, dose, and allowed operating modes should be exposed as explicit constraints, so that invalid actions can be excluded before execution, as the action mask did here. Automated aberration correction already relies on this kind of access to the corrector and microscope controls [15,16].

Second, the controller needs to know how a chosen scan becomes an image. In the first campaign, this was a single Gaussian blur law linking scan time to image sharpness, and Section 5.6 showed that changing this assumed law changes the outcome. A real observation model includes the scan-generator and detector response, the directional point-spread function, line settling, flyback, hysteresis, and the dependence on dwell time and signal level. Blur, noise, and geometric distortion are different effects and should not be collapsed into one number [63–65]. Manufacturers can provide the command semantics, hardware bandwidths, detector calibration, and timing specifications; the facility then needs to measure the effective response under its own alignment and operating conditions.

Third, the controller needs to know whether an image and a subsequent spectrum refer to the same place. The first campaign assumed perfect registration, and the second simulated drift and cross-correlation registration. On a live instrument, this requires timestamped transformations between stage, scan, and detector coordinates, together with alignment residuals and estimates of drift during and between acquisitions. Drift depends on elapsed time, scan direction, recent stage motion, temperature, and sample mounting, so a specification sheet is not enough: the instrument should stream its own position and environmental telemetry, and the workflow should be able to check it against reference images of the specimen. Time-stamped estimates should include their uncertainty and prediction horizon, so the controller can decide when a spectrum remains paired with its descriptor, when registration is worth its cost, and when a loss-of-registration fallback is required.

Fourth, the reward must be reproducible. Both campaigns used the integrated band of a stored spectrum as the scientific signal. On a live instrument, energy-axis drift, gain, dark current, background subtraction, saturation, exposure time, and beam current all change the number assigned to the same material [46]. The controller therefore needs the calibration state and the processing steps used to compute its reward, together with quality flags and the raw spectrum. Repeated measurements would allow measurement noise to be separated from uncertainty in the structure–property model; repeated spectroscopy and additional measurement noise were not included here.

Fifth, the cost model must reflect real time. Our budgets counted only acquisition time in normalized units. A real session also spends time moving and settling the stage, switching detectors, reading out and transferring data, refitting the model, and scoring candidates, and some of these steps can run in parallel. Timestamps and completion signals for every step establish the real schedule; offline calibration and policy-training costs should also be recorded separately so their reuse across campaigns can be assessed. Beam blanking, safe interruption, and a deterministic fallback should remain within the instrument's own control layer. Cumulative dose and specimen changes must also be recorded, because a measurement that alters the specimen cannot be evaluated as if the response map were unchanged.

Finally, every observation should be linked to the command that produced it, the calibration state, the specimen, the model version, the candidate scores, the selected action, and the outcome. Failed and unsaved operations also need to be recorded, because they consume time and influence the next decision [12]. Calibrations should carry validity ranges and dates, so that outdated prior knowledge can be detected. Data stewardship principles and acquisition frameworks provide a foundation for these links [13,14]. In short, manufacturers can provide reliable control, telemetry, and characterized instrument responses, whereas facilities and users must establish sample-dependent behavior and define the scientific objective. Neither the instrument nor the predictor can supply a universally appropriate reward.

## 8. Conclusions

Autonomous structure–property discovery is a coupled information-acquisition problem, not only a search over fixed image patches. Imaging, spectroscopy, image size, image fidelity, instrument stability, and computation all determine what can be learned within a finite budget. We explored two separate campaigns on the same real-data EELS benchmark. The first allowed the controller to choose between imaging and spectroscopy and to select image size and scan speed under perfect registration; the second fixed the image size but added drift, so that the controller also had to decide when to recalibrate. Pre-experiment analysis of the same data identified economical predictors, their data requirements, and full-data reference errors, and these resources were kept separate from information that is unavailable during acquisition.

The framework identifies objective-dependent choices rather than imposing a single model or controller ranking. DKL gave the best full-data reference and frequently found the maximum response, whereas RF gave the lowest common-input prediction error among the learned policies. Acquisition-based and fixed alternating controls define useful practical baselines, while learned policies provide an explicit mechanism for allocating modality and observation fidelity. The independent reward audit identifies where uncertainty calibration is needed, and distinguishes empirical learner maturity from the reliability of individual decisions. In the second campaign, explicit registration substantially improves positional fidelity, and the action records quantify its cost in spectroscopy throughput. Fast imaging and periodic calibration provide complementary operating points for discovery and learning, against which more elaborate control can be evaluated.

The present demonstration is limited to a small specimen, overlapping training and test neighborhoods, reflected window boundaries, a descriptor that couples physical context to the number of active input cells, and assumed acquisition and drift laws. We deliberately did not combine variable image size with drift in a single workflow. This is not technically difficult, but the result would be meaningful only if the scan-speed response, the drift behavior, and their interaction were calibrated on a specific instrument; with assumed laws, a combined workflow would mainly reflect the assumptions. The resulting contribution is a general, testable workflow framework that keeps experimental objectives, acquisition scores, rewards, computational feasibility, and independent scientific performance separate, together with a list of the instrument information needed to deploy it.

## AI use statement

Sergei V. Kalinin conceived and directed the study, defined the scientific objectives, physical assumptions, and reward functions, guided the analyses, interpreted the results, and revised the manuscript. ChatGPT (OpenAI) and associated tools assisted with methodological development, code generation and debugging, numerical execution and analysis, figure and schematic preparation, literature searches, and manuscript drafting and editing. Experimental images and spectra originated from the cited dataset. The author reviewed and revised the AI-assisted outputs and retains responsibility for the manuscript's scientific accuracy and integrity.

## Data and code availability

The reproducibility archive contains the reconstructed data cache, source code for all three predictors and both campaigns, calibration results, action histories, trained policy and predictor checkpoints, and the analysis notebooks. All manuscript figures are generated from this archived numerical record. Supplementary Section S9 documents provenance and verification. The archive is available at https://github.com/SergeiVKalinin/RL_and_Rewards/tree/main/RL_EELS .

**Acknowledgements**

This work was supported by the U.S. National Science Foundation through the Programmable Cloud Labs program, PCL-Test Bed: ATHENA - Advanced Testbed for High-throughput Experimentation in Nano- and Atomic Science (Award No. 2607469). Any opinions, findings, and conclusions or recommendations expressed in this material are those of the authors and do not necessarily reflect the views of the National Science Foundation.

## Supplementary Information

This Supplementary Information follows the structure of the main text. Sections S1–S9 support the first campaign (main Sections 2–5): the data and observation model (S1), the predictor implementations (S2), pre-experiment calibration (S3), the sequential controller and baselines (S4), complete performance comparisons (S5), additional acquisition diagnostics (S6), reward and uncertainty audits (S7), the alternative instrument-response scenario (S8), and verification (S9). Section S10 supports the second, drift campaign (main Section 6). All results are taken from the archived workflows; no predictor or policy was retrained to prepare this document. Figure and equation numbers without the prefix S refer to the main text, and references are numbered as in the main text.

### S1. Data provenance, scalarization, and observation model

This section describes the dataset shared by both campaigns, the definition of the scalar target, the data partition, and the assumed laws that convert the static data into a controllable acquisition environment for the first campaign (main Section 2).

### S1.1 Experimental records and split

The data are dataset 2 of the plasmonic EELS archive of ref. [20]. The cache contains 1,663 original training-pool records and 300 held-out records, each with a raw 17 × 17 structural patch and 101 targets: the original integrated band area followed by a 100-channel spectral representation. The original 33 × 63 structural image was reconstructed from overlapping patches with zero overlap disagreement and zero patch-reconstruction error. Every native image pixel is supported by at least 75 unpadded patch observations. This operation recovers an observed scene; it does not generate new specimen content. Spectroscopic responses are absent at 116 of the 2,079 image centers, and those centers remain ineligible.

The scalar response is retained directly from the source notebook, whose integration interval is 0.52–0.70 eV. The 100-channel representation spans approximately 0.20–3.7 eV. The scalar is not recomputed from that downsampled spectrum. No physical assignment to an individual plasmon mode is introduced. The first-campaign sequential experiments predict the scalar band area; the cached full spectra are preserved but are not claimed as new sequence-level regression outputs for that campaign.

Main Figure 1c retains the 100-channel spectra nearest the 10th, 50th, and 90th percentiles of stored area: row 30, column 54; row 23, column 31; and row 17, column 16. Both energy intervals are indicated in eV. The additional Figure 1d map uses linear interpolation at 0.8 and 1.0 eV and trapezoidal integration between these endpoints. This is an approximate integral of the cached spectral sampling for display, not a change to the original stored target or the campaign calculations. For display only, missing pixels in the two Figure 1 response maps are replaced by the mean of available neighboring values; the missing-label mask, eligible centers, and all fitted results remain unchanged.

A seed-101 partition removes 120 calibration records from the original pool, leaving 1,543 online candidates. The same 300 test responses are retained throughout. The fixed uncertainty survey comprises 32 clean, 17-pixel calibration images; their labels are not revealed to the online controller. Twelve initial seed pairs are selected independently for each experimental initialization. Spatially adjacent train/test centers and overlapping windows are intentionally retained, so the evaluation concerns within-specimen interpolation, not generalization to an independent specimen.

### S1.2 What is visible before a detailed image

Each unseen candidate is represented by a declared coarse preview: a 9-pixel crop with directional blur $\sigma = 3.2$, pooled to 3 × 3 and expanded into the common input canvas. A 5-unit overview cost is charged. This preview permits image-conditioned, non-random candidate ranking without revealing the eventual detailed observation. In addition, the 32 fixed reference images cost 32 units, and the twelve seed image–spectrum pairs cost 12 × (1 + 0.30). Consequently, the INITIAL record ends at 52.6 acquisition-time units, leaving 97.4 units for controlled actions within the total budget of 150. These costs are identical between paired methods. Offline calibration is treated as amortized prior work outside that campaign budget.

### S1.3 Linear-size imaging cost and directional blur

The first campaign uses eight window sides $L \in \{5, 7, 9, 13, 17, 21, 25, 33\}$ and six relative scan times $\tau \in \{1, 0.8, 0.6, 0.45, 0.3, 0.2\}$. The image cost is $(L/17)\tau$, and the spectrum cost is 0.30. The assumed blur width is $\sigma(\tau) = [(1/\tau - 1)/0.12]^{1/2}$. Both scan-time factors and costs are normalized simulation quantities, not calibrated microscope seconds. Figure S1 presents the two laws side by side.

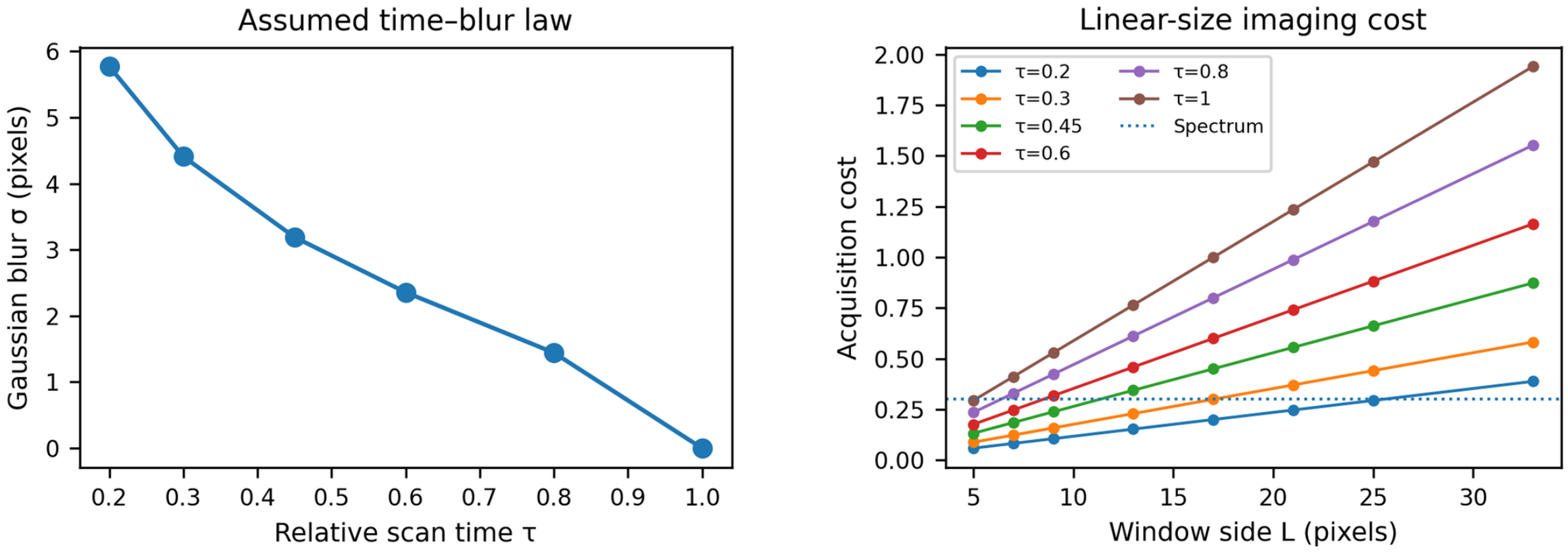

**(a)** **(b)**

**Figure S1.** (a) The deterministic relationship between relative scan time and directional Gaussian blur. This is the assumed instrument observation law used throughout the benchmark. (b) Imaging cost versus window side for every scan-time setting, together with the fixed cost of one spectrum. The plot shows the intended throughput trade-off; it does not include CPU training seconds.

The source scene is reflect-padded at its boundary. A window is then cropped, blurred along its columns with reflective crop boundaries, centered in a 33 × 33 zero-padded canvas, and adaptively pooled to the regressor's input grid. Blur is therefore expressed in native pixels before pooling. The fixed canvas preserves the acquired support, but couples physical context to the number of active numerical cells. RF_fast's 5 × 5 descriptor has only one active cell for $L = 5$, nine for $L = 17$, and 25 for $L = 33$. DKL's compact 7 × 7 grid has 9, 25, and 49 active cells at the same window sizes. Larger-window improvements consequently cannot be attributed exclusively to physical context.

Reflected boundary content does not constitute a larger independent specimen. Neither drift nor additional measurement noise, dose damage, line flyback, or scan-dependent spectral noise is synthesized. Images and spectra are never reacquired at the same center in the implemented environment.

## S2. Predictor architectures and training

This section specifies the three predictor families compared in main Section 3.1 and the linear baseline used in static screening. All predictors share the same input conventions and are refitted from scratch after every new spectrum.

### S2.1 Shared conventions

Each supervised refit uses all acquired pairs and only those pairs. A spectrum triggers a from-scratch fit with fixed online initialization seed 127; an image alone does not update any predictor parameters. Image/target normalization is fitted from the currently acquired training data, not hidden evaluation targets. All predictor implementations expose mean prediction, an uncertainty quantity, and measured fitting/inference CPU and elapsed times. Reusing the accumulated supervised dataset is distinct from experience replay in the RL controller, which is absent.

The profiles below are an explicit finite design set. They are not an exhaustive search over model architectures or hyperparameters. Timing is retained separately from the acquisition budget. One worker thread is used per predictor fit to keep the recorded CPU comparisons interpretable.

### S2.2 Random forest

The pooled canvas is flattened and fitted with bootstrap regression trees. The predictive mean is the tree-average response, and the uncertainty proxy is the population variance across tree predictions. No additional probabilistic calibration layer is fitted. Tree count, input dimension, minimum leaf size, and feature subsampling are varied as shown in Table S1a. The implemented tree variance should not be confused with inferential forest uncertainty [51,58].

**Table S1a. Random-forest profiles.**

| Profile | Input grid / dimension | Trees | Min. leaf | Feature fraction |
|---|---|---|---|---|
| RF_fast | 5 × 5 / 25 | 48 | 2 | 0.8 |
| RF_balanced | 7 × 7 / 49 | 128 | 2 | 0.8 |
| RF_detail | 9 × 9 / 81 | 128 | 1 | 1.0 |
| RF_smooth | 7 × 7 / 49 | 128 | 4 | 0.7 |
| RF_dense | 9 × 9 / 81 | 256 | 2 | 1.0 |

### S2.3 Deep convolutional neural network

The DCNN contains a 3 × 3 convolution with $w$ channels, ReLU, 2 × 2 average pooling, a second 3 × 3 convolution with $2w$ channels, ReLU, and adaptive average pooling to 3 × 3. The resulting $18w$ features feed a 48-unit ReLU layer, head dropout, and a scalar linear output. AdamW minimizes squared error in minibatches of 64. There is no batch normalization. The deterministic prediction disables dropout; repeated draws with dropout active in the head estimate the variance proxy. Thus this implementation samples uncertainty in the head, not a posterior over the entire network.

Training length, width, input resolution, and dropout sampling differ across the three tested profiles. These choices are held fixed once a profile is selected for a campaign.

**Table S1b. DCNN profiles.**

| Profile | Input / channels | Epochs | Dropout / draws | Learning rate / decay |
|---|---|---|---|---|
| DCNN_compact | 17 × 17 / 12→24 | 16 | 0.18 / 20 | 0.003 / 0.003 |
| DCNN_medium | 25 × 25 / 16→32 | 24 | 0.18 / 24 | 0.002 / 0.003 |
| DCNN_wide | 33 × 33 / 24→48 | 32 | 0.25 / 24 | 0.002 / 0.005 |

### S2.4 Deep-kernel learning

The DKL model maps flattened, pooled image descriptors through a multilayer perceptron with SiLU activations into a low-dimensional latent space. A Gaussian process with an automatic-relevance-determination RBF or Matérn-3/2 covariance acts on this learned representation. Encoder weights, kernel length scales, signal amplitude, and observation-noise variance are jointly optimized using the Gaussian log marginal likelihood [42]. This is supervised representation learning from acquired image–response pairs; it is not image-only manifold learning.

Exact conditional GP algebra is performed in float64. Length scales are constrained to 0.03–30, signal amplitude to 0.001–100, and standardized noise variance to $10^{-4}$–2. Initial length scales are 0.5, amplitude is 1, and noise variance is 0.05. Gradient norm is clipped at 10. Cholesky factorization uses jitter escalation from $10^{-8}$ to $10^{-3}$ as necessary, and the best recorded likelihood state is retained. Compact profiles use hidden layers 32 and 16 with a four-dimensional latent representation; the detailed profile uses 64 and 32 with six latent coordinates. All 1,543 candidates are retained in full-data fits: there is no sample cap, inducing-point substitution, or retrospective access to unmeasured online targets.

The acquisition signal is latent-function variance, conditional on the fitted neural weights and GP parameters. The final coverage audit also adds fitted observation noise. Exact conditional inference does not integrate parameter uncertainty and does not imply calibrated uncertainty over the complete learned model. Standard conditional GP inference [57] and learned-kernel limitations [56] apply to this distinction.

**Table S1c. DKL profiles.**

| Profile | Input / hidden layers / latent | Kernel | Epochs | Learning rate / decay |
|---|---|---|---|---|
| DKL_RBF_compact | 49 / 32,16 / 4 | ARD RBF | 40 | 0.010 / 0.0001 |
| DKL_Matern_compact | 49 / 32,16 / 4 | ARD Matérn-3/2 | 40 | 0.010 / 0.0001 |
| DKL_RBF_detail | 81 / 64,32 / 6 | ARD RBF | 60 | 0.008 / 0.0001 |

### S2.5 Linear control

Ridge regression supplies a static baseline using standardized grids of 5 × 5, 9 × 9, and 13 × 13, with regularization strengths 10, 10, and 100. It appears in the $N$ = 384 screening and Pareto analysis. It is not assigned an artificial uncertainty estimate and is not used as a sequential exploration regressor.

## S3. Pre-experiment calibration and empirical references

This section describes how the predictor profiles were screened, how the full-data reference errors were obtained, and how the maturity parameters used in the exploration reward were calibrated (main Sections 3.2–3.3).

### S3.1 Profile screening, global references, and Pareto construction

At $N$ = 384, fourteen profiles are fitted at each of the 48 image settings using three paired subset/fit repetitions, giving 2,016 fits. For each profile, its best calibration error on the common clean-L17 panel is found. Within each model family, the fastest profile within 12% of the family's best such error becomes the primary online profile: RF_fast, DCNN_compact, or DKL_RBF_compact. The 12% tolerance is an engineering choice, not a statistically inferred threshold. RF_detail, DCNN_wide, and DKL_Matern_compact are retained as alternatives.

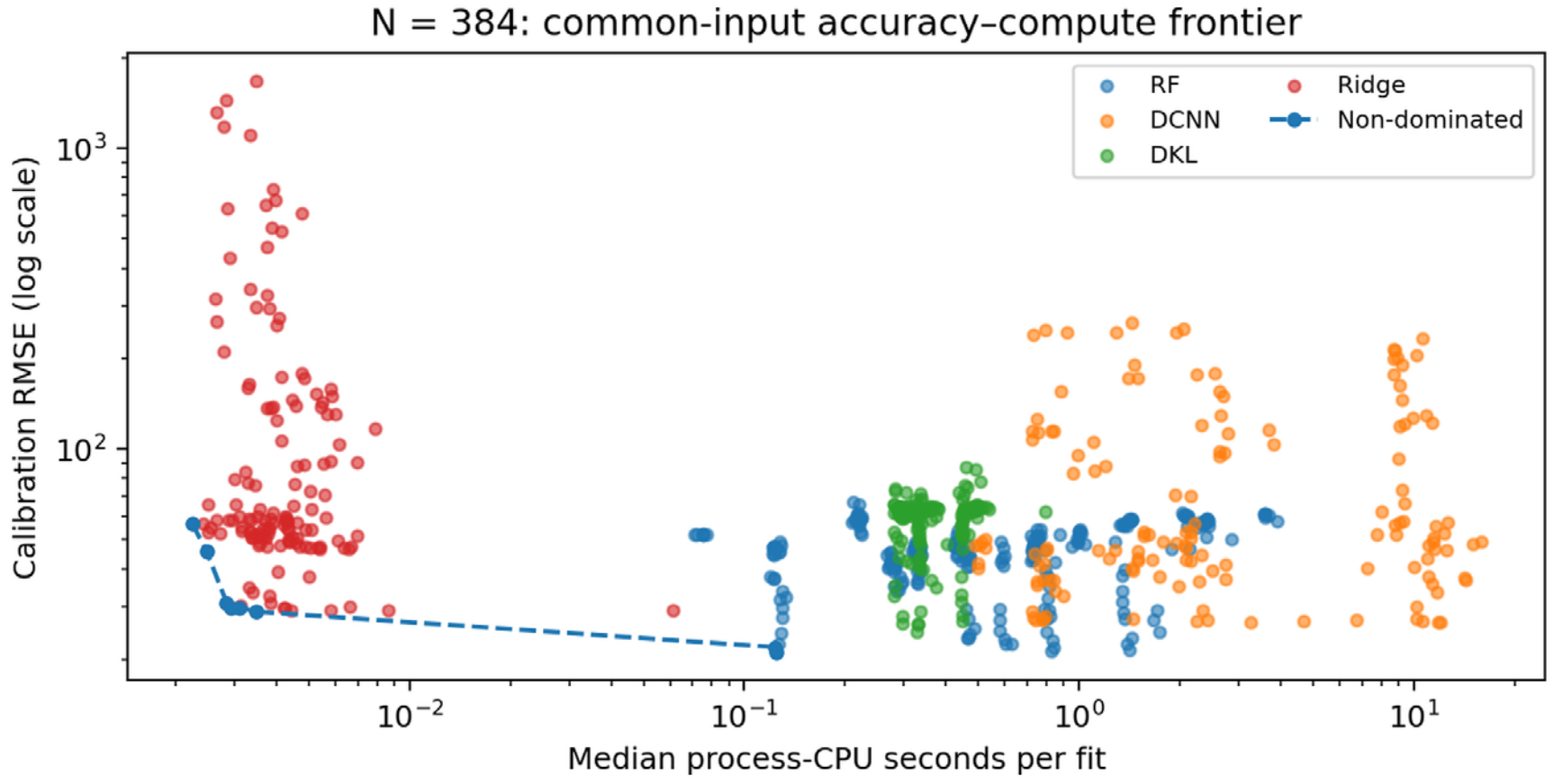

**(a)**

**Figure S2.** At $N$ = 384, common clean-L17 calibration error versus measured fitting CPU time for all screened profiles, including Ridge. This is the input panel used for selecting a practical online profile within each family.

The global-reference sweep instead fits all eleven non-Ridge profiles to all 1,543 candidates, at 48 imaging settings and with two fit seeds, giving 1,056 fits. For each family, the smallest mean matched-condition calibration error selects the reference configuration. Its test error is evaluated only after that choice. Matched descriptors use the same window and blur in fitting, calibration, and testing. Full-data references are empirical benchmarks rather than mathematical bounds; a smaller measured online gap, or even a negative gap, is not clipped. The full static workload is 3,072 fits.

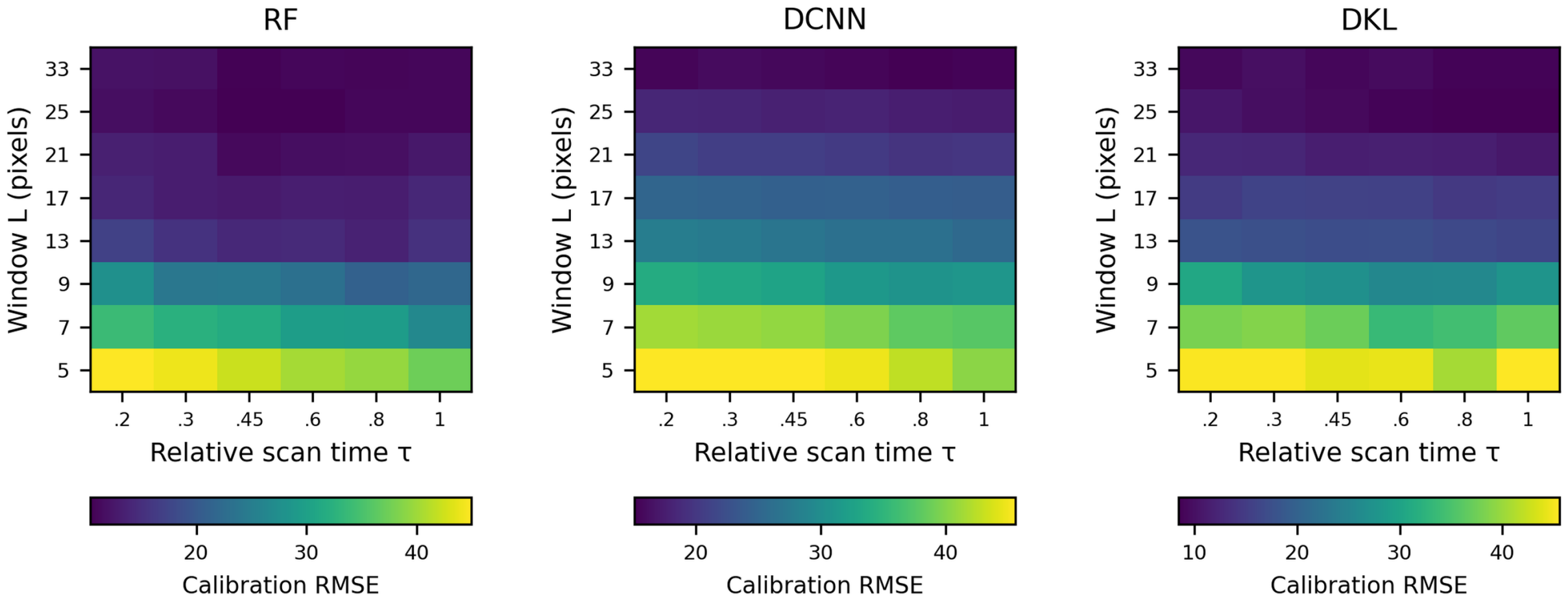

(a) (b) (c)

**Figure S3.** (a) RF: minimum full-data matched-condition calibration error across the tested profiles at each of the 48 image settings. This is an offline performance landscape, not a learned policy map. (b) DCNN: minimum full-data matched-condition calibration error across the tested profiles at each of the 48 image settings. This is an offline performance landscape, not a learned policy map. (c) DKL: minimum full-data matched-condition calibration error across the tested profiles at each of the 48 image settings. This is an offline performance landscape, not a learned policy map.

The Pareto test is applied separately at each sample count and evaluation convention. A configuration is non-dominated when no other tested configuration is at least as accurate and at least as fast with a strict improvement in one objective. CPU time is the principal timing axis, with elapsed time and numerical input dimension also recorded. The final matched-condition full-data frontier has ten RF and three DKL configurations. Family rankings on a common-input panel need not coincide with these matched-input rankings.

**Table S2. Profile-specific full-data configurations selected using calibration data; test errors and archive CPU times are reported afterwards.**

| Profile | L / τ | Hidden matched RMSE | Median archive CPU (s) |
|---|---|---|---|
| DCNN_compact | 33 / 1 | 18.66 | 2.44 |
| DCNN_medium | 33 / 0.8 | 17.90 | 5.86 |
| DCNN_wide | 33 / 0.2 | 18.37 | 39.31 |
| DKL_Matern_compact | 25 / 0.8 | 13.04 | 11.37 |
| DKL_RBF_compact | 33 / 0.2 | 11.86 | 9.64 |
| DKL_RBF_detail | 25 / 0.8 | 10.69 | 14.51 |
| RF_balanced | 33 / 0.6 | 12.76 | 4.96 |
| RF_dense | 25 / 0.6 | 13.21 | 10.87 |
| RF_detail | 25 / 0.6 | 13.18 | 9.93 |
| RF_fast | 25 / 0.6 | 14.15 | 2.19 |

| Profile | L / τ | Hidden matched RMSE | Median archive CPU (s) |
|---|---|---|---|
| RF_smooth | 33 / 0.8 | 13.00 | 3.61 |

A restricted same-session timing control refits each profile at one calibration-selected configuration. This control addresses timing sensitivity across the RF/DCNN and DKL execution sessions but is not a replacement for the full grid. Table S2 illustrates the practical consequence: DCNN_wide requires about seven times more CPU time than DCNN_medium without reaching a lower reference error.

### S3.2 Learning-curve maturity

Primary and alternative profiles are fitted on nested clean-L17 sample counts 12, 24, 48, 96, 192, 384, 768, and 1,543, with three repetitions. The constant-mean $N = 0$ baseline uses the full candidate-pool mean as part of offline calibration. A two-parameter saturating curve is fitted with that baseline fixed, giving an empirical asymptote and $N_0$. Two hundred residual-bootstrap fits provide the descriptive 10th–90th percentile ranges. The asymptote is distinct from the separately selected global oracle.

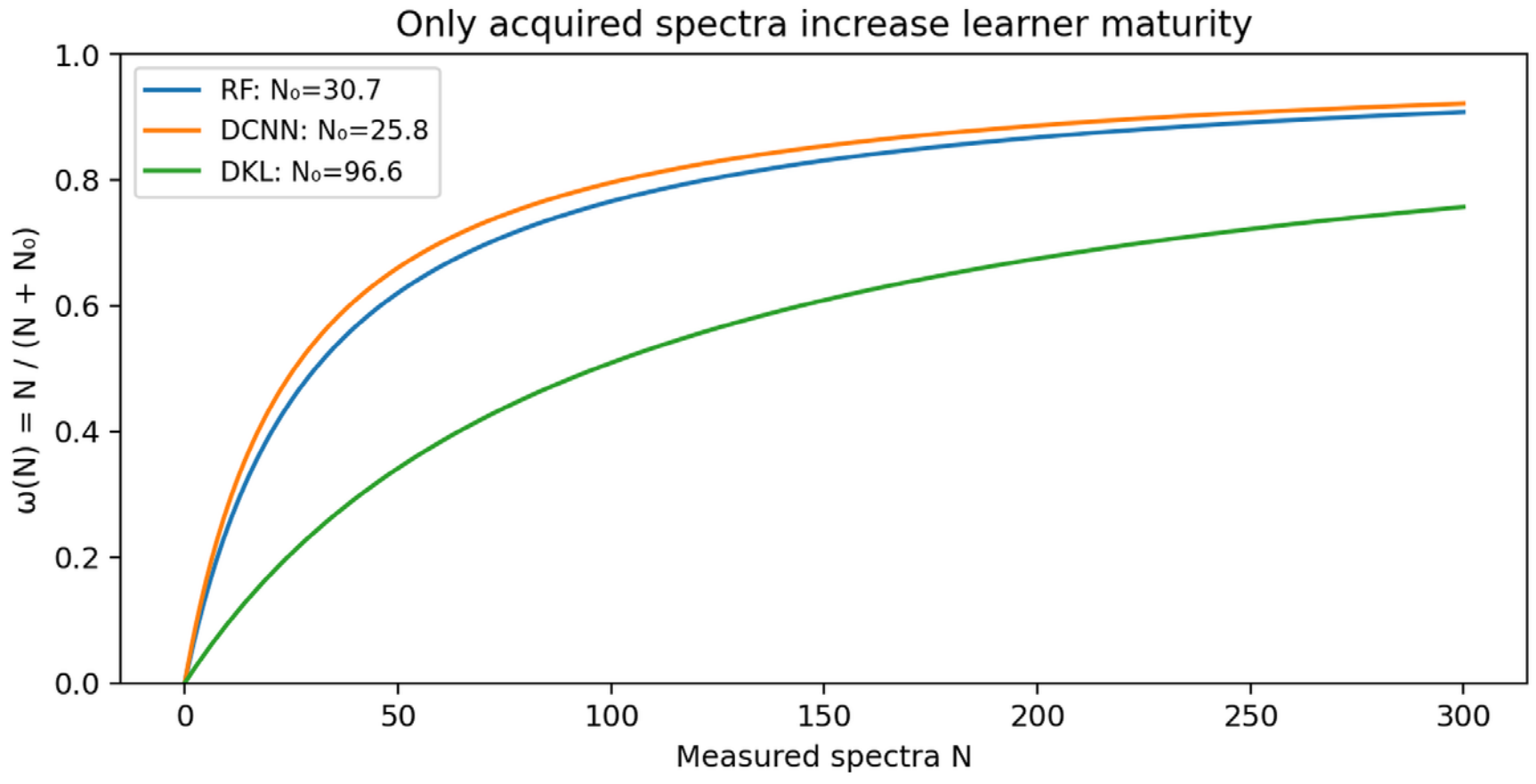

(a)

**Figure S4.** Maturity weights used by the three primary policies. Only measured spectra increase $N$. DKL's larger $N_0$ downweights early uncertainty changes more strongly; it does not imply that GP variance becomes calibrated at $N = N_0$.

**Table S3. Calibrated maturity parameters for primary and alternative profiles.**

| Profile | Role | $N_0$ | Bootstrap 10–90% | Fitted $E_\infty$ | Curve-fit residual RMSE |
|---|---|---|---|---|---|
| RF_fast | primary | 30.67 | 24.4–38.3 | 17.70 | 2.74 |
| RF_detail | alternative | 29.85 | 23.9–37.4 | 16.36 | 2.73 |
| DCNN_compact | primary | 25.78 | 19.8–34.9 | 24.49 | 2.86 |
| DCNN_wide | alternative | 37.46 | 17.7–91.8 | 21.56 | 8.82 |
| DKL_RBF_compact | primary | 96.58 | 69.3–154.2 | 16.24 | 4.51 |

| Profile | Role | $N_0$ | Bootstrap 10–90% | Fitted E∞ | Curve-fit residual RMSE |
|---|---|---|---|---|---|
| DKL_Matern_compact | alternative | 90.74 | 58.1–138.3 | 13.18 | 4.87 |

The parameter $N_0$ is fixed during a campaign and is not recalibrated after changing window size, blur, or sampling bias. A smaller $N_0$ describes earlier saturation of this learning curve, not necessarily a more accurate or more trustworthy predictor. Only spectra increase the labelled count. Uncertainty calibration and reward alignment are evaluated independently of maturity.

## S4. Sequential controller, baselines, and campaign replication

This section defines the legal actions and observable state of the actor–critic controller (main Section 4.4), its numerical settings, the baseline controllers, and the replication scheme behind the uncertainty bands in Figures 4 and 7.

### S4.1 Legal actions and observable state

Each candidate progresses from preview-only to imaged-but-unlabelled to imaged-and-labelled. IMAGE changes its available descriptor and candidate score; SPECTRUM reveals its target, updates the best measured response, and refits the regressor. The action set has 48 IMAGE configurations plus one SPECTRUM operation. Candidate selection is handled by the acquisition rule, not by the actor: predictive mean for exploitation, predictive standard deviation for exploration, and a uniform choice for the random baseline.

The pending queue contains at most eight centers. IMAGE is infeasible if that limit is reached, if no unseen candidate remains, or if the remaining budget cannot cover the image plus one future spectrum. SPECTRUM is infeasible without a pending center or 0.30 remaining time. Reserving one spectrum prevents a completely unusable final image, but does not require that every pending image eventually receives a label. Permission applies only to the image's indexed center, not every pixel contained in its window.

The actor and critic use the same 20-component observable summary. It includes budget, label and operation counts, pending availability, fixed-panel uncertainty, maturity, already observed best response, candidate-score summaries, recent reward, and mean acquired settings. This is a compact state representation, not proof that it renders the full scientific posterior Markov. Hidden test errors, oracle gaps, and unrevealed responses are excluded.

### S4.2 Replay-free actor–critic

The actor is a linear softmax policy and the critic a linear value function. Actor weights are initialized from a normal distribution with standard deviation 0.006 using policy seeds 11 and 29. An operation-balancing normalization subtracts the logarithm of the feasible IMAGE count from image logits so that the large number of imaging settings does not overwhelm SPECTRUM at initialization. Infeasible actions are masked before sampling.

The implementation uses undiscounted TD(0), an actor step size of 0.015, a critic step size of 0.01, and an entropy-gradient coefficient of 0.003. The update uses 30 times the logged scientific reward, and the TD residual is clipped to [−2, 2]. Actor weights are clipped to [−6, 6]. These numerical transformations are separate from the unscaled rewards plotted in the manuscript. Frozen deployment retains stochastic action sampling; it is not greedy argmax deployment. No replay buffer, synthetic transition planning, or generative image look-ahead is used.

### S4.3 Comparison policies and independent replication

The acquisition-rule baseline uses the same non-random candidate ranking as RL but randomized operation/settings choices, with a preference for clearing the pending set once four images are queued. The fixed clean baseline alternates IMAGE at $L = 17$, $\tau = 1$ and SPECTRUM. The random-location baseline changes candidate selection to uniform sampling while retaining comparable legal actions. The common mask permits up to eight pending images. Baselines therefore test different components of the workflow and should not be relabelled as identical controllers.

For each family and objective, two policy seeds are trained for twelve campaigns, giving 144 RL training campaigns. Frozen RL evaluation uses six experimental initializations, seeds 1001–1006, crossed with both policy seeds. The 72 primary RL evaluations are supplemented by 108 primary-profile baseline evaluations and 36 alternative-profile acquisition-rule evaluations: 216 evaluations and 360 total recorded campaigns. Random-baseline rows under the two objective labels can repeat the same trajectory and are not independent additional evidence.

Uncertainty ribbons first average the two policy replicates within each experimental initialization and then show the mean and one standard deviation across the six initializations. The twelve deployments are not twelve independent specimens. Operation histories chosen for the main figures use the fixed first initialization (1001) and policy seed 11, rather than the best or median final outcome.

## S5. Complete sequential performance comparisons

This section reports the complete endpoint comparisons summarized in main Sections 5.1–5.2, including both evaluation panels, all baselines, and the alternative predictor profiles.

### S5.1 Common-input and family-reference evaluation

All campaigns are evaluated on the same held-out response locations using two descriptor conventions. The common-input panel always supplies clean 17-pixel images to all models. The family-reference panel instead uses each family's fixed calibration-selected oracle descriptor. The corresponding errors answer different questions, and an oracle gap may reflect both sample limitations and the difference between an economical online profile and its family's best offline profile. The family-reference histories are retained in Figure S5 rather than substituted for the common-input main-text comparison.

**Table S4. Primary RL results under the linear window-cost model. Prediction error uses identical clean 17-pixel test inputs. Values are means ± between-initialization SD; exact hits retain the twelve-deployment denominator.**

| Predictor | Objective | Common-input RMSE | Best measured area | Exact-maximum hits |
|---|---|---|---|---|
| RF | Exploitation | 35.17 ± 2.23 | 250.27 ± 2.11 | 2/12 |
| RF | Exploration | 33.96 ± 1.32 | 234.77 ± 6.45 | 0/12 |
| DCNN | Exploitation | 40.32 ± 3.98 | 247.97 ± 0.00 | 0/12 |
| DCNN | Exploration | 36.55 ± 1.68 | 249.74 ± 1.93 | 3/12 |
| DKL | Exploitation | 45.52 ± 5.34 | 251.45 ± 2.92 | 5/12 |
| DKL | Exploration | 37.97 ± 2.41 | 252.84 ± 2.18 | 8/12 |

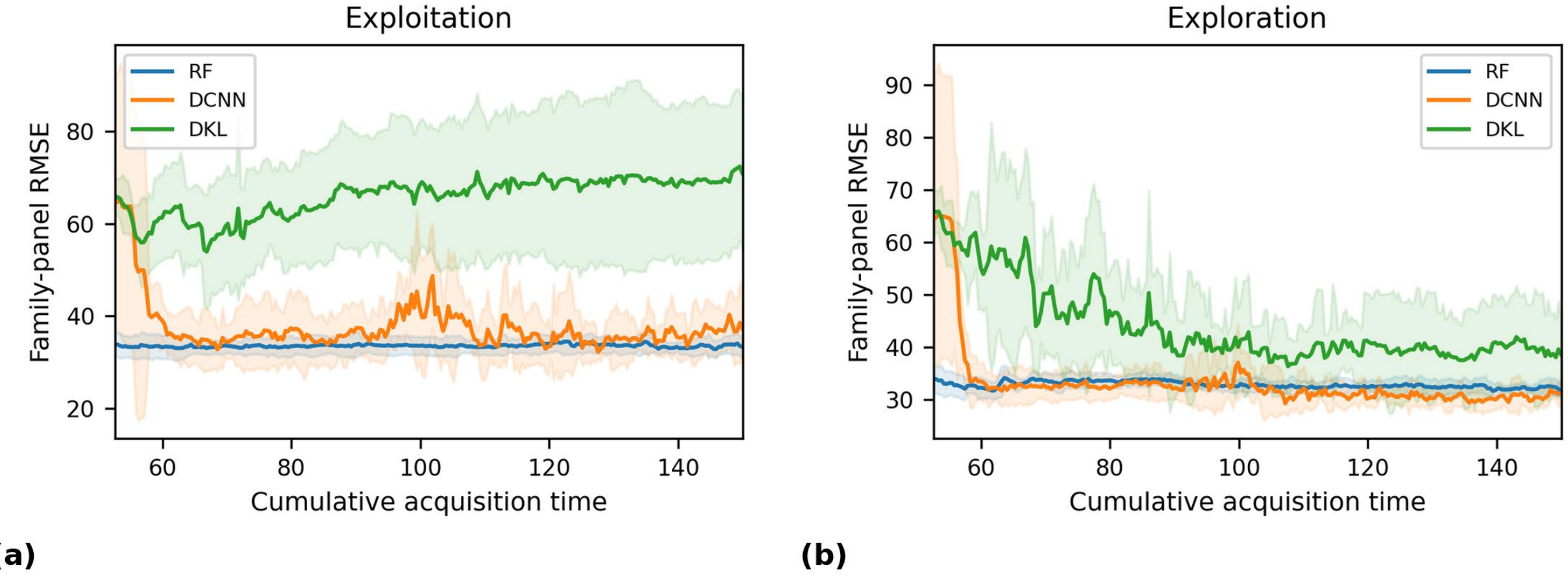

**Figure S5.** (a) Family-oracle input error under exploitation. Each family has a different fixed audit descriptor, so these curves are not an identical-input architecture ranking. (b) Family-oracle input error under exploration. Both panels evaluate the same 300 held-out response locations but use different descriptor conventions.

### S5.2 Baselines and alternative profiles

Table S5 includes both learned and simpler controllers. The source uses campaign means for its endpoints, and exact-maximum counts retain the number of deployments in their denominators. Figure S6 provides the corresponding discovery and prediction comparisons.

**Table S5. Full primary-profile comparison against acquisition-rule, fixed-clean, and random-location baselines.**

| Family | Objective | Policy | Common RMSE | Best area | Maximum found |
|---|---|---|---|---|---|
| RF | exploitation | RL | 35.17 | 250.27 | 2/12 |
| RF | exploitation | acquisition | 35.16 | 251.85 | 2/6 |
| RF | exploitation | random | 34.51 | 239.62 | 1/6 |
| RF | exploitation | alternating | 36.51 | 250.33 | 2/6 |
| RF | exploration | RL | 33.96 | 234.77 | 0/12 |
| RF | exploration | acquisition | 33.74 | 233.84 | 0/6 |
| RF | exploration | random | 34.51 | 239.62 | 1/6 |
| RF | exploration | alternating | 32.90 | 238.73 | 0/6 |
| DCNN | exploitation | RL | 40.32 | 247.97 | 0/12 |
| DCNN | exploitation | acquisition | 40.81 | 248.35 | 0/6 |
| DCNN | exploitation | random | 35.71 | 239.62 | 1/6 |
| DCNN | exploitation | alternating | 40.97 | 247.97 | 0/6 |
| DCNN | exploration | RL | 36.55 | 249.74 | 3/12 |
| DCNN | exploration | acquisition | 36.17 | 250.33 | 2/6 |

| Family | Objective | Policy | Common RMSE | Best area | Maximum found |
|---|---|---|---|---|---|
| DCNN | exploration | random | 35.71 | 239.62 | 1/6 |
| DCNN | exploration | alternating | 35.90 | 247.97 | 0/6 |
| DKL | exploitation | RL | 45.52 | 251.45 | 5/12 |
| DKL | exploitation | acquisition | 46.63 | 251.11 | 1/6 |
| DKL | exploitation | random | 43.07 | 239.62 | 1/6 |
| DKL | exploitation | alternating | 33.26 | 251.09 | 2/6 |
| DKL | exploration | RL | 37.97 | 252.84 | 8/12 |
| DKL | exploration | acquisition | 40.90 | 253.44 | 4/6 |
| DKL | exploration | random | 43.07 | 239.62 | 1/6 |
| DKL | exploration | alternating | 33.59 | 237.10 | 1/6 |

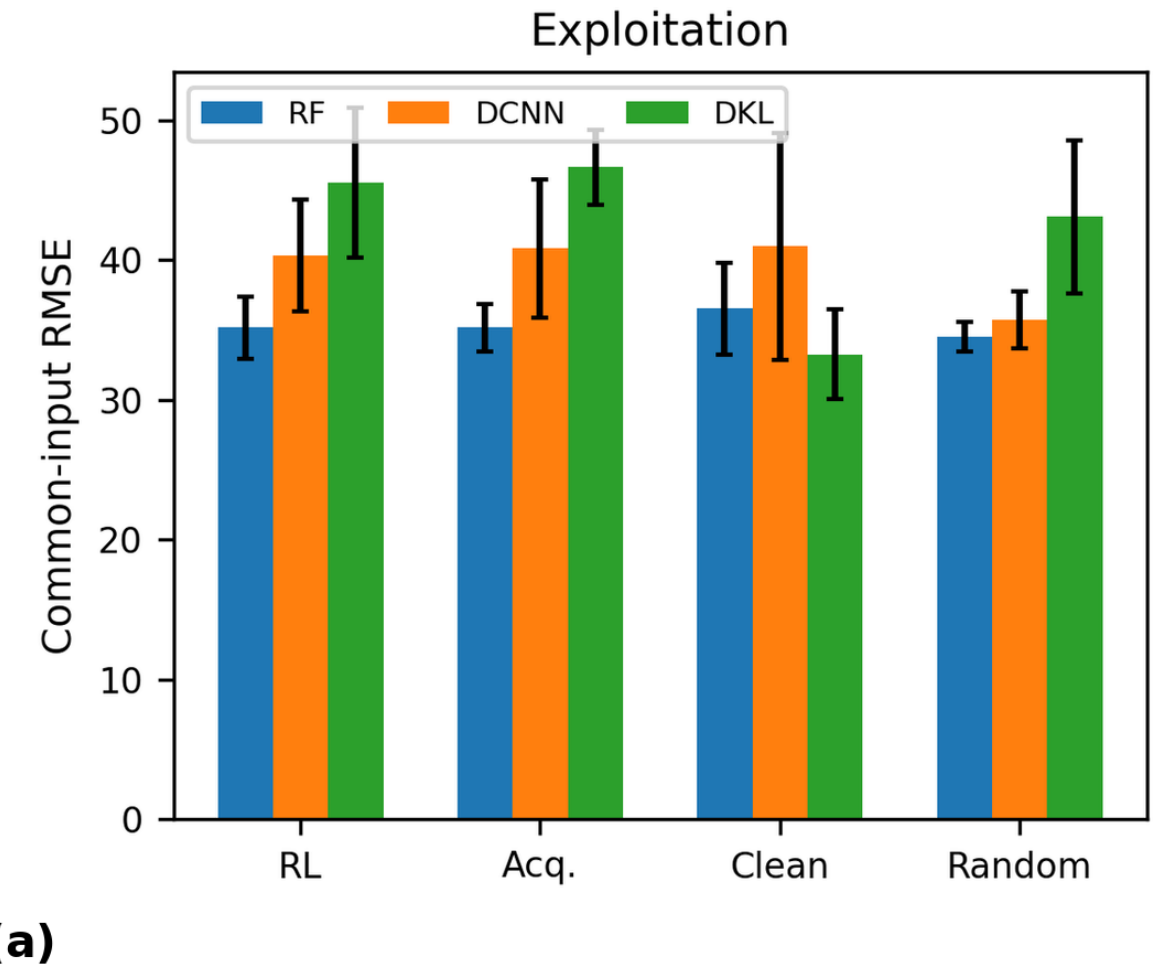

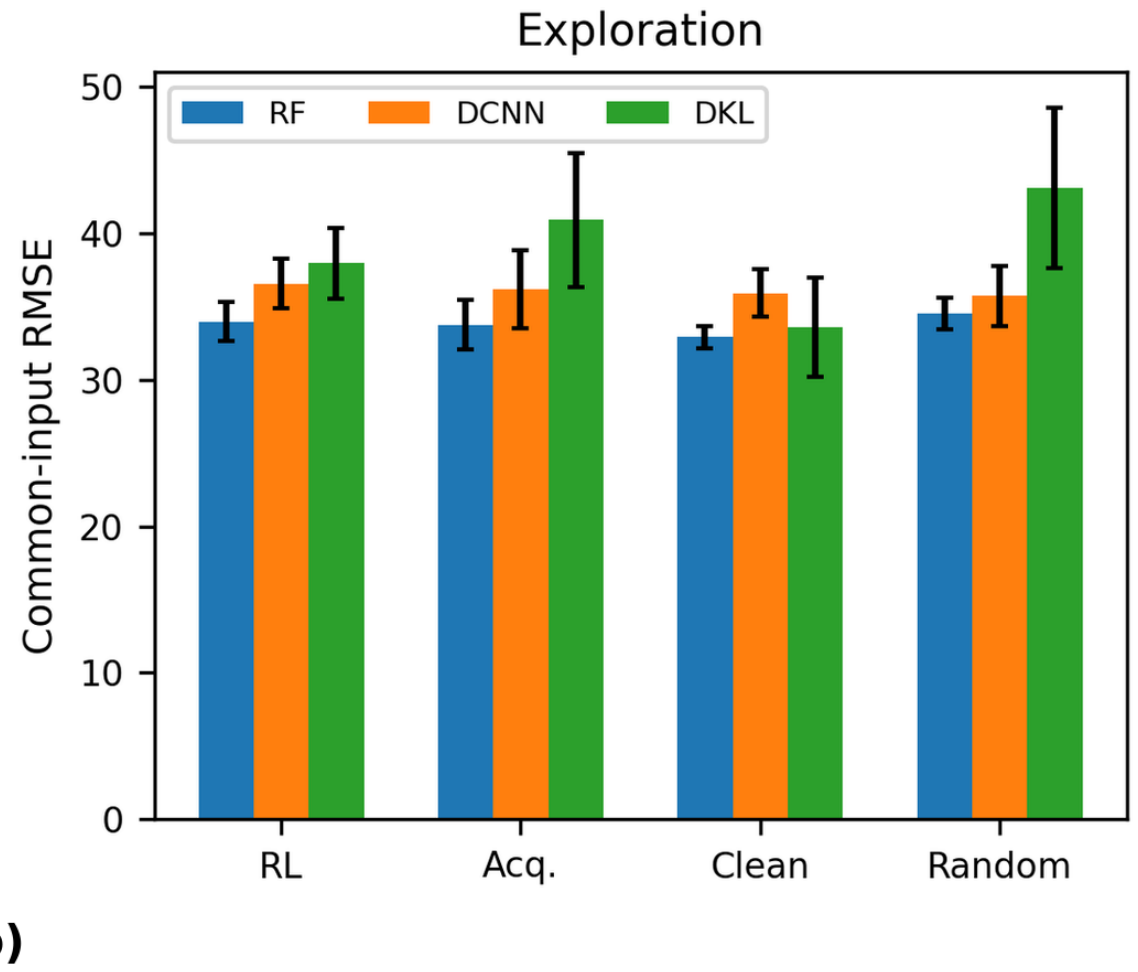

**Figure S6.** (a) Exploitation: final common-input RMSE for the learned controller and three primary baselines. Error bars are between-initialization SD, with policy replicas collapsed before computing spread. (b) Exploration: final common-input RMSE for the learned controller and three primary baselines. Error bars are between-initialization SD, with policy replicas collapsed before computing spread.

Alternative predictive profiles are evaluated with the acquisition-rule controller; they are not additional retrained RL policies. Their outcomes therefore test the profile–acquisition combination rather than an isolated change in policy learning.

**Table S6. Alternative-profile acquisition-rule deployments.**

| Family / profile | Objective | Common RMSE | Best area | Maximum found |
|---|---|---|---|---|
| DCNN / DCNN_wide | exploitation | 34.26 | 249.49 | 0/6 |
| DCNN / DCNN_wide | exploration | 35.45 | 246.80 | 0/6 |

| Family / profile | Objective | Common RMSE | Best area | Maximum found |
|---|---|---|---|---|
| DKL / DKL_Matern_compact | exploitation | 46.48 | 249.87 | 2/6 |
| DKL / DKL_Matern_compact | exploration | 37.59 | 249.48 | 2/6 |
| RF / RF_detail | exploitation | 36.67 | 251.09 | 2/6 |
| RF / RF_detail | exploration | 33.62 | 239.38 | 0/6 |

RF exploitation with the acquisition-rule baseline reaches mean best area 251.85, versus 250.27 for RL. DKL exploration with that baseline reaches approximately 253.44, versus 252.84 for RL, while RL has lower common-input prediction error, approximately 37.97 versus 40.90. Thus the present data support objective-dependent trade-offs, not universal superiority of the learned controller. The accessible response maximum is 255.0.

**S6. Additional operation, window, scan-time, and spatial diagnostics**

This section supports main Sections 5.3–5.4 with the complete three-model acquisition histories. Figure S7 pairs window-side and relative-time histories for exploitation; Figure S8 gives the corresponding exploration histories. Figures S9 and S10 pair implied blur and actual IMAGE duration for exploitation and exploration, respectively. Duration depends jointly on $L$ and $\tau$ in this primary experiment. These histories accompany the redesigned representative operation and spatial-trajectory plots in the main manuscript. The histories are choices of a bounded controller; they do not by themselves establish convergence to an optimal schedule.

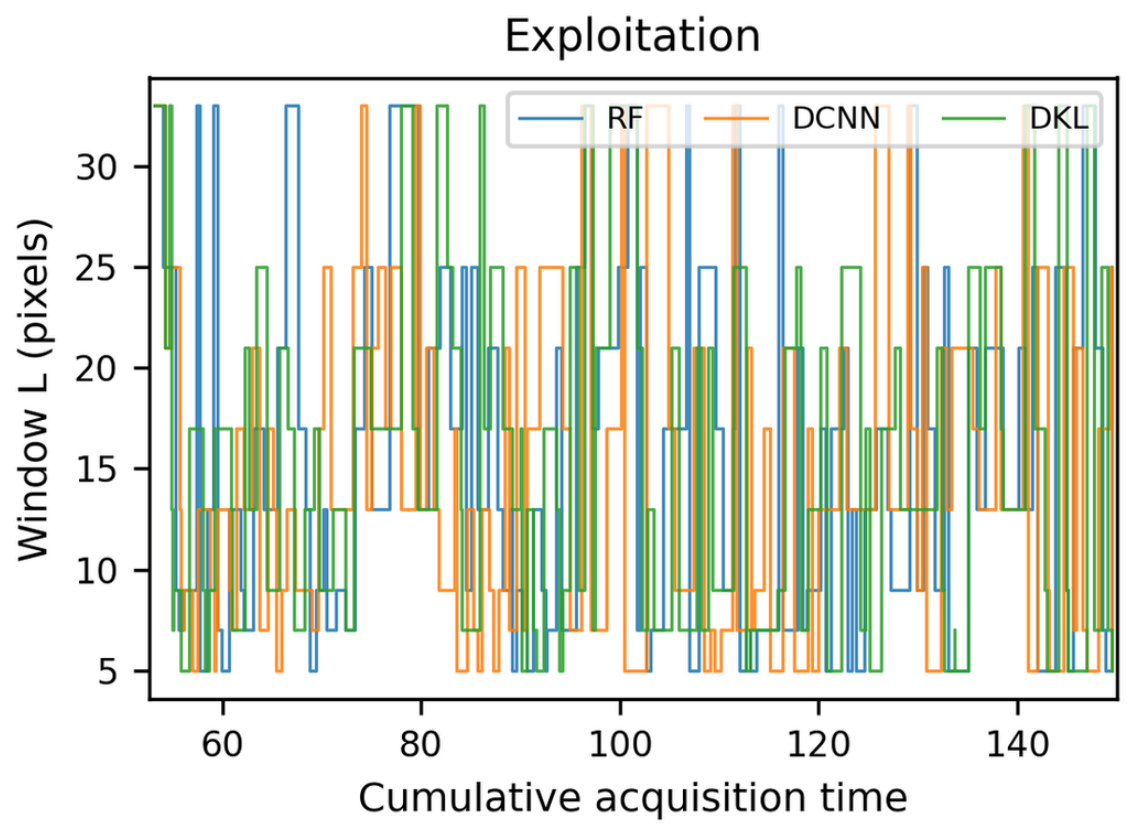

(a)

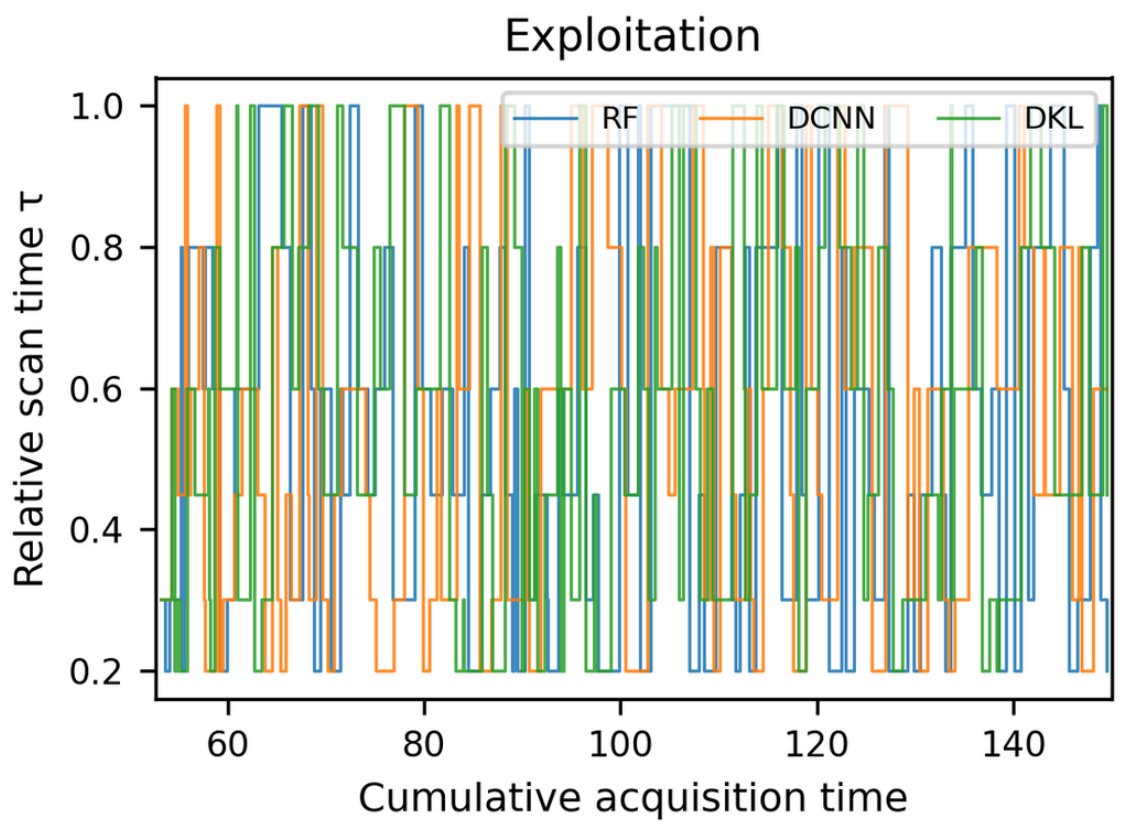

(b)

**Figure S7.** (a) Exploitation: actual image-window side versus cumulative time for the representative three-model comparison. Abrupt changes are selected controls, not changes in particle size. (b) Exploitation: relative scan-time settings for the same trajectories. Lower τ means faster scanning and larger assumed blur.

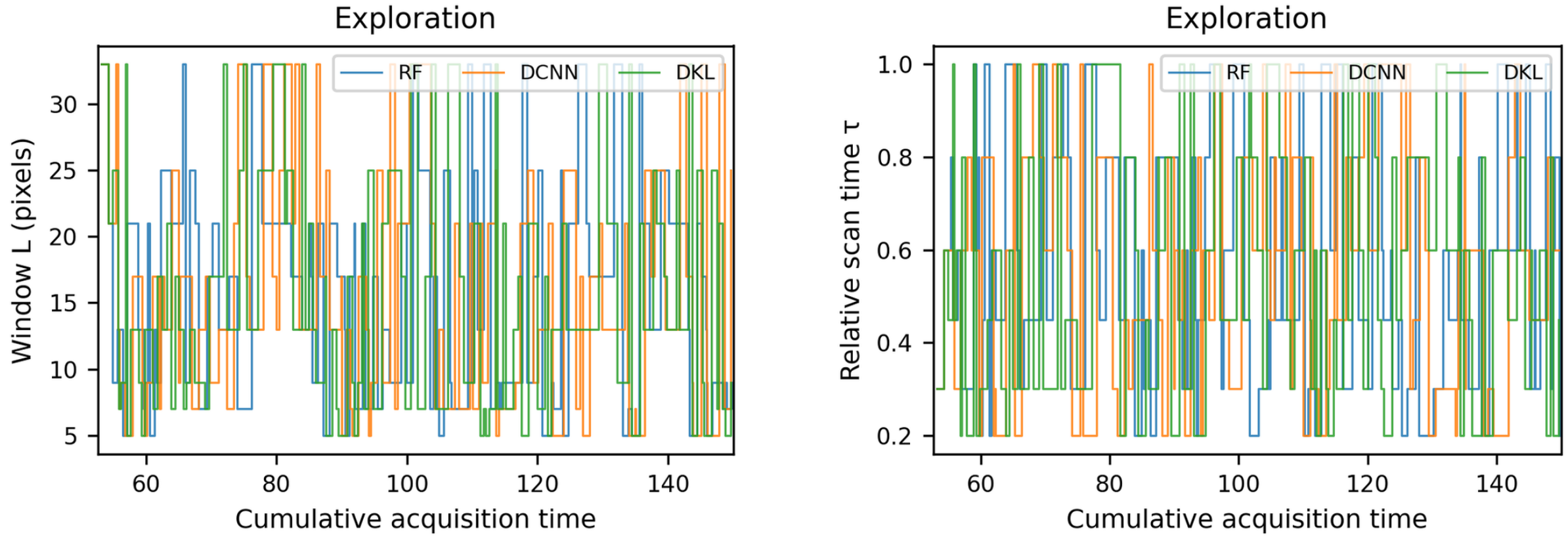

**(a)** **(b)**

**Figure S8.** (a) Exploration: actual image-window side versus cumulative time for the representative three-model comparison. Abrupt changes are selected controls, not changes in particle size. (b) Exploration: relative scan-time settings for the same trajectories. Lower τ means faster scanning and larger assumed blur.

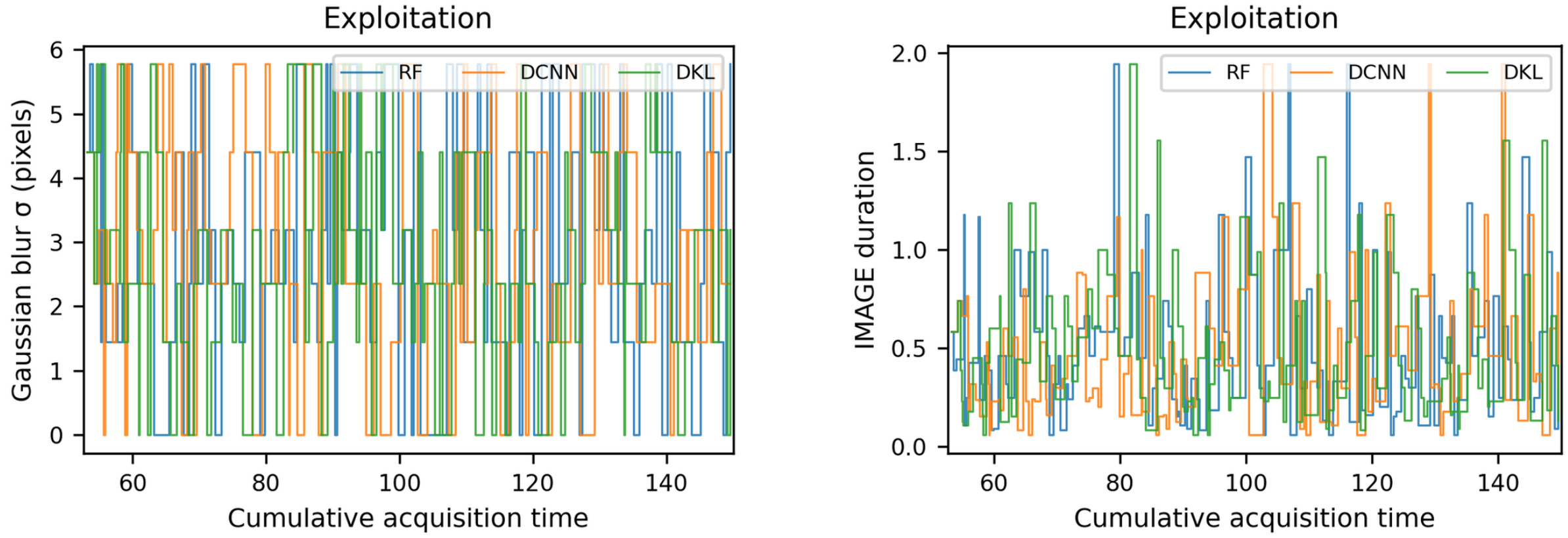

**(a)** **(b)**

**Figure S9.** (a) Exploitation: blur implied by each selected scan time under the fixed simulator law. This plot is not an independently measured instrumental resolution trace. (b) Exploitation: duration of each IMAGE operation. Duration depends on both $L$ and $\tau$, so scan-time factor alone does not describe the cost actually paid.

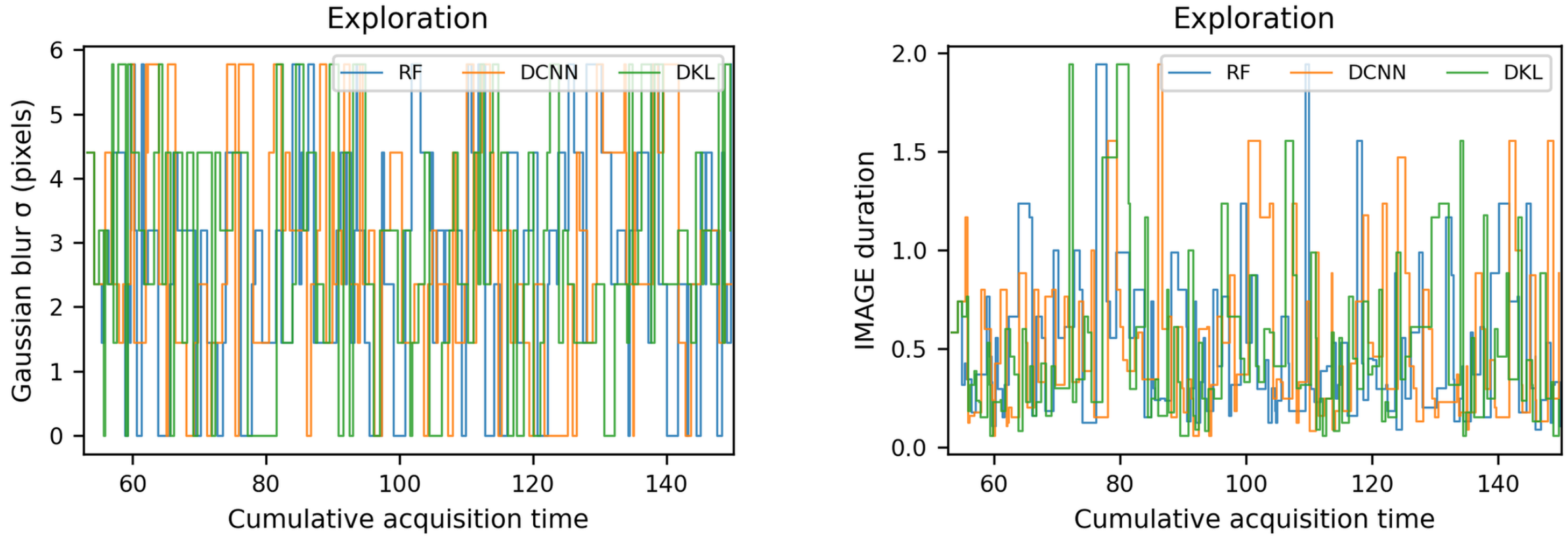

**(a)** **(b)**

**Figure S10.** (a) Exploration: blur implied by each selected scan time under the fixed simulator law. This plot is not an independently measured instrumental resolution trace. (b) Exploration: duration of each IMAGE operation. Duration depends on both $L$ and $\tau$, so scan-time factor alone does not describe the cost actually paid.

A new image buys a descriptor at one center but does not create a label. Image count, spectral count, pending queue occupancy, and acquisition time must therefore be interpreted together. Faster images can increase throughput, but more spectra are not automatically more informative. In the revised main trajectory figure, discrete parameter cells are separated and repeated settings receive display-only offsets; the spatial panels show measurement points without connecting paths. No move-time penalty or continuous beam trajectory is implied.

## S7. Internal rewards, uncertainty, and independent error audits

This section supports main Section 5.5 with the instantaneous rewards and the independent error audits. The normalized fixed-panel uncertainty is the mean predictive variance over the 32 unchanged images divided by the seed-response variance scale. The exploration reward weights its decrease by the pre-action maturity. Discovery reward is the increase in the best measured response divided by the seed-response standard-deviation scale. IMAGE gives zero immediate reward for both objectives; spectroscopy can give negative exploration reward but cannot decrease the observed record in the noiseless cached-response setting. Negative uncertainty gains are preserved.

The record-improvement reward telescopes before update scaling and clipping. Maturity-weighted uncertainty increments do not telescope because maturity changes with acquired label count. Neither diminishing uncertainty nor a rising shaped return is algebraically equivalent to decreasing prediction error. Figure S11 retains instantaneous signals, while main Figure 7 shows cumulative rewards on separate scales. Figure S12 compares uncertainty-reward and independent error diagnostics.

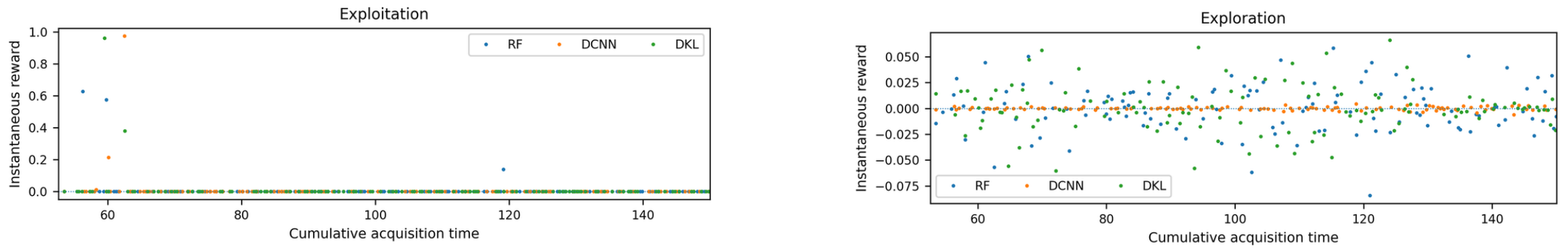

**(a)** **(b)**

**Figure S11.** (a) Exploitation: one-step spectroscopy rewards for the representative deployments. IMAGE rewards are identically zero. Objective-specific figures prevent small exploration rewards from being visually suppressed by discovery jumps. (b) Exploration: one-step spectroscopy rewards for the representative deployments. IMAGE rewards are identically zero. Objective-specific figures prevent small exploration rewards from being visually suppressed by discovery jumps.

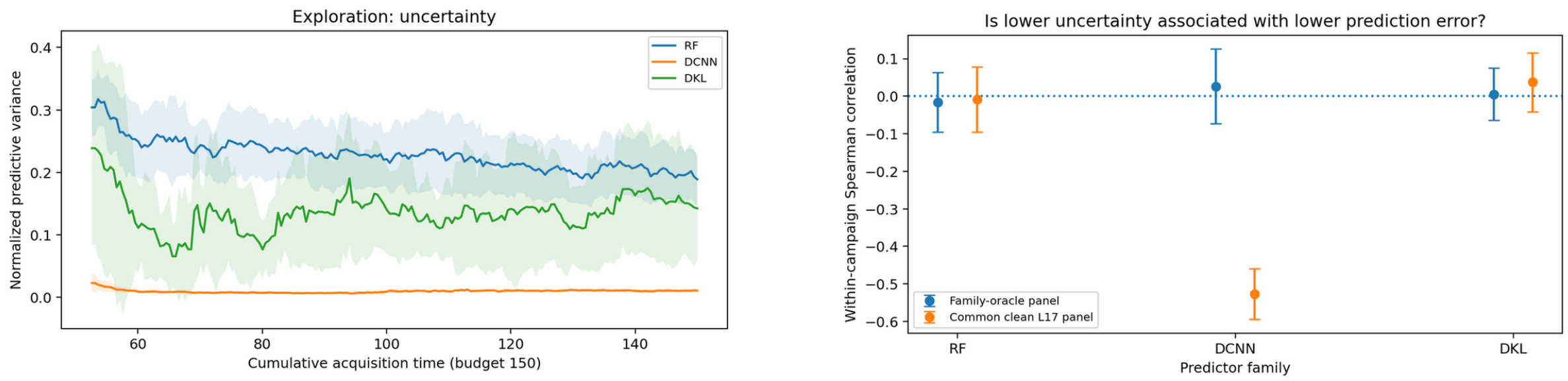

**(a)** **(b)**

**Figure S12.** (a) Normalized survey uncertainty under exploration for all families. Its estimator-specific scale is distinct from common-input prediction error and from measured-response discovery. (b) Association of realized uncertainty reduction with true MSE reduction, using the family-specific and common-L17 audit panels separately. Error bars describe between-initialization variation, not a significance test over independent specimens.

The mean within-campaign Spearman rank association between uncertainty reduction and common-input MSE improvement is approximately −0.009 for RF, −0.527 for DCNN, and 0.037 for DKL in the primary pure-exploration RL comparisons. The mean fraction of positive uncertainty-gain actions that nevertheless worsen common-input error is approximately 50%, 70%, and 48%, respectively. These are diagnostics of the actions executed, not counterfactual ranking tests of every feasible action. Their weakness does not imply that uncertainty is useless for candidate selection over a long horizon; it does prevent interpreting every positive uncertainty reward as demonstrated scientific progress.

The ±1.96-standard-deviation coverage calculation is a diagnostic of the recorded uncertainty proxies. For RF and DCNN those bands are not automatically Gaussian predictive intervals. For DKL, latent variance excludes fitted observation noise; a second column includes it. Even that observation variance is conditional on fitted encoder/kernel parameters rather than a fully Bayesian posterior over all parameters.

**Table S7. Final prediction-uncertainty diagnostics; coverage is not assumed to be calibrated.**

| Family | Objective | ±1.96√variance coverage | Coverage incl. DKL noise | SD–absolute-error rank corr. |
|---|---|---|---|---|
| RF | exploitation | 86.8% | — | 0.232 |
| RF | exploration | 83.8% | — | 0.228 |
| DCNN | exploitation | 51.7% | — | -0.081 |
| DCNN | exploration | 46.3% | — | -0.104 |
| DKL | exploitation | 69.2% | 78.2% | 0.534 |
| DKL | exploration | 72.9% | 88.4% | 0.192 |

### S8. Size-independent acquisition cost and size-dependent blur

This section reports the alternative instrument-response scenario summarized in main Section 5.6. Only the image-cost and blur laws are changed; the policies are retrained and all other settings are kept fixed, so that the comparison isolates the effect of the assumed instrument response.

### S8.1 Controlled change to the observation and cost laws

The alternative scenario replaces the size penalty in the image cost with stronger blur for large windows acquired at the same time. All controls, initial information, scientific objectives, primary predictor profiles, fitted maturity parameters, campaign seeds, controller updates, and 150-unit budgets are held fixed. Spectroscopy remains 0.30 units. New policies are trained using the changed observations and costs; this is not a post hoc relabelling of old trajectories.

The new law is

$$c_I(L,\tau)=\tau,\qquad v_{\mathrm{rel}}(L,\tau)=\frac{(L/17)^p}{\tau},\qquad \sigma(L,\tau)=\left[\frac{\max(v_{\mathrm{rel}}-1,0)}{0.12}\right]^{1/2}. \tag{S1}$$

The executed exponent is $p = 2$. At fixed native pixel pitch, an $L \times L$ raster has approximately $L$ lines of length $L$, so neglecting flyback gives a path-length proxy proportional to $L^2$. At fixed line count the corresponding exponent could be $p = 1$, but that alternative is shown only as a mathematical sensitivity, not as an additional sequential experiment. These laws are assumed, not inferred from instrument timing or measured transfer functions. The old and new scenarios agree exactly at $L = 17$ but are not two globally consistent physical calibrations of one microscope.

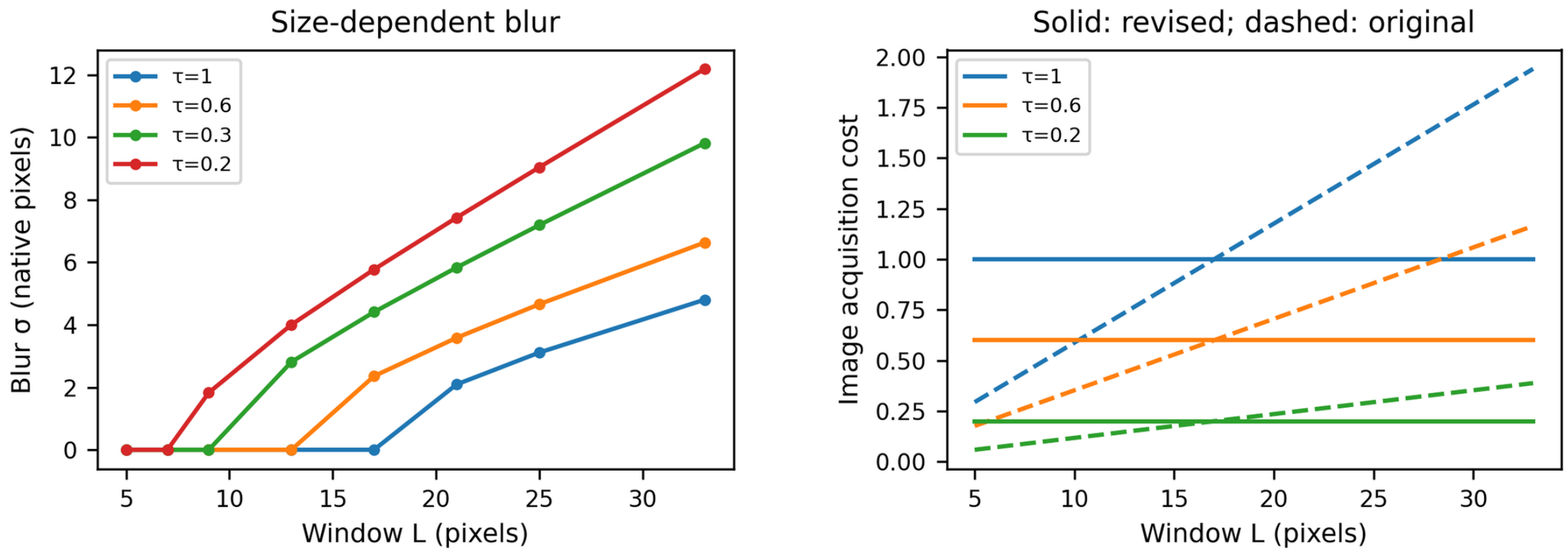

**(a)** **(b)**

**Figure S13.** Size-independent-cost sensitivity study. (a) Revised Gaussian bandwidth in native pixels. At equal image time, larger windows experience greater blur. The inherited threshold produces zero blur below the reference speed. (b) Original and revised image costs. Under the revised rule, all sizes cost the same at a fixed time setting; the spectrum remains a separate 0.30-unit operation.

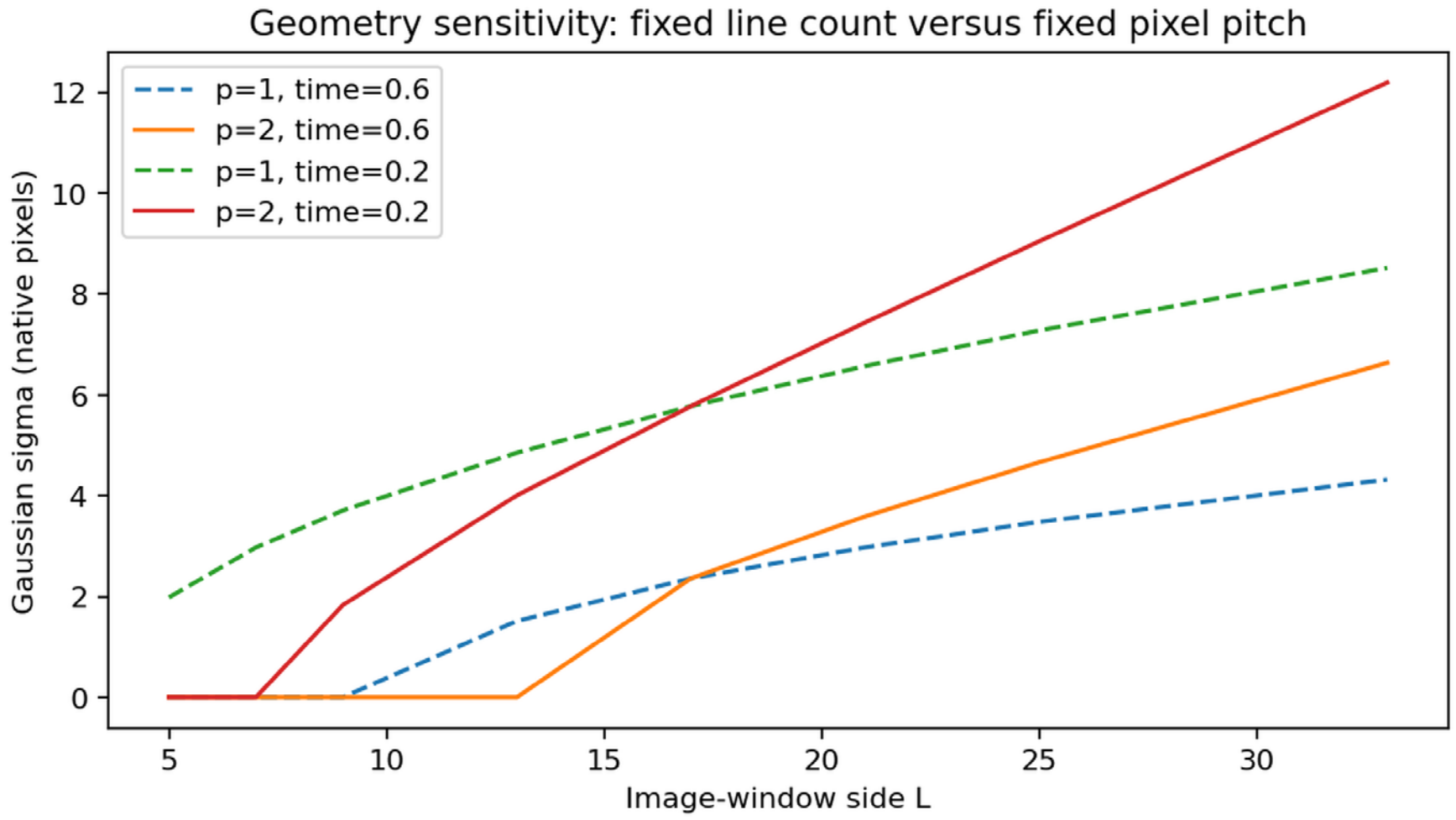

**(a)**

**Figure S14.** Size-independent-cost sensitivity study. Geometry sensitivity. $p$ = 2 is the executed fixed-pixel-pitch model; $p$ = 1 is a possible fixed-line-count alternative. These curves alone are not results of two RL experiments.

Blur is applied in native pixels before pooling. The inherited threshold means that a small window can have zero added blur at a fast time setting, while a large window can retain blur even at $\tau = 1$. For $L = 33$, reaching the zero-blur threshold would require $\tau \approx 3.77$, outside the retained action range 0.2–1.0. Thus the current action set does not contain a clean 33-pixel image in the revised scenario. This restriction must be considered when comparing selected configurations.

**Table S8. Revised blur and the acquisition time needed to reach the zero-added-blur threshold.**

| L | σ at τ=1 | σ at τ=0.45 | σ at τ=0.2 | τ for vrel=1 |
|---|---|---|---|---|
| 5 | 0.00 | 0.00 | 0.00 | 0.09 |
| 9 | 0.00 | 0.00 | 1.83 | 0.28 |
| 17 | 0.00 | 3.19 | 5.77 | 1.00 |
| 25 | 3.11 | 5.63 | 9.04 | 2.16 |
| 33 | 4.80 | 7.84 | 12.19 | 3.77 |

### S8.2 Static recalibration and computational frontiers

The revised static study uses the established primary profiles, not a new full hyperparameter sweep. Three profiles × 48 image settings × two sample counts (384 and 1,543) × two fit seeds give 576 fits. The same pre-experiment normalization and evaluation conventions are retained. All 36 full-data fits at the shared $L = 17$ reference reproduce the original RMSE exactly, providing a control on the implementation change. The original maturity functions and family prediction references are frozen for paired sequential evaluation. The newly selected matched-input static references below are a distinct diagnostic and are not silently substituted for those frozen trajectory references.

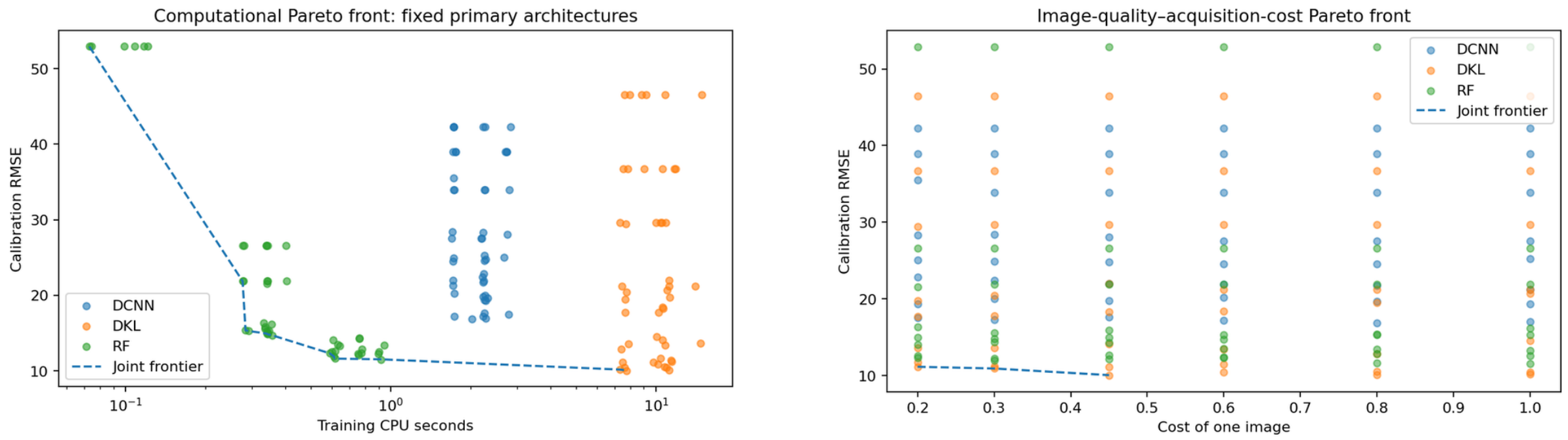

**(a)** **(b)**

**Figure S15.** Size-independent-cost sensitivity study. (a) Accuracy versus current-session training CPU time. The frontier covers the three frozen primary architectures and all image settings. It must not be compared to the old all-architecture frontier as though the candidate sets were identical. (b) Static acquisition-cost frontier. Each point uses 1,543 training spectra, so the horizontal axis is one image's cost rather than total campaign cost. The campaign comparisons are reported separately.

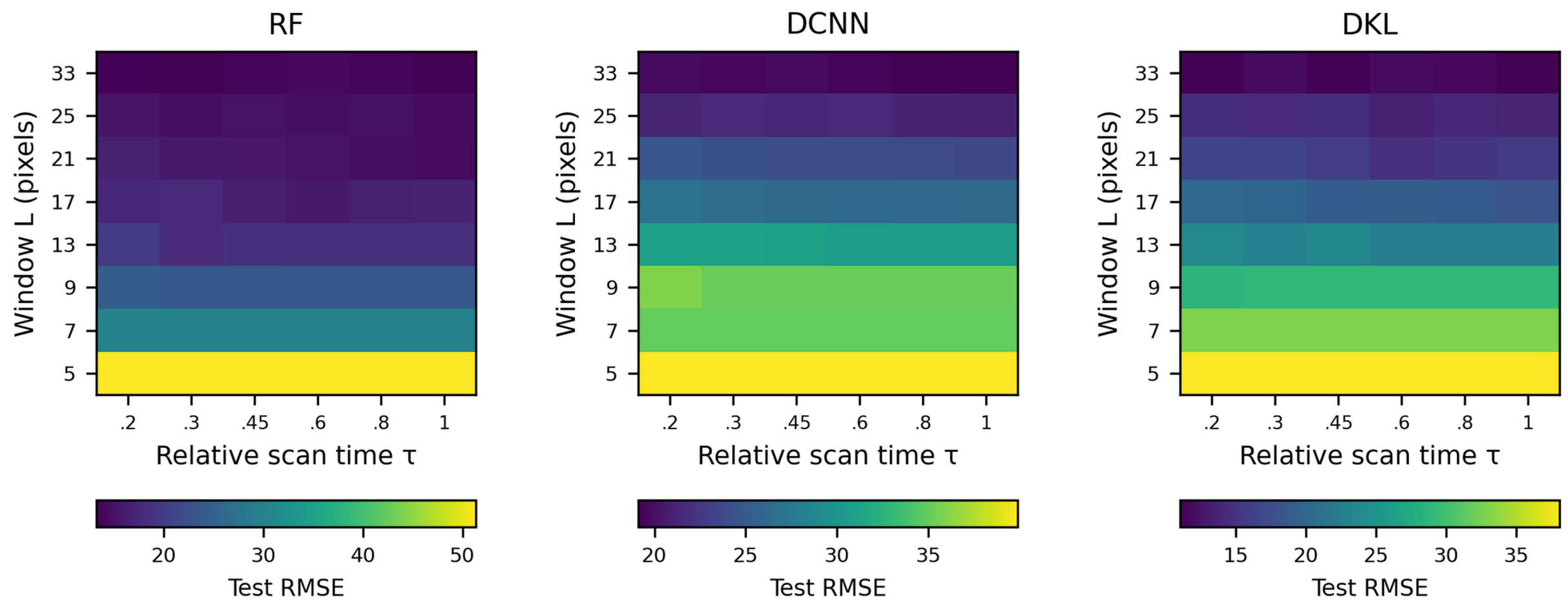

**(a)** **(b)** **(c)**

**Figure S16.** Size-independent-cost sensitivity study. (a) RF matched-condition test RMSE under the revised law, averaged over two fits at every setting. Calibration data, not this test surface, select reported settings. (b) DCNN matched-condition test RMSE under the revised law. Wider context may remain useful despite a larger Gaussian bandwidth. (c) DKL matched-condition test RMSE. The posterior is conditioned on the actual blurred images; no hidden clean descriptor is substituted for an acquired input.

**Table S9. Revised matched-condition full-data configurations for the fixed primary profiles.**

| Model | New L / τ | New σ | New RMSE | Old selected RMSE | CPU / fit |
|---|---|---|---|---|---|
| RF | 25 / 1 | 3.11 | 14.39 | 14.15 | 0.917 |
| DCNN | 33 / 0.8 | 5.56 | 19.10 | 18.66 | 2.021 |
| DKL | 33 / 0.45 | 7.84 | 11.16 | 11.86 | 7.761 |

The selected windows remain 25 pixels for RF and 33 pixels for DCNN and DKL. Increased blur therefore does not necessarily eliminate the predictive value of broader context. This interpretation remains subject to reflection at scene boundaries and the fixed-canvas coupling between window size and active descriptor cells. The acquisition-cost frontier uses one-image cost on its horizontal axis at a fixed full-data fit; it is not the same as the computational frontier or a complete-campaign budget frontier.

### S8.3 Paired sequential results and uncertainty

The revised study trains 144 policies/campaigns under the same replication schedule and evaluates 180 primary-profile RL/baseline campaigns, for 324 new recorded trajectories. Alternative-profile comparisons are not added. Each primary RL row has twelve deployments from the six experimental initializations and two policy seeds. Paired intervals first average the policy replicates within each initialization; they summarize variability within this specimen.

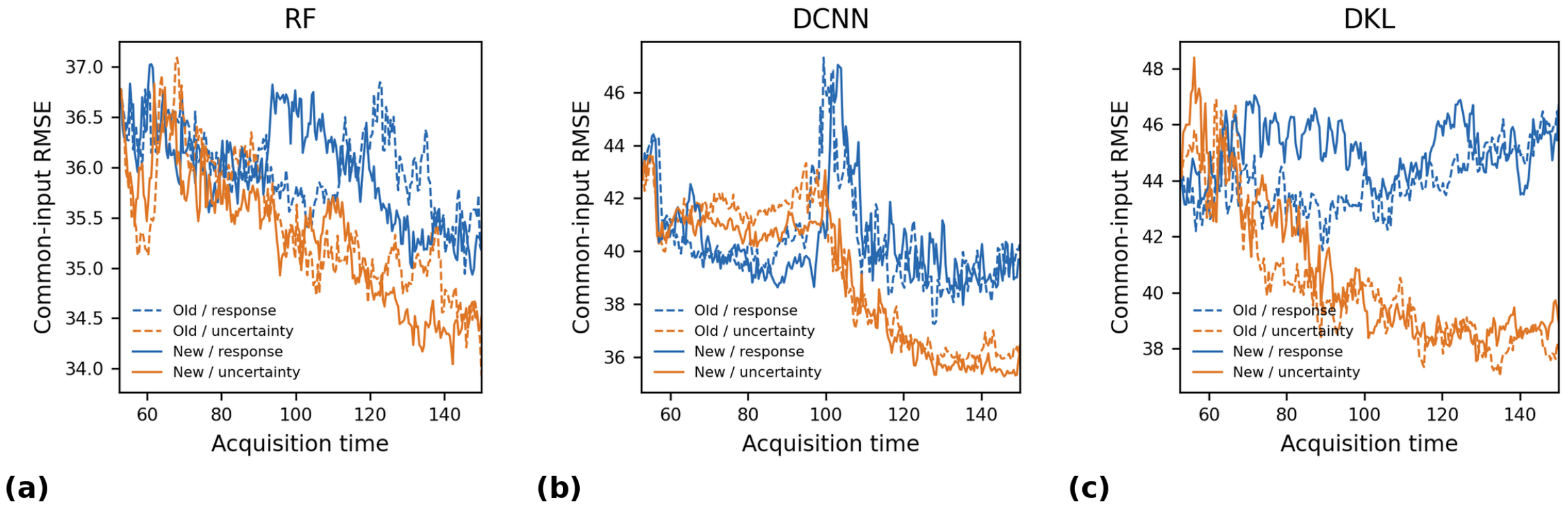

**Figure S17.** Size-independent-cost sensitivity study. (a) RF common-input RMSE versus accumulated acquisition time. Dashed curves are the original scenario; solid curves use the new law. Curves are means of 12 RL deployments. (b) DCNN evaluated on the identical clean-17 input panel. A lower curve therefore cannot be attributed to making the audit images easier or more blurred. (c) DKL common-input prediction error. A better full-data fit or smaller GP variance does not automatically imply a better finite-budget online model.

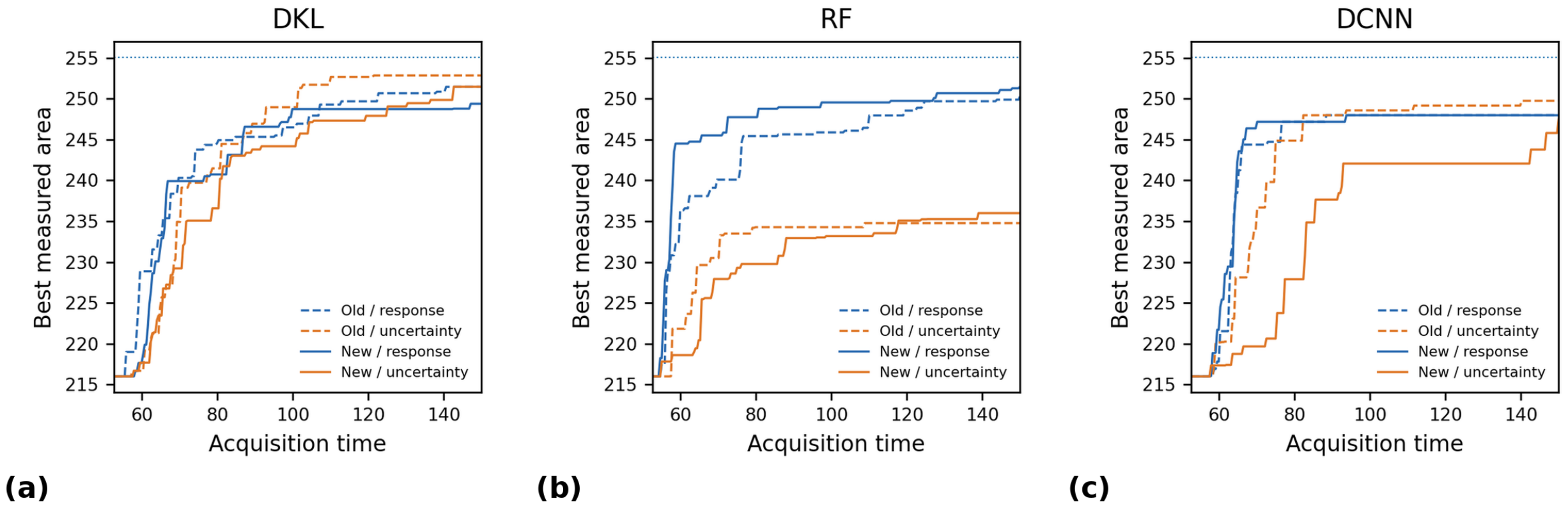

**Figure S18.** Size-independent-cost sensitivity study. (a) DKL best actually measured response. The discovery oracle is the unchanged accessible maximum; predicted high values do not count as discoveries. (b) RF discovery trajectories for the two objectives and two physical models. (c) DCNN discovery trajectories. All curves are evaluated against the same response values and time budget.

**Table S10. Revised minus original scenario: paired endpoint differences and intervals.**

| Model | Objective | Δ RMSE | Paired 95% interval | Δ best area | Δ spectra |
|---|---|---|---|---|---|
| RF | exploitation | 0.08 | [-1.51, 1.67] | 1.19 | -6.25 |
| RF | exploration | 0.56 | [-0.64, 1.76] | 1.22 | -5.33 |
| DCNN | exploitation | -0.64 | [-5.11, 3.83] | 0.00 | -4.67 |
| DCNN | exploration | -0.28 | [-2.15, 1.60] | -2.26 | -3.33 |
| DKL | exploitation | 1.17 | [-4.03, 6.38] | -2.07 | -4.00 |
| DKL | exploration | 0.88 | [-1.51, 3.27] | -1.38 | -4.50 |

Every paired RMSE interval includes zero. The revised law therefore produces measured changes in behavior without establishing a general prediction improvement. RF exploitation increases exact-maximum discoveries from 2/12 to 4/12, whereas DKL exploration decreases from 8/12 to 6/12. The accessible maximum remains 255.0 because the underlying spectra have not changed.

Simpler controllers remain competitive. In the revised DKL exploration experiment the acquisition-rule baseline reaches mean best area 253.85, compared with 251.46 for RL. For revised DCNN exploration the corresponding values are 249.15 and 247.47. These outcomes should not be hidden behind comparisons of internal rewards alone.

### S8.4 Operation allocation and acquired fidelity

Figure S19 compares the fixed family-reference prediction gaps. Figure S20 records labelled-data accumulation. Figure S21 displays the representative DKL exploration window, IMAGE duration, and blur; duration now equals $\tau$ rather than $(L/17)\tau$. Figure S22 shows its explicit IMAGE/SPECTRUM sequence. Making large windows cheaper does not make every action cheaper: windows smaller than 17 pixels lose their original size discount. The revised RL campaigns acquire approximately three to six fewer spectra on average, depending on family and objective. In both scenarios, the mean selected window and scan time (Table S11) stay close to the averages over the action set (16.3 pixels and 0.56), consistent with the limited shift in mean settings discussed in main Section 5.2; the change in observation law alters the acquired blur without a large shift in these averages; mean settings alone do not determine the action distribution.

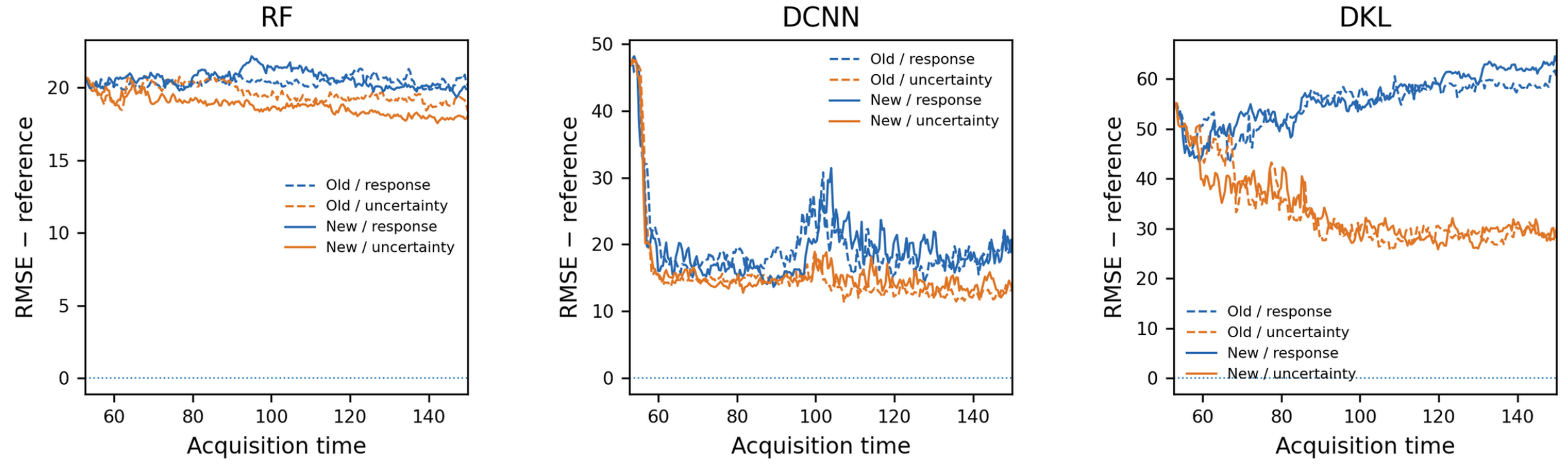

**(a)** **(b)** **(c)**

**Figure S19.** Size-independent-cost sensitivity study. (a) RF error on the original family-oracle input panel minus the fixed family reference. Both the input panel and the oracle value are held unchanged between scenarios. (b) DCNN global-reference gap. This is different from error on the common clean-17 panel; the two audits answer different questions. (c) DKL approach to its fixed

global reference. Stronger blur in newly acquired training images does not alter the evaluation image definition.

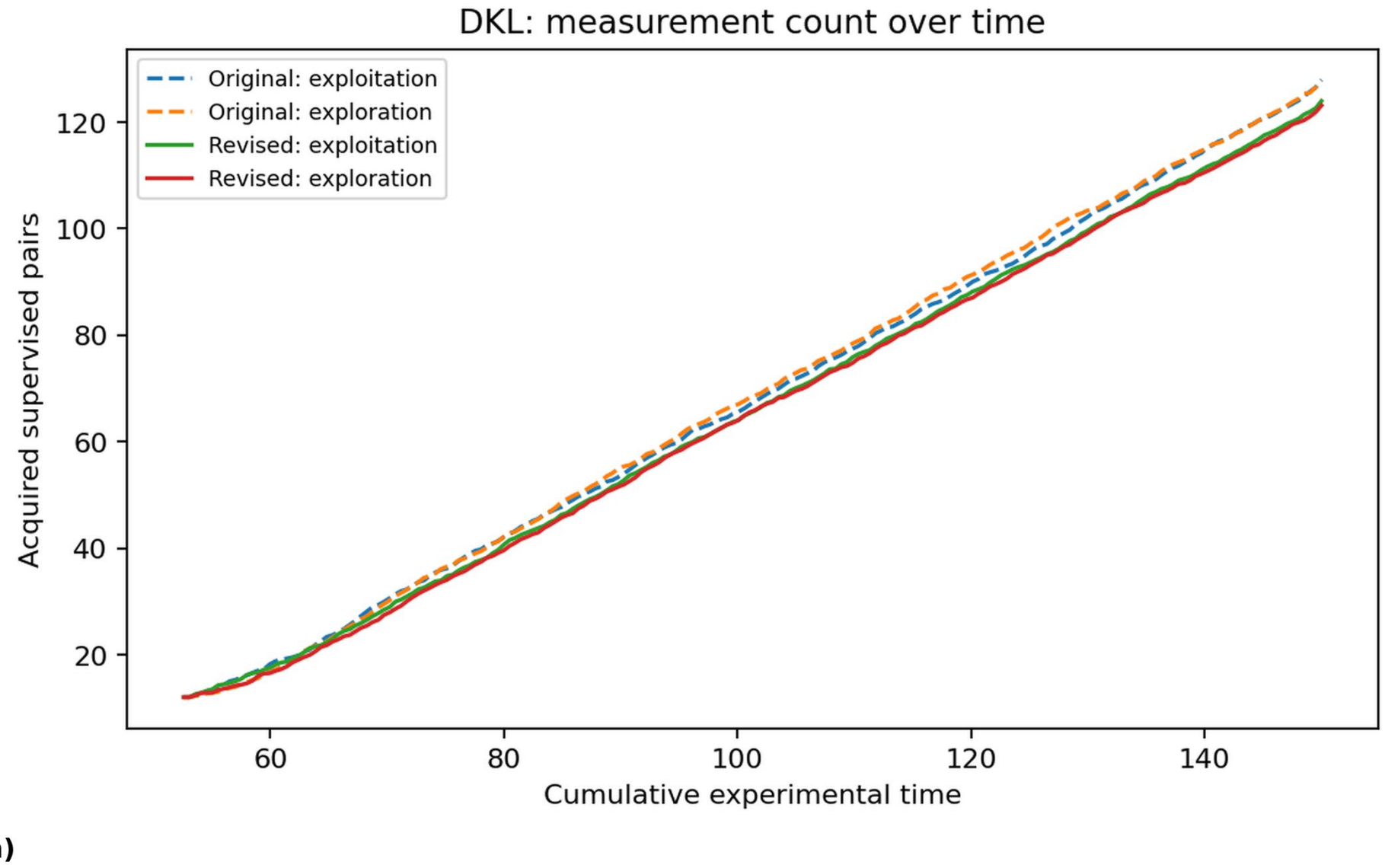

**Figure S20.** Size-independent-cost sensitivity study. Number of supervised pairs versus cumulative acquisition time. Only spectroscopy increases the count; image-only steps leave it unchanged.

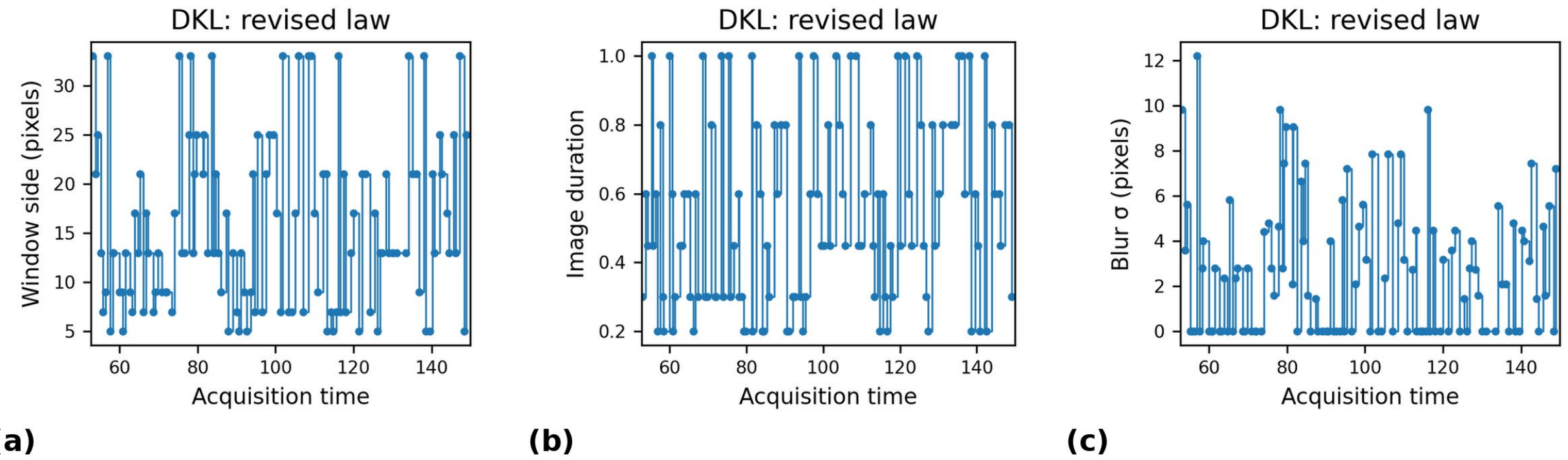

**Figure S21.** Size-independent-cost sensitivity study. (a) Representative DKL exploration window sequence, initialization 1001 and policy seed 11. The window choice is explicit, while the candidate is ranked by the acquisition rule. (b) Image acquisition duration versus cumulative experimental time. An image of any side length has exactly this time cost. (c) Actual acquired Gaussian bandwidth versus cumulative time for the same campaign. It changes jointly with window size and selected scan duration, rather than being determined by duration alone.

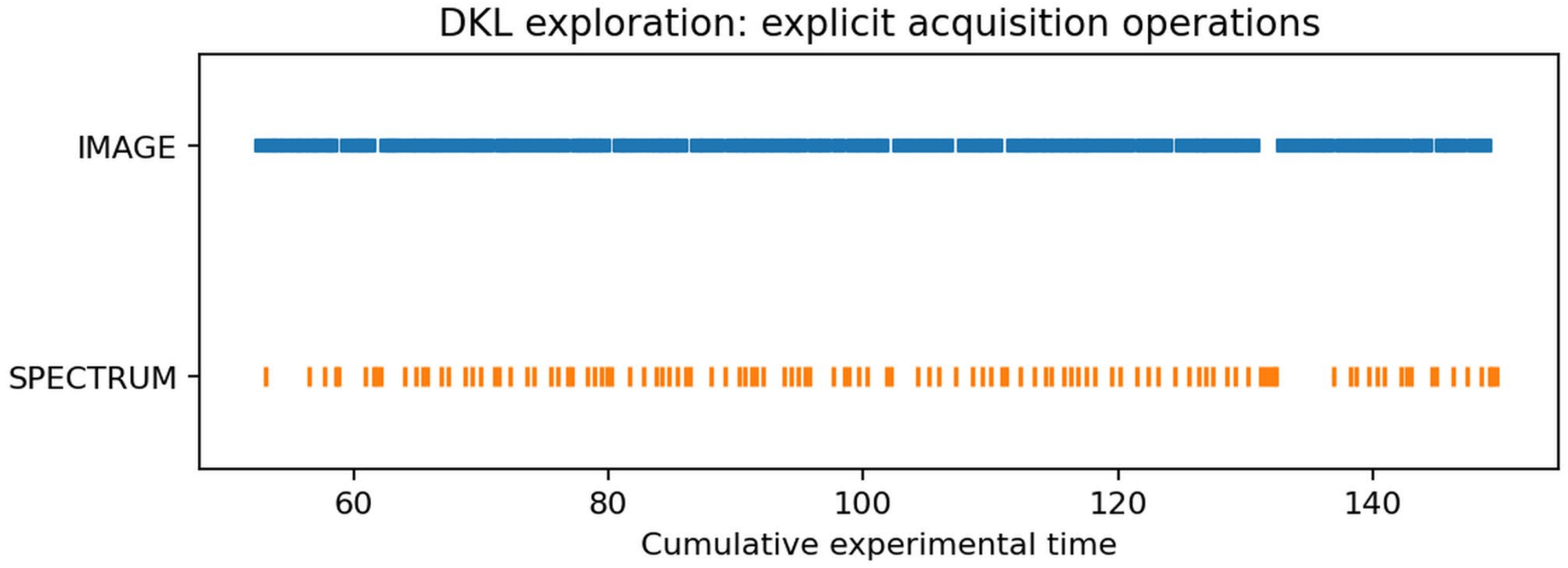

**(a)**

**Figure S22.** Size-independent-cost sensitivity study. IMAGE and SPECTRUM decisions for the representative DKL exploration campaign. Each spectrum is acquired at a previously imaged, unlabelled center; no automatic image–spectrum batch is imposed.

**Table S11. Mean selected window, relative time, and native-pixel blur under the revised law.**

| Model | Objective | Mean window side | Mean selected time | Mean blur |
|---|---|---|---|---|
| RF | exploitation | 15.8 → 16.0 | 0.549 → 0.552 | 2.93 → 2.71 |
| RF | exploration | 16.0 → 15.9 | 0.557 → 0.564 | 2.87 → 2.66 |
| DCNN | exploitation | 15.8 → 16.0 | 0.549 → 0.548 | 2.92 → 2.72 |
| DCNN | exploration | 16.0 → 16.0 | 0.541 → 0.539 | 2.97 → 2.81 |
| DKL | exploitation | 15.9 → 16.0 | 0.547 → 0.538 | 2.95 → 2.78 |
| DKL | exploration | 15.8 → 15.9 | 0.545 → 0.544 | 2.96 → 2.67 |

### S8.5 Reward alignment under the revised law

Figure S23 presents realized exploration rewards and fixed-panel uncertainty. The revised Gaussian response does not make reduced model disagreement equivalent to lower error. Table S12 retains both raw uncertainty gain and maturity-weighted reward correlations. Its positive-gain/worse-error statistic is a pooled action-level ratio, which should not be confused with the mean-of-campaign ratios quoted for the primary experiment in Section S7.

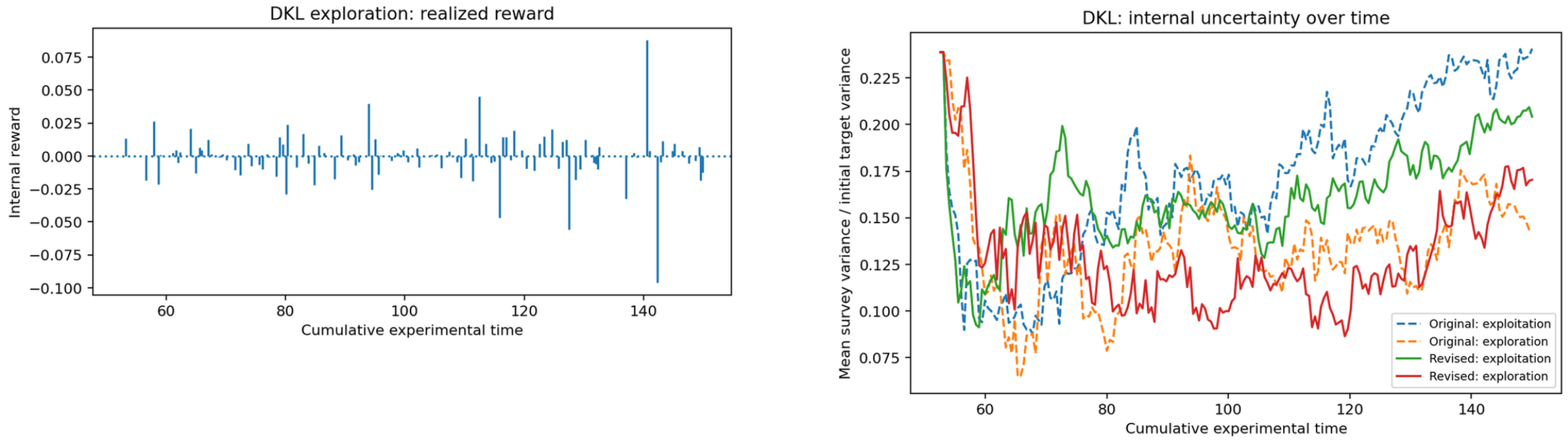

**(a)** **(b)**

**Figure S23.** Size-independent-cost sensitivity study. (a) Realized exploration reward. Images do not directly reduce fixed-survey uncertainty in this implementation; their value is delayed until

spectroscopy and supervised model updating. Negative rewards are retained. (b) Internal survey uncertainty for DKL. This curve is a reward-related quantity, not a substitute for the prediction-error and discovery curves.

**Table S12. Exploration reward audit before and after the observation-law change.**

| Model | Scenario | Raw-U rank correlation | Weighted-reward rank correlation | Positive gain but worse error |
|---|---|---|---|---|
| RF | Original | -0.009 | -0.013 | 49.9% |
| RF | Revised | -0.087 | -0.096 | 50.3% |
| DCNN | Original | -0.527 | -0.536 | 70.2% |
| DCNN | Revised | -0.497 | -0.494 | 70.6% |
| DKL | Original | 0.037 | 0.038 | 48.0% |
| DKL | Revised | 0.024 | 0.027 | 46.7% |

The revised study reports all 324 acquisition traces passing its budget, access, blur, and reward-arithmetic checks, and all 180 saved evaluation predictors reproducing their recorded endpoints. The recorded main computation consumed approximately 25.3 minutes elapsed and 1.61 aggregate CPU-hours across parallel workers. These are computational measurements, not simulated acquisition time. The controlled comparison demonstrates an explicit context–resolution trade-off; it is not validation of the assumed scan-speed law on a live microscope.

## S9. Verification, implementation state, and figure provenance

This section lists the controller state, the verification checks performed on the archived campaigns, the provenance of the figures, and the main limitations of the first campaign.

### S9.1 Observable state and reproducibility

The original workflow exposes the following normalized state components to its linear actor and critic. Their definitions establish what the controller can use; unrevealed responses and evaluator-only errors do not enter this vector.

**Table S13. Twenty-component controller state.**

| Index | Component |
|---|---|
| 1 | Constant 1 |
| 2 | Remaining budget / 150 |
| 3 | Elapsed budget fraction |
| 4 | Number of labelled pairs / 300 |
| 5 | New IMAGE count / 300 |
| 6 | New SPECTRUM count / 300 |
| 7 | Pending-image count / 8 |
| 8 | Survey U clipped to [0,5], divided by 5 |
| 9 | $N / (N + N_0)$ |

| Index | Component |
|---|---|
| 10 | (Best response − initial mean) / initial scale, clipped to [−5,5], then /5 |
| 11 | Maximum acquisition score over pending centers, clipped to [−5,5], then /5 |
| 12 | Maximum acquisition score over unseen centers, clipped to [−5,5], then /5 |
| 13 | Unseen candidate fraction |
| 14 | Indicator for exploration objective |
| 15 | Last realized reward clipped to [−1,1] |
| 16 | Mean labelled-center window side / 33 |
| 17 | Mean labelled-center relative scan time |
| 18 | Indicator that a pending center exists |
| 19 | Indicator that an unseen center exists |
| 20 | log(1+N) / 6 |

The original RF/DCNN and DKL computations occurred in separate sessions. Their records are integrated here with unchanged targets, splits, action conventions, and profile-selection rules. All 360 original acquisition traces passed the archived accounting audit, and all 216 original evaluation models reproduced their saved results within documented numerical tolerances. The revised acquisition-law study is additional and is not counted as part of those 360 traces.

**Table S14. Verification recorded for the joint RF/DCNN/DKL workflow.**

| Check | Recorded outcome | Scope |
|---|---|---|
| Real-image reconstruction | All 2,079 pixels covered; zero patch disagreement | The working image matches the cached observations |
| Fresh joint trace audit | 360 / 360 passed | Recomputed legal operation order, costs, label increments, rewards and budget |
| Archived final-model reproduction | 216 / 216 passed | Saved predictors reproduce recorded outputs within numerical tolerances |
| Exact-GP likelihood gradient | Agreement with Cholesky autodiff to the documented tolerance | Checks the implemented conditional likelihood calculation |
| DKL posterior and reload | Direct algebra and exact reload checks passed | Checks mean/variance computation and serialization |
| Hidden-label perturbation | Observable state/ranking and IMAGE reward unchanged | Checks the intended online information boundary |
| Source parity audit | Environment/controller/action routines unchanged across the DKL addition | Supports the same-parameter comparison |

### S9.2 Figure provenance

All figures are assembled from the stored experimental image, the numerical calibration tables, and the recorded campaigns; no predictive model or policy was retrained to prepare them. Display offsets in the control-space plots (Figure 6) do not alter any acquisition setting or metric. The 0.8–1.0 eV map and neighbor-mean gap filling in Figure 1 are computed for display only; missing spectroscopy remains excluded from all analyses. Figures 8 and 11 use the supplied replacement panels, with standardized panel labels. The remaining drift-campaign figures are reformatted from the archived plots; hatching in Figure S24 marks source pixels hidden by the original colorbar, and no markers are reconstructed there.

### S9.3 Limits and deferred extensions

The dataset is one small specimen with spatially overlapping training/test neighborhoods. Large windows use reflected boundaries. Physical context and numerical descriptor support are coupled. The coarse preview, fixed reference panel, initial seed data, and offline calibration define an explicit information and cost contract, not a universal operating condition. Compute time is measured but not deducted from acquisition time. The spectrum cost is fixed and no extra noise or dose process is introduced. RL reward scaling and clipping act differently on differently scaled uncertainty estimators. These restrictions limit the conclusions to the tested workflow and prevent treating the campaign count as specimen-level replication.

Future work could calibrate the scan-speed/image-quality response on the actual instrument, incorporate measured computation into the critical path, separate context from numerical resolution, calibrate uncertainty prospectively, and test transfer to independent specimens. Newly acquired unlabelled images might also update a structural manifold before spectroscopy. That image-only learning extension remains explicitly deferred: it would require a different representation-learning component, separate representation maturity, and a revised credit-assignment analysis. It is not performed by the present supervised DKL encoder merely because that encoder is neural.

### S9.4 Source records and code availability

The evidence consists of the joint RF/DCNN/DKL workflow archive, the size-independent-cost archive, and the drift-campaign archive, each with its executed notebook, together with the figure-generation code and the numerical records used for plotting. Training checkpoints and all campaign traces are retained in these archives. A public repository identifier will be provided upon publication.

## S10. Drift, scan-speed blur, calibration, and full-campaign metrology

This section documents the second campaign, reported in main Section 6. It is separate from the fixed-registration first campaign ($B$ = 150) with RF, DCNN, and DKL. Its structural input and spectral labels originate in the same recovered dataset, while drift dynamics, blur, cross-correlation registration, operation costs, and controller decisions are simulated. Results from the static test, speed-only baseline, and 300-unit calibration-control replay are not pooled into a single performance estimate because their actions, metrics, and budgets differ.

### S10.1 Static oracle analysis of feature drift

The reconstructed structural field has 33 × 63 native pixels. The saved cache supplies measured scalar band responses and 100-channel EELS spectra at 1,963 locations; 116 locations lack cached spectroscopy labels. Clean image patches of fixed side 17 pixels were used to fit a two-dimensional convolutional autoencoder. In a static counterfactual test, the structural feature center is shifted by a prescribed vector while the EELS target remains that of the intended center. This isolates feature–target misregistration without asserting that a live instrument followed the simulated trajectory. Fractional-pixel shifts use interpolation of the image. Near edges, reflected image content is used; the 15-pixel sweep therefore measures the specified boundary-padded simulator, not an independent larger specimen.

The latent mixing coefficient uses differences in measured band response between neighboring embedded points, normalized by twice the global target variance. Population spreading is the RMS distance between a clean embedding and embeddings after displacement in multiple directions. Both are representation-dependent measurements; the first changes with latent neighbor topology, whereas the second remains tied to each original point. The frozen autoencoder makes scores comparable across radii. The bottleneck is a visualization device [68], not an assumed isometric representation.

Table S15. Selected observed static drift metrics from the real EELS cache.

| Drift (px) | kNN mixing | Mean population spread | Population p90 |
|---|---|---|---|
| 0 | 0.214 | 0.000 | 0.000 |
| 1 | 0.213 | 0.141 | 0.309 |
| 2 | 0.239 | 0.279 | 0.638 |
| 5 | 0.290 | 0.674 | 1.304 |
| 8 | 0.329 | 0.990 | 1.687 |
| 10 | 0.313 | 1.164 | 1.942 |
| 12 | 0.447 | 1.302 | 2.121 |
| 15 | 0.590 | 1.449 | 2.319 |

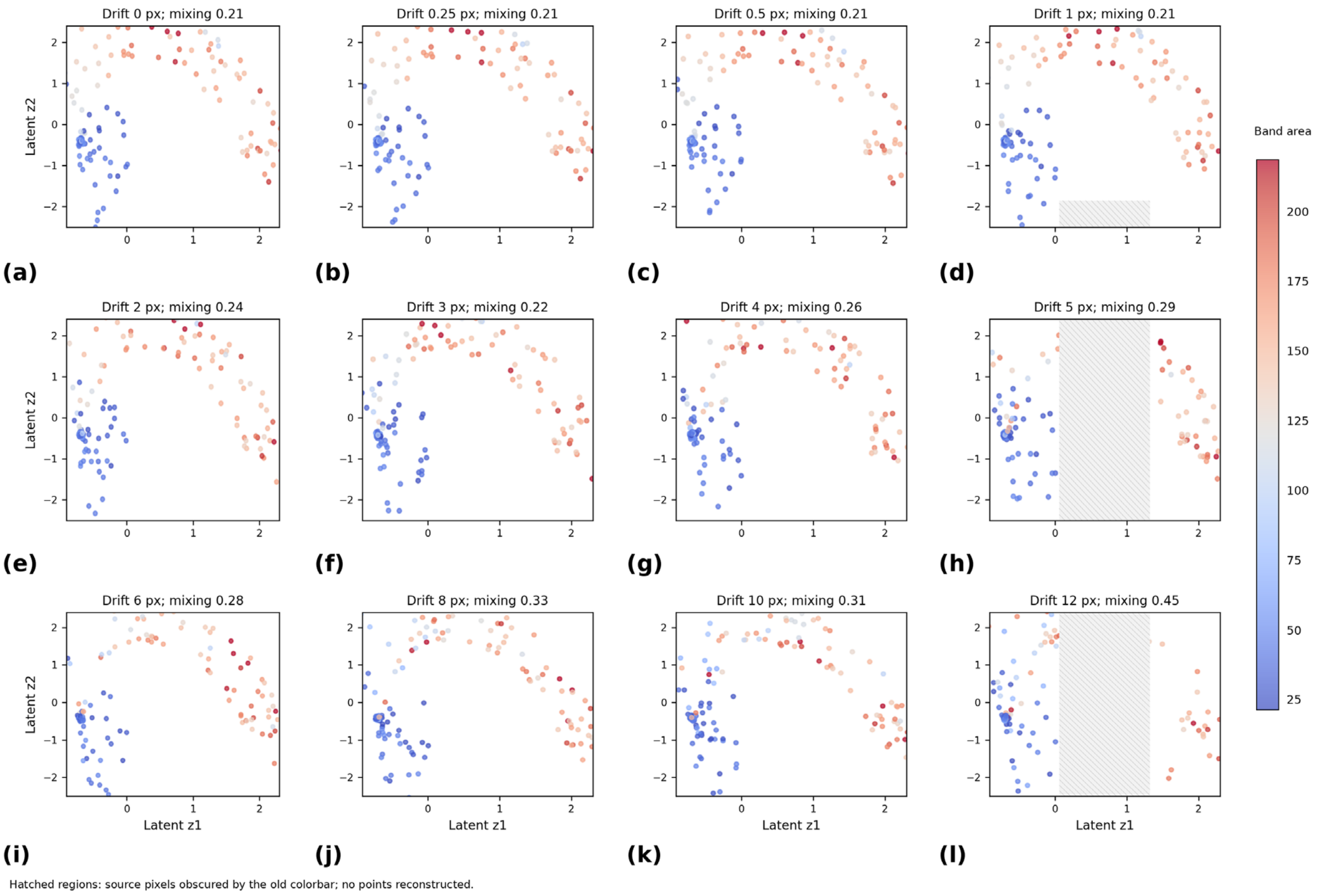

Figure S24. Latent distributions at twelve displacement radii, arranged in three rows and four columns with one external response color scale. The 15-pixel display panel is omitted; its numerical results remain in Tables S15–S16. Point colors retain the measured target at the nominal coordinate. Hatching at 1, 5, and 12 pixels marks regions obscured by the original raster colorbar; unavailable points are not reconstructed. The 0- and 10-pixel conditions are also shown in Figure 8b,c.

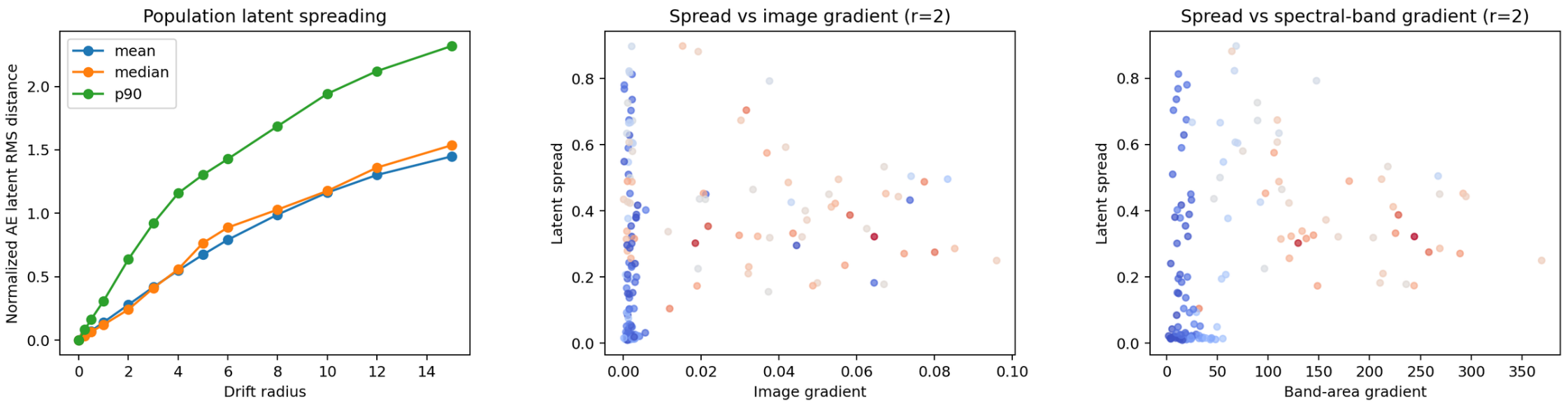

**(a)** **(b)** **(c)**

Figure S25. Population-average, median, and upper-tail latent spreading, with auxiliary correlations to image gradient and target gradient. Gradient comparisons are retrospective oracle diagnostics; missing target pixels are filled only for the auxiliary spatial-gradient display. The population-spreading panel is reproduced as main Figure 8d.

Model sensitivity was tested by fitting Ridge, RF, CNN, and DKL to clean structural patches and predicting on displaced held-out patches, with unchanged spectral targets. The source training and testing subsets were kept disjoint by center, while spatially overlapping patches were allowed. The 100-channel cached spectrum and scalar band intensity were evaluated separately. The tested DKL training schedule and sample count differ from those of the other regressors; model rankings should be interpreted as achieved configuration performance, not matched-compute architecture limits.

Table S16. Selected NRMSE under clean training and drifted-feature evaluation.

| Target | Model | 0 px | 5 px | 10 px | 15 px |
|---|---|---|---|---|---|
| Band | Ridge | 0.459 | 0.571 | 0.636 | 0.855 |
| Band | RF | 0.276 | 0.526 | 0.616 | 0.803 |
| Band | CNN | 0.434 | 0.528 | 0.621 | 0.828 |
| Band | DKL | 0.441 | 0.583 | 0.677 | 0.834 |
| Spectrum | Ridge | 0.324 | 0.382 | 0.440 | 0.550 |
| Spectrum | RF | 0.245 | 0.369 | 0.435 | 0.535 |
| Spectrum | CNN | 0.328 | 0.370 | 0.431 | 0.532 |
| Spectrum | DKL | 0.440 | 0.446 | 0.478 | 0.557 |

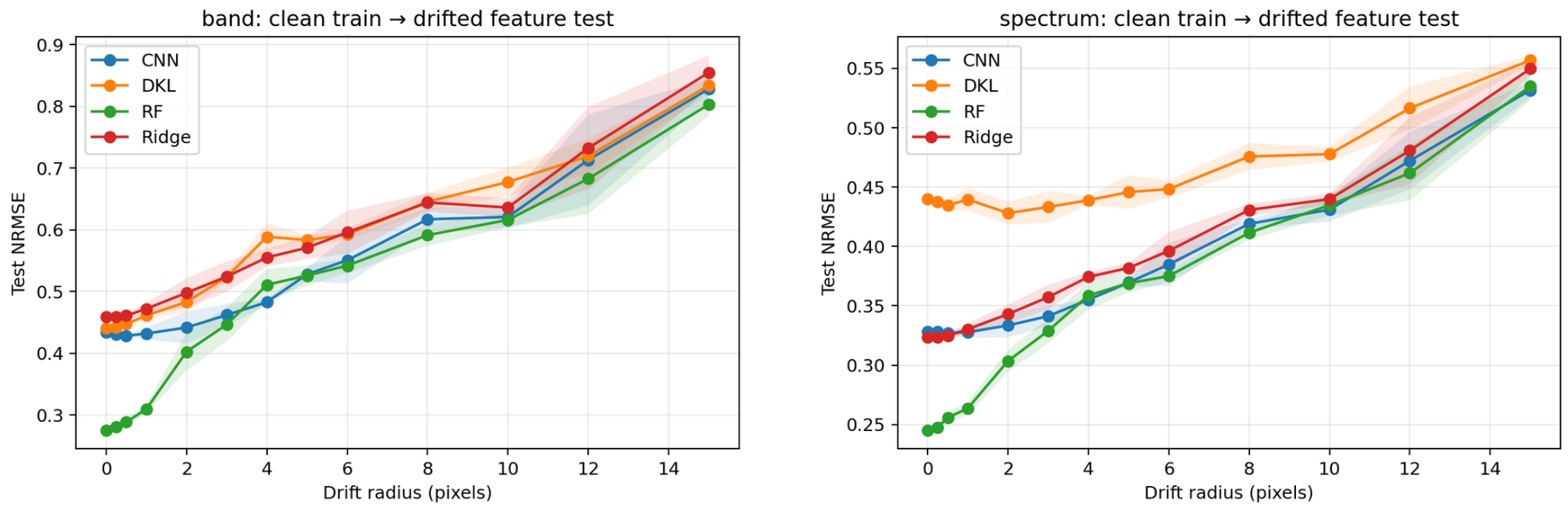

**(a)** **(b)**

Figure S26. Model-specific normalized test error as structural patches are displaced relative to their original EELS labels. The left panel reports integrated-band prediction; the right reports prediction of the cached 100-channel spectra. Curves retain the executed model settings and repeated displacement realizations. The scalar band-response regression panel is reproduced as main Figure 8e.

### S10.2 Scan-speed baseline and competing fidelity mechanisms

Before allowing calibration, IMAGE and SPECTRUM were paired at a common nominated target, with scan time chosen from six relative durations. The inherited phenomenological blur law is $\sigma(\tau) = [(1/\tau - 1)/0.12]^{1/2}$ native pixels for $\tau \in \{0.2, 0.3, 0.45, 0.6, 0.8, 1.0\}$; blur is directional along the raster fast-scan axis. Fast imaging reduces additional elapsed time but does not erase drift already accumulated. The baseline simulation compares fixed speeds, repeated imaging, and a simple confidence-triggered reimage heuristic under matched drift realizations and a fixed campaign budget. It does not contain true cross-correlation position correction. The effect of speed is therefore a compromise between structural information lost to blur and spectroscopy capacity gained through shorter image acquisitions.

The scan-speed-only study used image cost proportional to relative scan duration and a separate spectroscopy charge, whereas the later explicit calibration experiment uses image cost $3\tau$, a calibration charge of 1, and spectroscopy charge of 1. Scores from these different resource contracts are not placed on the same numerical cost axis. Across speed-only runs a fast image frequently improves response discovery by enabling more spectra, while a longer scan can improve the quality of learned image–spectrum relationships. Neither outcome establishes a single universally best instrument setting.

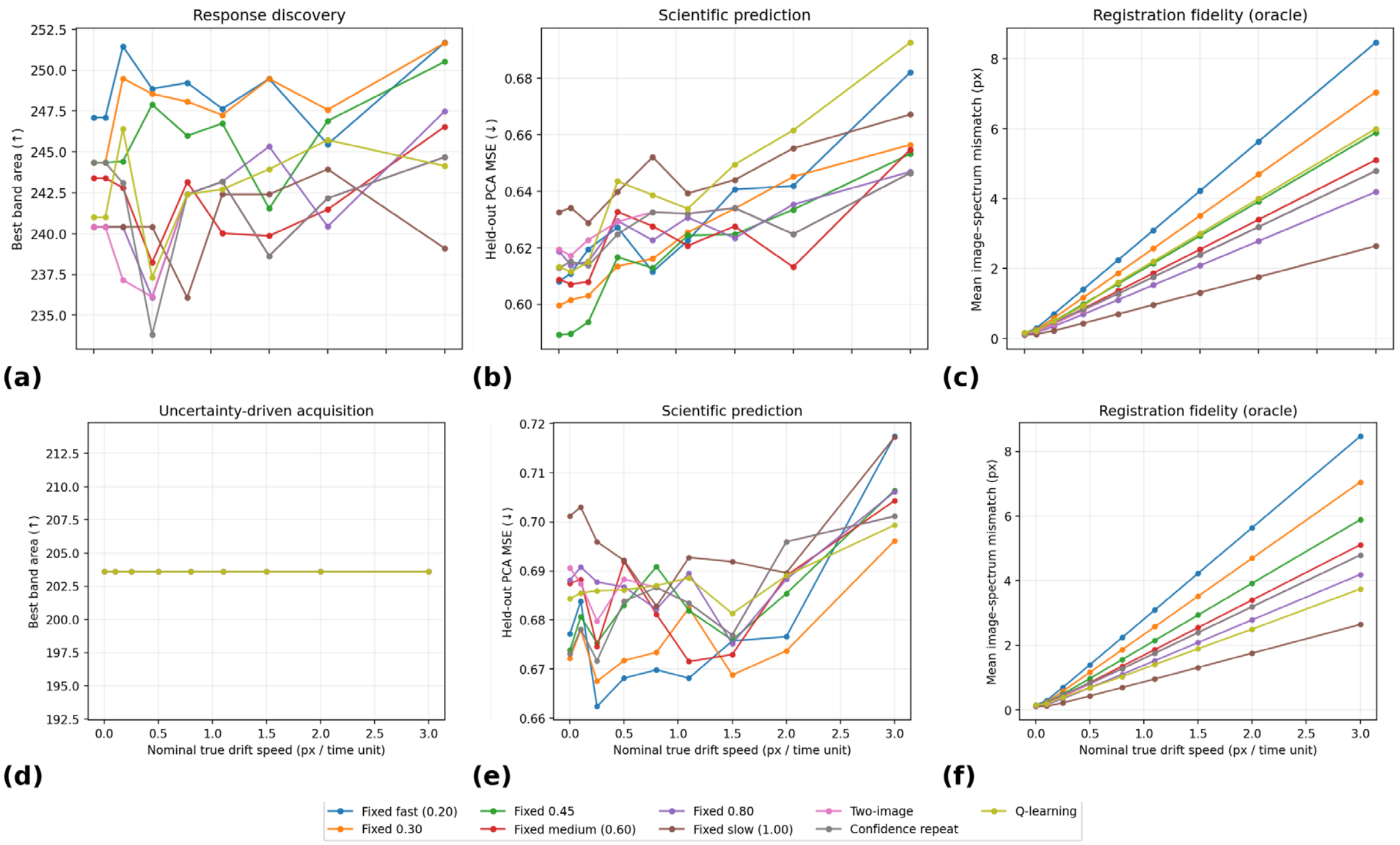

Figure S27. Fixed-scan baseline results over the simulated drift-speed sweep, separating response discovery, scientific prediction, and physical measurement quality. This scan-speed-only study uses a different campaign budget from the $B = 300$ explicit-calibration benchmark.

### S10.3 Eight-action instrument environment and cross-correlation

The full-control environment retains a 17 × 17 structural window and eight legal operations: IMAGE at $\tau$ = 0.2, 0.3, 0.45, 0.6, 0.8, or 1.0; CALIBRATE; and SPECTRUM. IMAGE costs $3\tau$, CALIBRATE costs 1, and SPECTRUM costs 1 normalized acquisition-time unit. IMAGE produces a drifted and directionally blurred observation, but does not automatically correct a position or reveal a spectral label. CALIBRATE operates on an available image by normalized cross-correlation with a trusted reference, estimates the shift, and modifies the commanded position. A corrected target must be imaged anew before spectroscopy. SPECTRUM returns one preacquired experimental response at the simulated actual beam position and updates the scientific learner. All operations advance the hidden drift process and consume budget; the initial acquisition and reference resources are included under the executed ledger.

Cross-correlation quality is measured on the simulated blurred image, not assigned solely by a convenient noise formula. The calibrated image produces the shift estimate and signal quality indicators, whereas only the evaluator sees the true displacement and registration error. A fast scan may lead to a broad or ambiguous cross-correlation peak. The experiment tested constant-velocity/random drift, linear drift, and linear-plus-periodic drift, including separate held-out motion realizations for frozen policy evaluation. The structured K2 estimator uses candidate periodic components; its hypothesis set contains the simulated periodicity and is therefore a favorable structured-drift comparison. Registration and drift-correction methods motivate independent validation of this observation channel [63–65,69].

Table S17. Assumed full-control action costs and scan-dependent Gaussian blur.

| IMAGE τ | Time cost | Blur σ (px) |
|---|---|---|
| 0.20 | 0.60 | 5.774 |
| 0.30 | 0.90 | 4.410 |
| 0.45 | 1.35 | 3.191 |
| 0.60 | 1.80 | 2.357 |
| 0.80 | 2.40 | 1.443 |
| 1.00 | 3.00 | 0.000 |
| CALIBRATE | 1.00 | N/A |
| SPECTRUM | 1.00 | N/A |

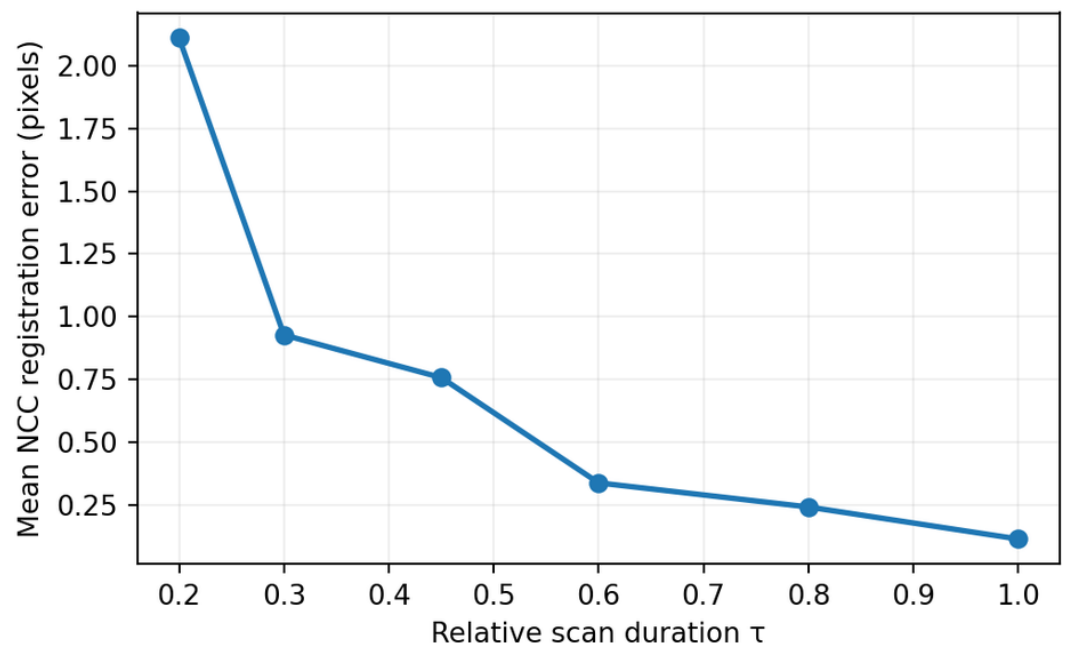

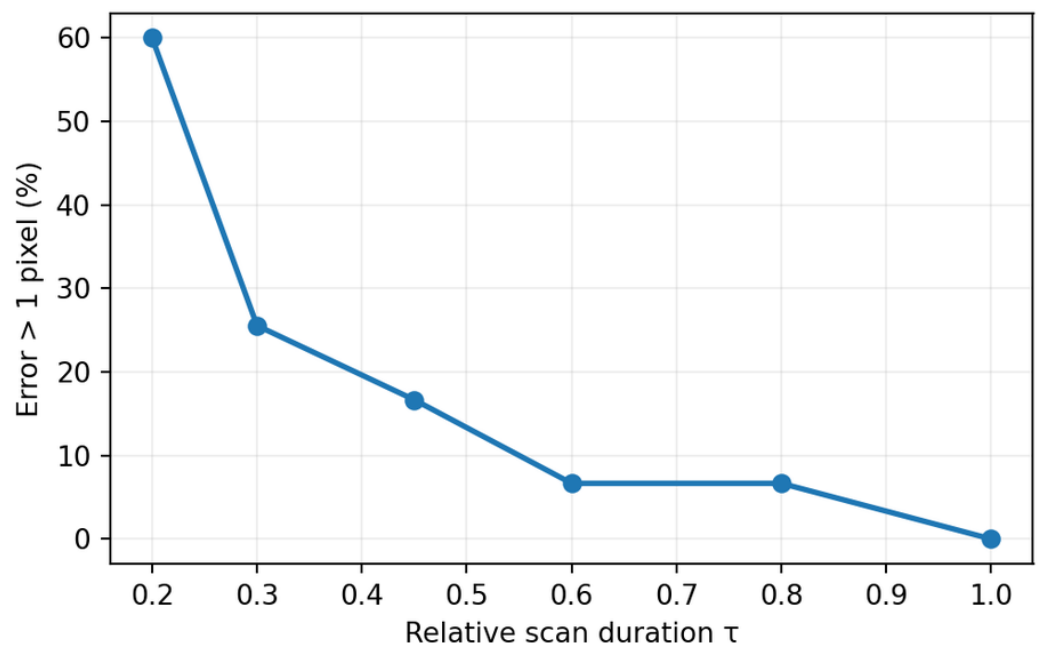

Figure S28. Cross-correlation registration accuracy and failure rate versus image acquisition speed, measured using known applied shifts in the reconstructed real structural image. At high blur, correlation-derived displacement can be substantially less accurate.

### S10.4 Control strategies, rewards, and instrument beliefs

The model-free controller uses Q-learning with discretized observations (budget remaining, time since calibration, acquired-image status, model confidence and other legally available history), without an explicit Kalman dynamics model. Its operation set is the same as the calibration controllers. K1 is a constant-velocity Kalman state estimator for the two displacement coordinates and their velocities. K2 augments linear motion with a bank of periodic candidate components. A filter is not itself an action policy: both generic and structured Kalman estimators require either a periodic calibration schedule or a covariance-triggered decision rule. To isolate estimation quality from decision scheduling, K1 and K2 are also evaluated under identical periodic calibration times and input observations. Raw periodic and fixed-speed rules are additional controls. The action-log audit of the $B$ = 300 experiment confirms that model-free RL selected explicit calibration actions in every held-out evaluation campaign. Across 30 matched drift realizations, the response-maximization RL policy used a mean of 10.07 calibrations per campaign (range 9–12), whereas the uncertainty-reduction RL policy used a mean of 4.70 (range 2–11). The corresponding mean numbers of acquired spectra were 115.3 and 90.0. Calibration is thus sparse but present; low calibration counts should not be mistaken for an action mask or missing code branch. The logs contain only model-free RL and independent Kalman policies; an RL policy explicitly conditioned on a K1/K2 posterior was not trained. The algorithmic references are Q-learning [70] and Kalman filtering [71]; the periodic-component bank is a problem-specific extension.

Exploitation rewards the realized increase in the best measured band response relative to the initial record. This reward is zero for IMAGE and CALIBRATE and nonnegative upon spectroscopy. Exploration instead receives the pre-calibrated learner-maturity factor $N/(N + N_0)$ multiplied by the observed reduction in RF inter-tree variance on a fixed, accessible survey panel after a spectrum. The maturity weight is a heuristic from offline learning curves, not a posterior probability or proof that confidence tracks accuracy. A fixed total acquisition budget incorporates time cost directly; the hidden evaluator measures true spectral PCA loss, physical displacement, relative image–spectrum mismatch, observed response, acquisition counts, and fraction of available response improvement realized.

Crucially, the accessible maximum measured band response is a retrospective discovery ceiling rather than a budget-limited optimal policy value. Similarly, a clean full-data RF fit provides an empirical reference error and not an intrinsic Bayes limit. The original action logs do not contain a separately trained RL-plus-Kalman hybrid, nor do they log every per-action Kalman posterior and actual RF uncertainty reduction needed to reconstruct an exact clairvoyant controller. We therefore compare time-resolved policies with the full-field response ceiling, clean-data reference, and an empirical best-of-tested-policy envelope, explicitly distinguishing all three.

### S10.5 Full B = 300 campaign results and oracle proximity

The full campaign contains 540 frozen-policy evaluations: nine controllers, two separately trained scientific objectives, ten matched drift realizations in each of three motion regimes, and budget $B$ = 300 for every campaign. Position correction requires a fresh image before spectroscopy. The corresponding model-free discovery policy uses measured record improvement, while the exploration policy uses measured RF variance reduction. The evaluated controllers are not RL policies all trained jointly; K1 and K2 use fixed or triggered drift estimators and schedule rules. The action ledger, hidden losses and acquired measurement counts are retained for each run.

Table S18. Representative campaign means for the 300-unit response-maximization objective.

| Policy | Band record | Hidden PC MSE | Spectra | Calibrations |
|---|---|---|---|---|
| Fixed fast | 246.50 | 0.714 | 152.0 | 0.0 |
| Raw periodic | 241.81 | 0.695 | 75.0 | 11.0 |
| K1 periodic | 243.67 | 0.698 | 75.0 | 11.0 |
| K2 periodic | 241.54 | 0.695 | 75.0 | 11.0 |
| K1 triggered | 242.89 | 0.698 | 76.0 | 10.0 |
| K2 triggered | 239.84 | 0.701 | 77.7 | 8.3 |

| Policy | Band record | Hidden PC MSE | Spectra | Calibrations |
|---|---|---|---|---|
| Model-free RL | 244.85 | 0.693 | 115.3 | 10.1 |

Table S19. Intrinsic-reference diagnostics under the response-maximization objective.

| Policy | Mean gap to 255.0 | Available improvement realized | Position error (px) |
|---|---|---|---|
| Fixed fast | 8.53 | 63.5% | 7.20 |
| Raw periodic | 13.22 | 48.4% | 1.29 |
| K1 periodic | 11.36 | 51.7% | 1.36 |
| K2 periodic | 13.49 | 49.7% | 1.14 |
| Model-free RL | 10.18 | 55.0% | 3.46 |

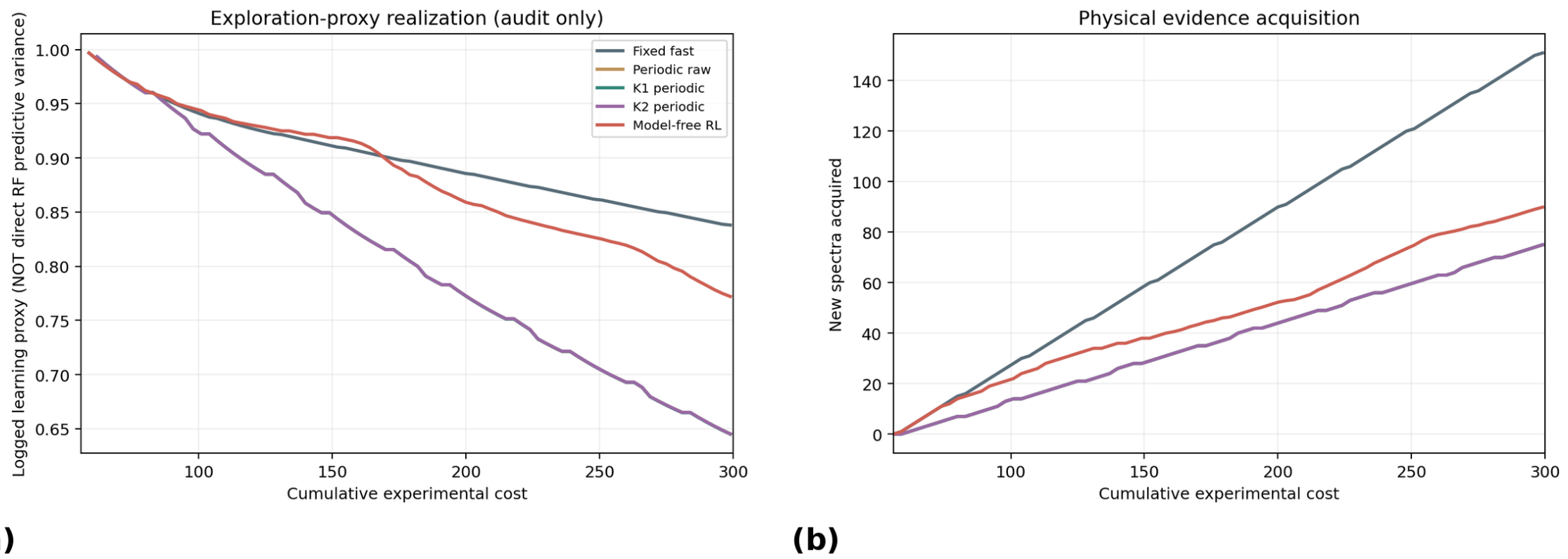

Figure S29. Controller-visible exploration uncertainty proxy and accumulated spectroscopy count versus acquisition cost. The proxy is not itself an oracle prediction-error curve; differences in uncertainty and error remain independent evaluation questions.

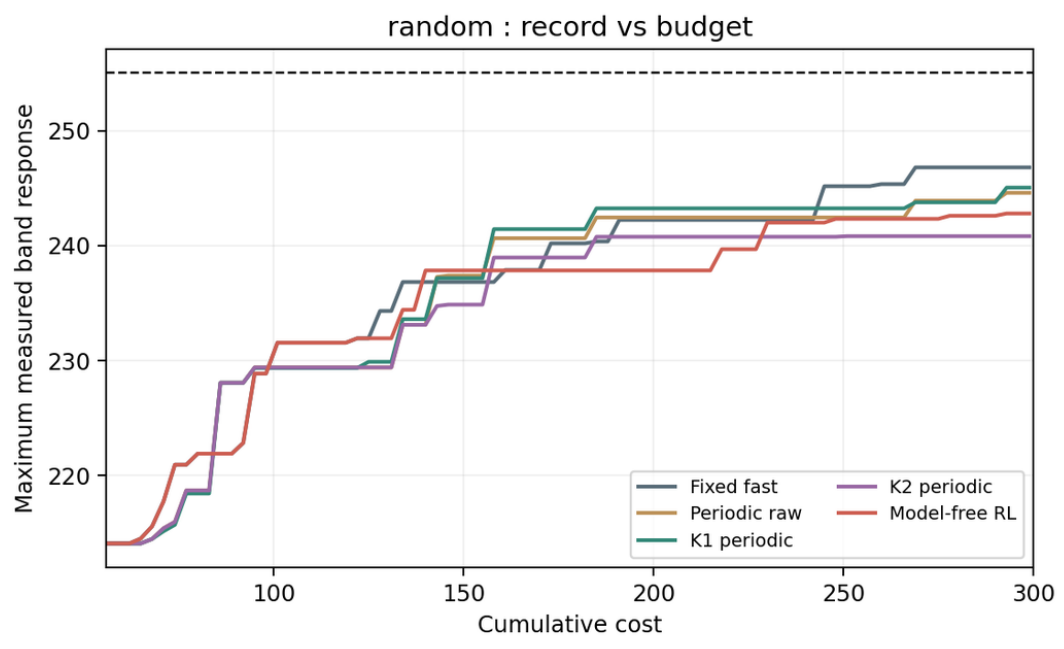

**(a)**

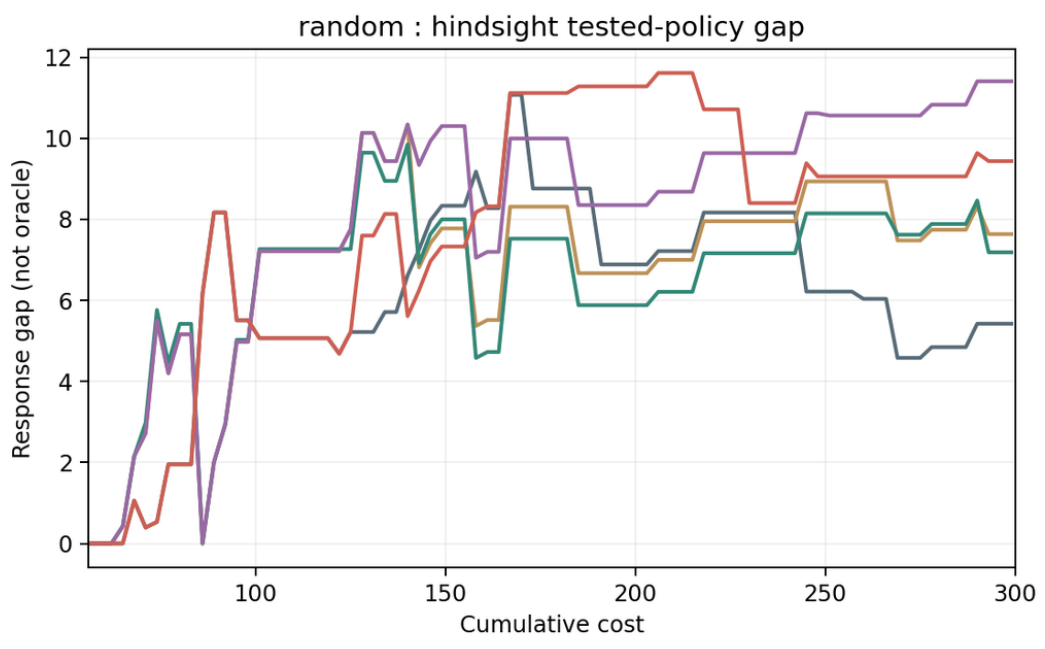

**(b)**

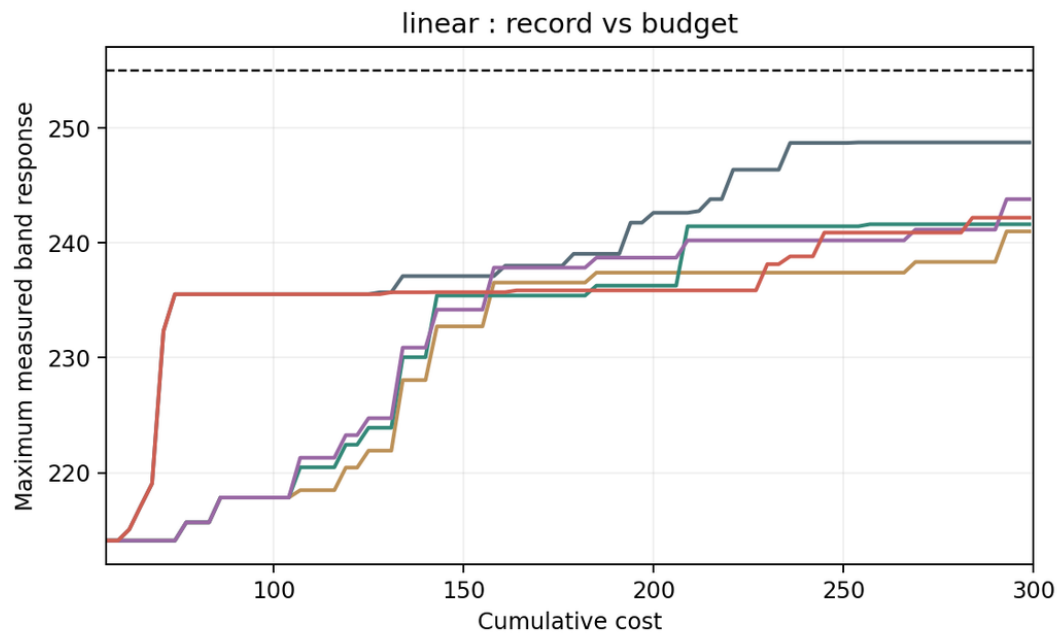

**(c)**

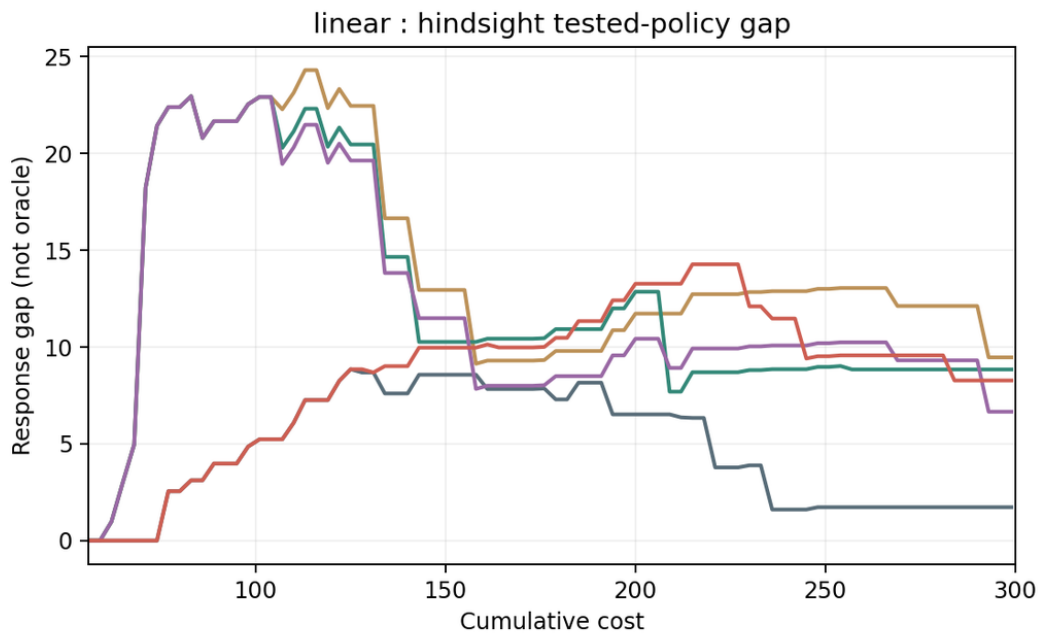

**(d)**

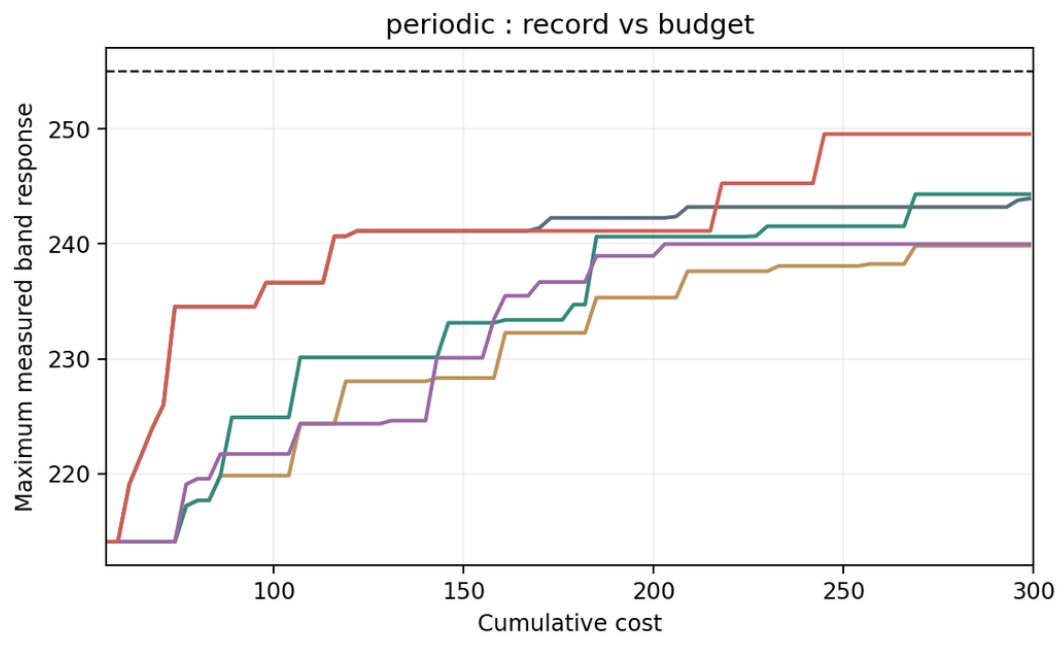

**(e)**

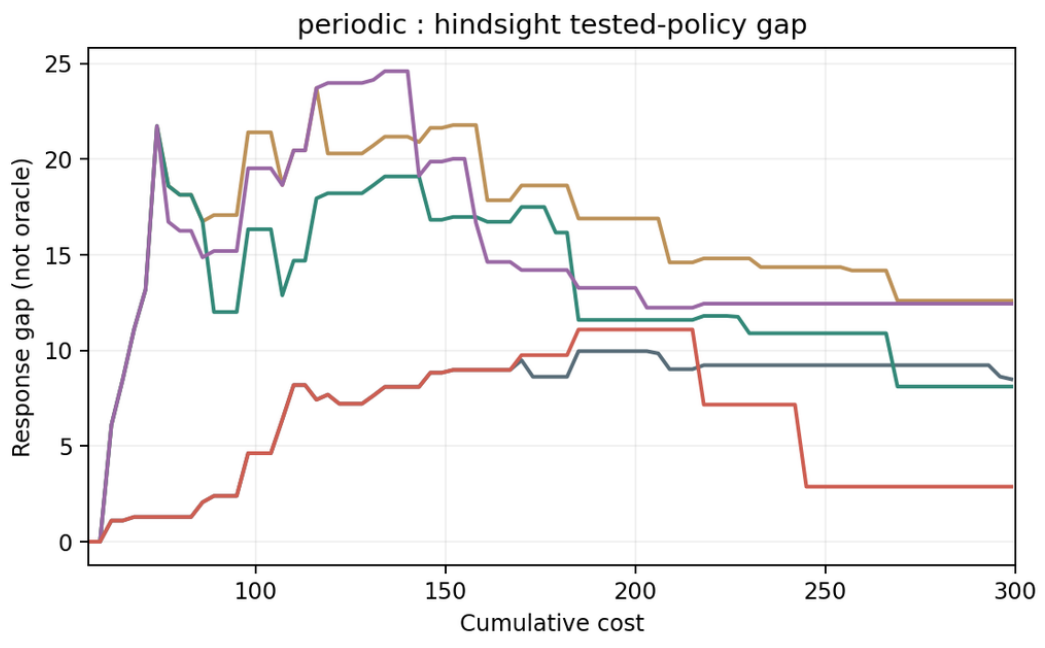

**(f)**

Figure S30. Time-resolved response realization under random, linear, and linear-plus-periodic drift. Each panel uses the same accessible full-field response ceiling but different simulated drift histories; absolute records need not reach the ceiling at a finite budget.

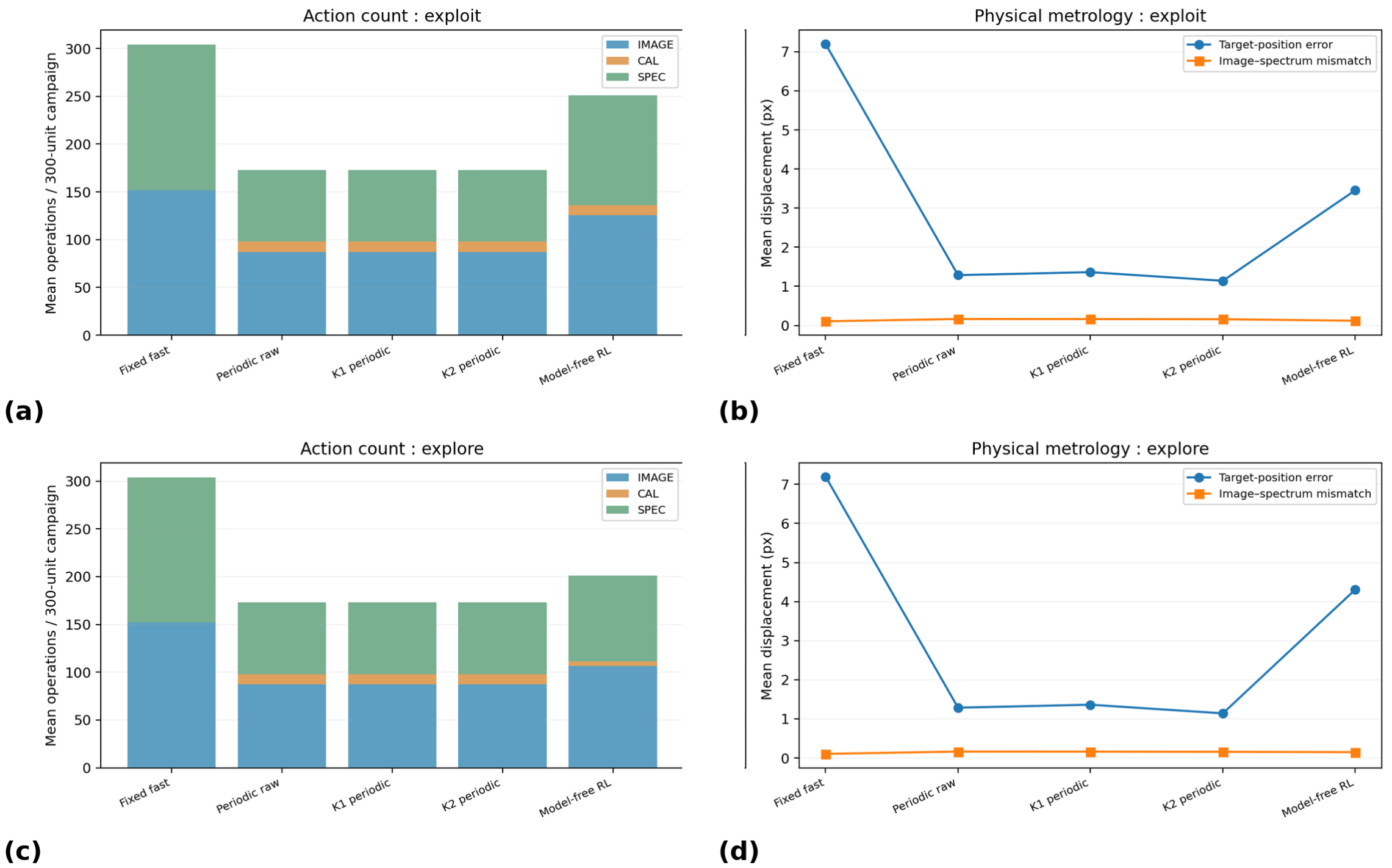

Figure S31. Full operation allocation for the evaluated controllers. Imaging, calibration, spectroscopy, and resulting overhead must be considered together; the same registration accuracy can imply different measurement throughput.

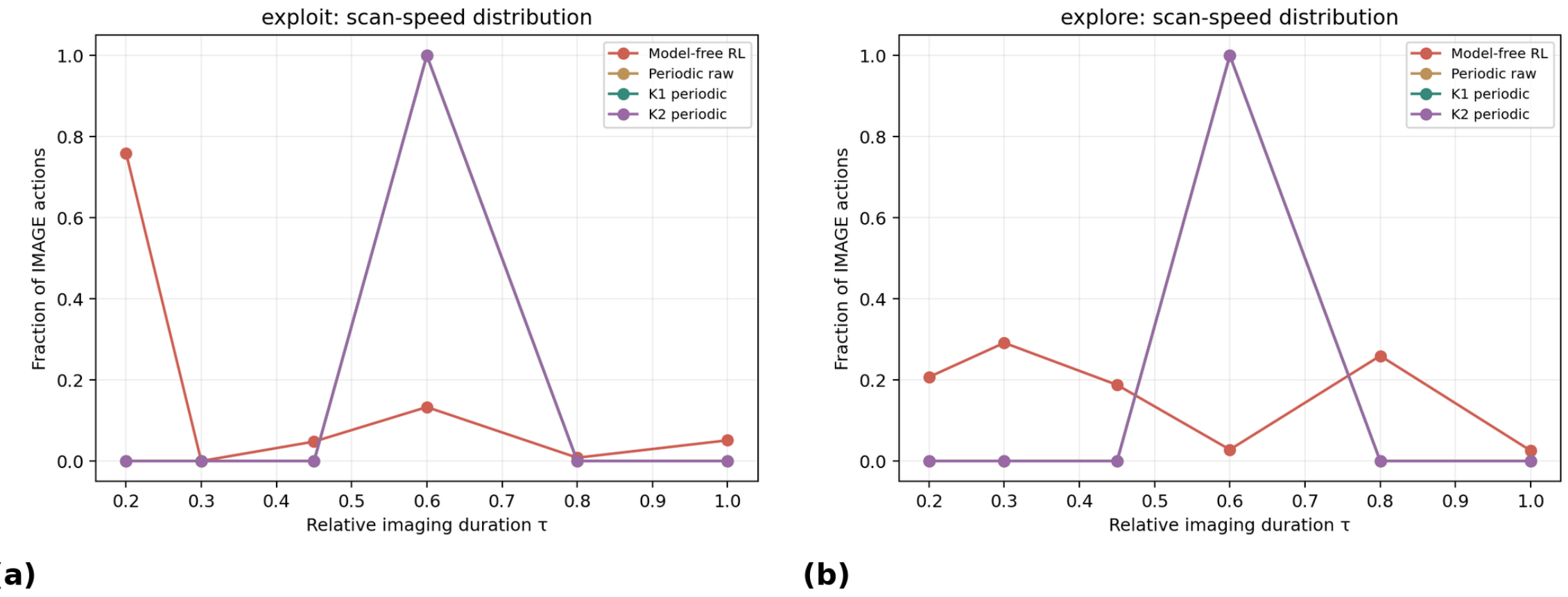

Figure S32. Scan-duration choices and imaging-speed distributions selected over the campaigns. Fast imaging causes greater modeled directional blur but costs less instrument time; slow imaging improves sharpness at greater expense.

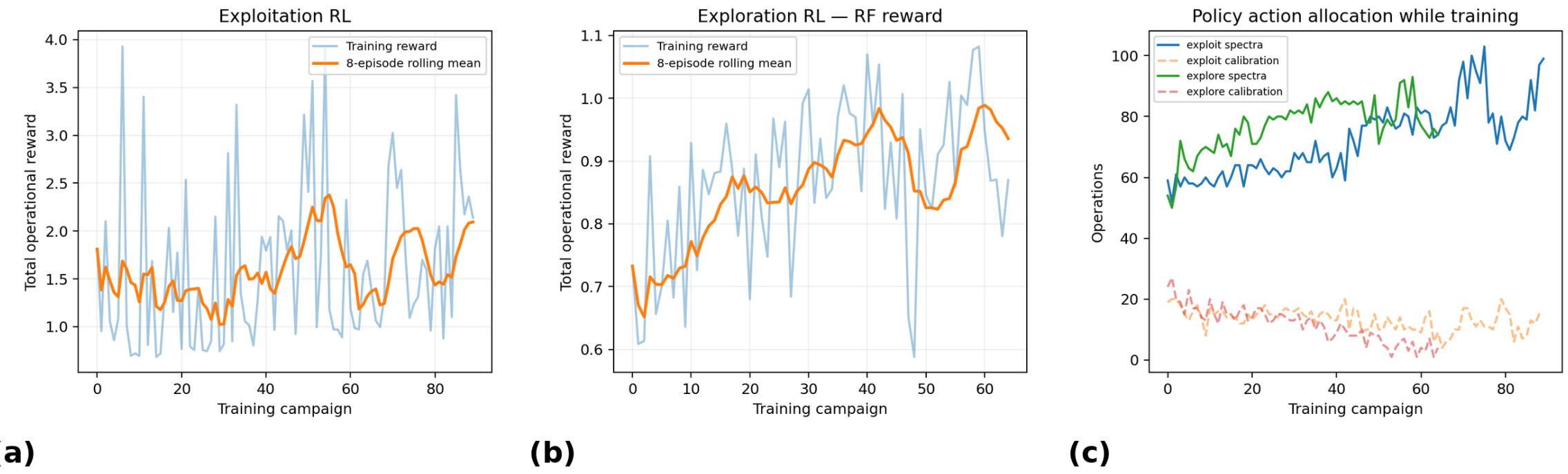

Figure S33. Training telemetry for model-free RL policies optimized separately for discovery and RF-uncertainty reduction. These are finite bounded training curves and do not establish convergence or superiority to periodic calibration.

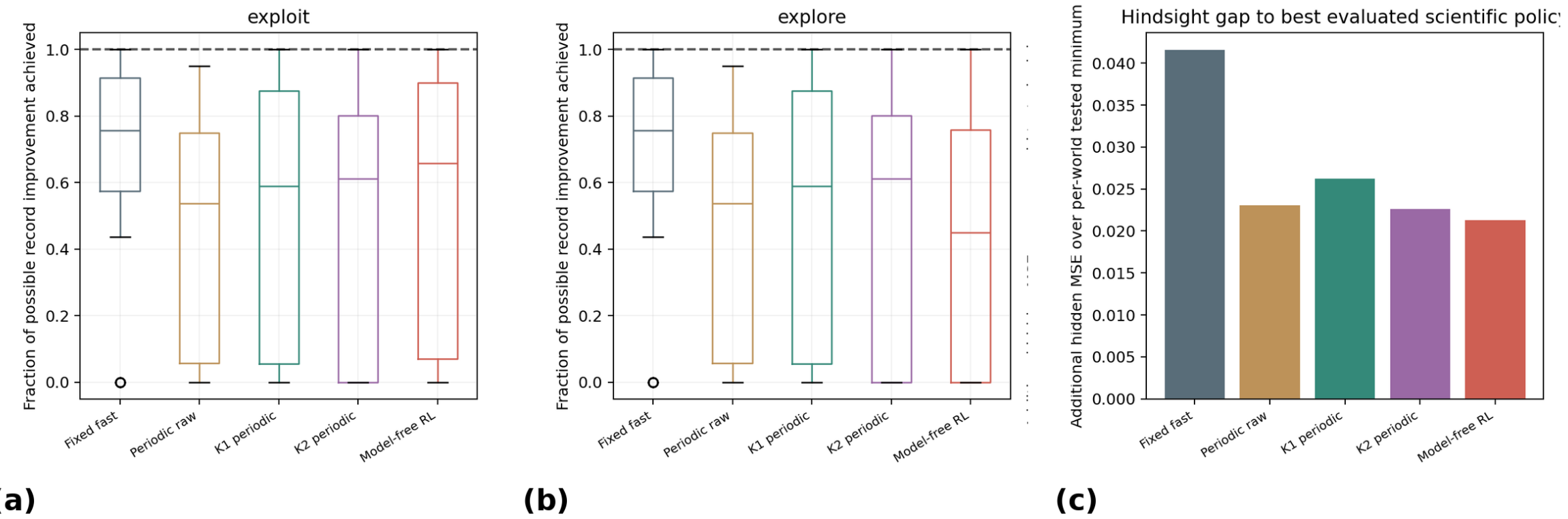

Figure S34. Spread of oracle-referenced discovery realization across held-out drift realizations. The full-field ceiling is fixed, while individual response records and fractions of available gain vary with initialization and acquired trajectory.

K1 and K2 have identical nominal imaging and calibration counts under the matched periodic schedule, allowing a direct comparison of position estimation. In the reported means, K2 has smaller residual position error, but the hidden scientific prediction differences are much smaller. The model-free discovery policy improves hidden prediction error over uncontrolled fast acquisition, yet does not establish a clear advantage over periodic calibration. On the exploration objective, lower internal uncertainty does not necessarily imply lower hidden error. Together, these comparisons identify practical quality–throughput trade-offs and make the incremental value required of a more elaborate controller explicit.

### S10.6 Methodological limitations and evidence boundaries

All spectroscopy targets arise from one reconstructed experimental field. Training and held-out measurement centers are intermingled, and their structural patches overlap; the analysis concerns within-specimen interpolation. The drift traces, temporal blur and command correction are simulated rather than recorded at the microscope. Spectra at displaced positions use nearest available measured responses where a direct fractional-position experimental spectrum does not exist. Large drift near the scene boundary samples reflected image content. The comparison has finite RL training, a structured K2 hypothesis bank favorable to periodic drift, and triggered-calibration thresholds that may not be globally optimal. The initial budget and common fixed-panel resources are charged according to the logged $B = 300$ replay rather than inferred from the $B = 150$ workflow.

Main Figure 8 and Figures S24–S26 document the static drift analysis; Figures 9–11, S27–S34, and Tables S15–S19 document the distinct scan-speed and $B = 300$ studies. They do not change the results of the first campaign. They do not demonstrate live-microscope drift correction or optimality of a clairvoyant controller.